\documentclass[runningheads]{llncs}
\usepackage{graphicx}

\usepackage{tikz}
\usepackage{comment}
\usepackage{amsmath,amssymb} 
\usepackage{color}
\usepackage{epsfig}
\usepackage{graphicx}
\usepackage{amsmath}
\usepackage{amssymb}
\usepackage{subcaption}
\usepackage{multirow}
\usepackage{booktabs}
\usepackage[breaklinks=true,
            bookmarks=false,
            colorlinks,
            linkcolor=red,       
            anchorcolor=blue,  
            citecolor=green, 
            ]{hyperref}
\usepackage{wrapfig}

\usepackage[accsupp]{axessibility}  

\mainmatter
\def\ECCVSubNumber{859}  

\title{MovieCuts: A New Dataset and Benchmark for Cut Type Recognition} 

\newcommand{\etal}{\textit{et al}. }
\newcommand{\ie}{\textit{i}.\textit{e}. }
\newcommand{\eg}{\textit{e}.\textit{g}. }
\newcommand{\wrt}{w.r.t. }

\titlerunning{MovieCuts: A New Dataset and Benchmark for Cut Type Recognition}
\author{Alejandro Pardo\inst{1}\index{Pardo, Alejandro} \and
Fabian Caba Heilbron\inst{2}\index{Caba Heilbron, Fabian} \and
Juan Le\'on Alc\'azar\inst{1}\index{Le\'on Alc\'azar, Juan} \and \\
Ali Thabet\inst{1,3}\index{Thabet, Ali} \and
Bernard Ghanem\inst{1}\index{Ghanem, Bernard}
}
\authorrunning{Alejandro Pardo}
\institute{King Abdullah University of Science and Technology, KAUST
\email{\{alejandro.pardo,juancarlo.alcazar,ali.thabet,bernard.ghanem\}@kaust.edu.sa}\\
\and 
Adobe Research \email{\{caba\}@adobe.com}
\and 
Meta Reality Labs \email{\{thabetak\}@meta.com} 
}

\begin{document}

\maketitle

\begin{abstract}
Understanding movies and their structural patterns is a crucial task in decoding the craft of video editing. While previous works have developed tools for general analysis, such as detecting characters or recognizing cinematography properties at the shot level, less effort has been devoted to understanding the most basic video edit, \emph{the Cut}.
This paper introduces the Cut type recognition task, which requires modeling multi-modal information. To ignite research in this new task, we construct a large-scale dataset called MovieCuts, which contains $173,967$ video clips labeled with ten cut types defined by professionals in the movie industry.
We benchmark a set of audio-visual approaches, including some dealing with the problem's multi-modal nature. Our best model achieves $47.7\%$ mAP, which suggests that the task is challenging and that attaining highly accurate Cut type recognition is an open research problem.
Advances in automatic Cut-type recognition can unleash new experiences in the video editing industry, such as movie analysis for education, video re-editing, virtual cinematography, machine-assisted trailer generation, machine-assisted video editing, among others. Our data and code are publicly available: \href{https://github.com/PardoAlejo/MovieCuts}{https://github.com/PardoAlejo/MovieCuts}.

\end{abstract}

\section{Introduction}
\label{sec:introduction}

%

Professionally edited movies use the film grammar \cite{filmgrammar} as a convention to tell visual stories. 
Through the lenses of the film grammar, a movie can be deconstructed into a hierarchical structure: a string of contiguous frames form a shot, a sequence of shots build a scene, and a series of scenes compose the movie.
Typically, scenes portray events that happen in single locations using shots recorded with a multi-camera setup \cite{katz2005film}.
Like punctuation in the written grammar, careful transition between shots is also an important component of the film grammar. Indeed, shot transitions can be viewed as the most basic video editing device \cite{kozlovic2007anatomy,tsivian2009cinemetrics}. They create changes of perspective, highlight emotions, and help to advance stories \cite{kozlovic2007anatomy,tsivian2009cinemetrics}. 
Several types of \textit{soft} shot transitions like wipes, or fades are commonly used between scenes. However, within a scene, the most used transition between shots \cite{cutting2016evolution} is the Cut, which simply joins two shots without any special effect.
\begin{figure}[t]
    \raggedleft
        \includegraphics[width=1.0\linewidth]{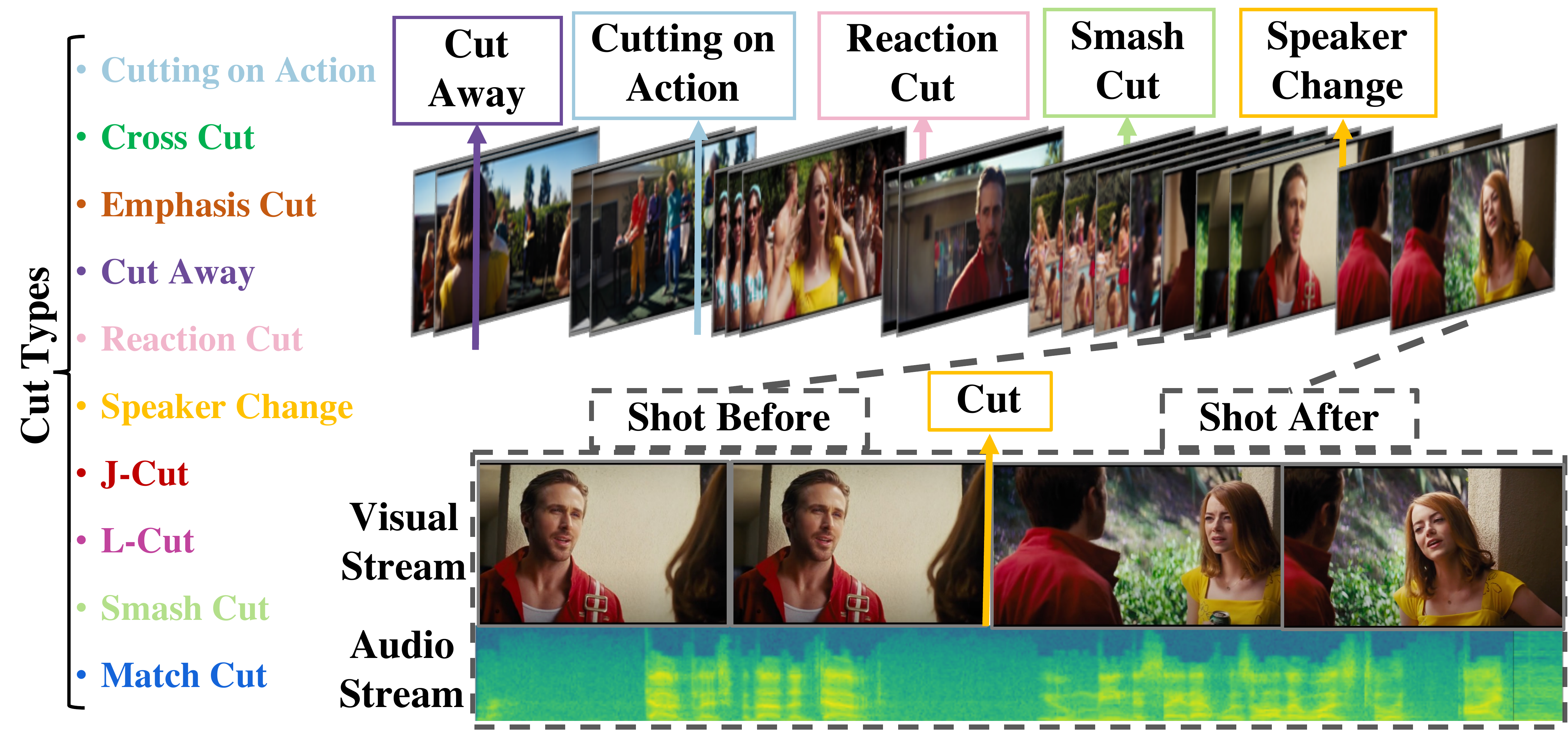}
    \caption{\textbf{Cut Type Recognition Task.}
    Within a movie scene, most of the shot transitions use the \textit{Straight Cut}, or Cut. Cuts are designed to preserve audio-visual continuity \cite{smith2008edit} across time, space, or story, and can be classified by their the semantic meaning \cite{smith2012window}. This figure illustrates a scene with different cut types happening one after the other. Towards the end of the scene, there is a dialogue portrayed by cutting when the active speaker changes. Understanding this task requires audio-visual analysis. The visual stream helps detect the camera change to focus on one of the two actors. The audio stream exhibits a clear change of frequencies when the cut happens. Combining these two cues allows us to predict the cut type: \textit{Speaker Change}.}

    \label{fig:teaser}
\end{figure}

Cuts in professionally edited movies are not random but rather have language, structure, and a taxonomy \cite{filmgrammar}. Each cut in a movie has a purpose and a specific meaning. Thus, to understand movie editing styles, one has to understand the cuts. Tsivian \etal introduced a platform called Cinemetrics to analyze movies by analysing their cut frequency \cite{cutting2016evolution}. While Cinemetrics is helpful in characterizing the rhythm and pace of cuts, analyzing and understanding the semantic meaning of these cuts remains a rather difficult task. 
In the computer vision community, recent works have tackled the problem of analyzing different cinematography components at the shot level \cite{wu2016analysing,wu2018thinking,huang2020movienet,rao2020local} for automatic film analysis. However, only a few works, have focused on shot transitions for film analysis \cite{wu2017analyzing} and continuity editing \cite{Pardo_2021_ICCV,galvane2015continuity}.
We argue that automatically recognizing and understanding cut types would make an important step towards computationally characterize the principles of video editing, enabling new experiences for movie analysis for education, video re-editing, virtual cinematography, machine-assisted trailer generation, and machine-assisted video editing. We showcase one example of the latter in Section \ref{sec:application}.


Figure \ref{fig:teaser} illustrates the cut type recognition task introduced in this work. 
\textbf{\emph{A Cut is composed of two adjacent shots and the transition between them.}}
Cuts are not only made of frames but also of their time-aligned sound stream. 
In many situations, sounds and speech drive the cut and shape its meaning. 
Our goal is then to recognize the cut type by analyzing the clip's audio-visual information across shots. Multiple research challenges emerge from this new multi-shot video understanding task.
First, there is a need for a high-level understanding of visual and audio relationships over time to identify the intended cut type. 
To identify Speaker Change in Figure \ref{fig:teaser}, one needs a detailed audio-visual inspection to associate the sounds before and after the cut to the corresponding actor's voice. Although it sounds trivial, small changes in the signal can change the cut type. For instance, if the speakers are in different locations, the cut type would no longer be Speaker Change but rather a Cross Cut (see Figure \ref{fig:dataset}). If the active speaker does not change after the cut, the cut type would be Reaction Cut instead. Thus, it is essential to understand the fine-grained details of both signals.
We argue that these challenges can promote the development of new techniques to address the multi-modal and multi-shot nature of the problem.\\
\indent Understanding the audio-visual properties of movies has a long-standing track of interest \cite{laptev2008learning,gu2018ava,rohrbach2017movie,moviegraphs}. 
The community has developed methods to recognize characters and speaker \cite{nagrani2018benedict,everingham2006hello,brown2020playing}, events and actions \cite{liu2020multi,gu2018ava,hoai2014thread}, story-lines \cite{bain2020condensed,moviegraphs,huang2020movienet}, shot-level cinematography properties such as shot-scale and camera motion \cite{rao2020unified,canini2013classifying}, and mine shot-sequencing patterns \cite{wu2018thinking,wu2017analyzing,tsivian2009cinemetrics}.
While these approaches have set an initial framework for understanding editing in movies, there is still a lack of automated tools that understand the most basic and used editing technique, the Cut.\\
\indent This work aims to study and bootstrap research in Cut type recognition. To do so, we introduce MovieCuts, a new large-scale dataset with manually curated Cut type annotations. 
Our new dataset contains $173,967$ clips (with cuts) labeled with ten different cut categories taken from the literature \cite{bordwell1993film,transitionsref} and movie industry \cite{transitionsblog}.
We hired professional and qualified annotators to label the cut type categories. 
MovieCuts offers the opportunity to benchmark core research tasks such as multi-modal analysis, long-tailed distribution learning, and multi-label classification. Furthermore, the study of this task, might benefit other areas like machine-assited video editing.
While we observe improvements by leveraging recent techniques for audio-visual blending \cite{wang2020makes}, there is ample room for improvement, and the task remains an open research problem.

\vspace{9pt}\noindent\textbf{Contributions.} 
Our contributions are threefold:
\noindent\textbf{(1)} We introduce the cut type recognition task. To the best of our knowledge, our work is the first to address and formalize the task from a machine learning perspective.
\noindent\textbf{(2)} We collect a large-scale dataset containing qualified human annotations that verify the presence of different cut types. We do an extensive analysis of the dataset to highlight its properties and the challenges it presents. We call this dataset MovieCuts (Section \ref{sec:dataset}).
\noindent\textbf{(3)} We implement multiple audio-visual baselines and establish a benchmark in cut type recognition (Section \ref{sec:experiments}).

\section{Related Work}
\label{sec:related}

\noindent \textbf{The Anatomy of a Movie Scene.} 
Scenes are key building blocks for storytelling in film. They are built from a sequence of shots to depict an event, action, or element of film narration. Extensive literature in film studies has analyzed and characterized the structure of a scene. It includes (among others) the properties and categorization of shots and, to the interest of our work, the type of shot transitions \cite{filmgrammar,burch2014theory,murch2001blink}. There are four basic shot transitions: the wipe, the fade, the dissolve, and the cut. Each one of these four transitions has its purpose and appropriate usage. For instance, soft transitions like wipe, fade, and dissolve, are commonly used to transition between scenes and evoke a passage of time or change in location. Our work studies the cut, the instantaneous (hard) change from one shot to another, which is arguably the most frequently used. 

Film theory has developed multiple taxonomies to organize the types of cuts \cite{filmgrammar,editgrammar,bordwell1993film}. Case in point, Thompson and Bowen \cite{editgrammar} divide the types of cuts (or edits) into five different categories: action edit, screen position edit, form edit, concept edit, and combined edit. While such categorization provides a high-level grouping, it is too coarse. The categorization centers around the emotional aspects of the edit rather than the audio-visual properties of the cut. Film courses \cite{transitionsref} and practitioners \cite{transitionsblog} have also developed a taxonomy of cut types. These tend to be more specific and closely describe the audio-visual properties of the shot pair forming the cut. We choose our list of Cut types based on the existing literature and narrow it down to categories that video editors recognize in their daily routine.



\vspace{2pt}\noindent \textbf{Edited Content in Video Understanding.} Edited video content such as movies has been a rich source of data for general video understanding. Such video sources contain various human actions, objects, and situations occurring in people's daily life. In the early stages of action recognition, the Hollywood Human Actions (HOHA) \cite{laptev2008learning} and HMDB51 \cite{kuehne2011hmdb}, introduced human action recognition benchmarks using short clips from Hollywood movies. 
Another group of works used a limited number of films to train, and test methods for character recognition \cite{sivic2009you}, human action localization \cite{duchenne2009automatic}, event localization \cite{liu2020multi}, and spatio-temporal action and character localization \cite{bojanowski2013finding}.
With the development of deep-learning techniques and the need for large-scale datasets to train deep models, Gu \etal proposed the AVA dataset \cite{gu2018ava}. AVA is a large-scale dataset with spatio-temporal annotations, actors, and actions, whose primary data sources are movies and TV shows. Furthermore, other works have focused on action, and event recognition across shots \cite{liu2020multi,hoai2014thread}. Finally,  Pavlakos \etal leverage information across shots from TV shows to do human mesh reconstruction \cite{pavlakos2020human}. Instead of leveraging movie data to learn representations for traditional tasks, we propose a new task to analyze movie cut types automatically. 


\noindent \textbf{Stories, Plots, and Cinematography.}
Movies and TV shows are rich in complexity and content, which makes their analysis and understanding a challenging task.
Movies are a natural multi-modal source of data, with audio, video, and even transcripts being often available. Several works in the literature focus on the task of understanding movie content. Recent works have addressed the task of movie trailer creation \cite{irie2010automatic,hesham2018smart,xu2015trailer,smith2017harnessing}, TV show summarization \cite{bost2019remembering,bost2020serial}, and automated video editing \cite{Pardo_2021_ICCV}. Moreover,
Vicol \etal proposed MovieGraphs \cite{moviegraphs}, a dataset that uses movies to analyze human-centric situations. Rohrbach \etal presented a Movie Description dataset \cite{rohrbach2017movie}, which contains audio narratives and movie scripts aligned to the movies' full-length. Using this dataset, a Large Scale Movie Description Challenge (LSMDC) has hosted competitions for a variety of tasks, including Movie Fill-In-The-Blank \cite{maharaj2017dataset}, and movie Q\&A \cite{MovieQA}, among others. 
Like LSMDC, MovieNet \cite{huang2020movienet} and Condensed Movies \cite{bain2020condensed} are big projects that contain several tasks, data, and annotations related to movie understanding.
MovieNet includes works related to person re-identification \cite{huang2020caption,xia2020online,huang2018person,Huang_2018_CVPR}, Movie Scene Temporal Segmentation \cite{rao2020local}, and trailer and synopsis analysis \cite{Xiong_2019_ICCV,huang2018trailers}. 
All these works have shown that movies have rich information about human actions, including their specific challenges. However, only a few of them have focused on artistic aspects of movies, such as shot scales \cite{rao2020unified,canini2013classifying,benini2016shot}, shot taxonomy and classification \cite{wang2009taxonomy}, and their editing structure and cinematography \cite{wu2016analysing,wu2017analyzing,wu2018thinking}. These studies form the foundations to analyze movie editing properties but miss one of the most used techniques, the Cut. Understanding cuts is crucial for decoding the grammar of the film language. Our work represents a step towards that goal.

\section{The MovieCuts Dataset}
\label{sec:dataset}

\subsection{Building MovieCuts}

\begin{figure*}[t!]
    \begin{center}
        \includegraphics[width=\linewidth]{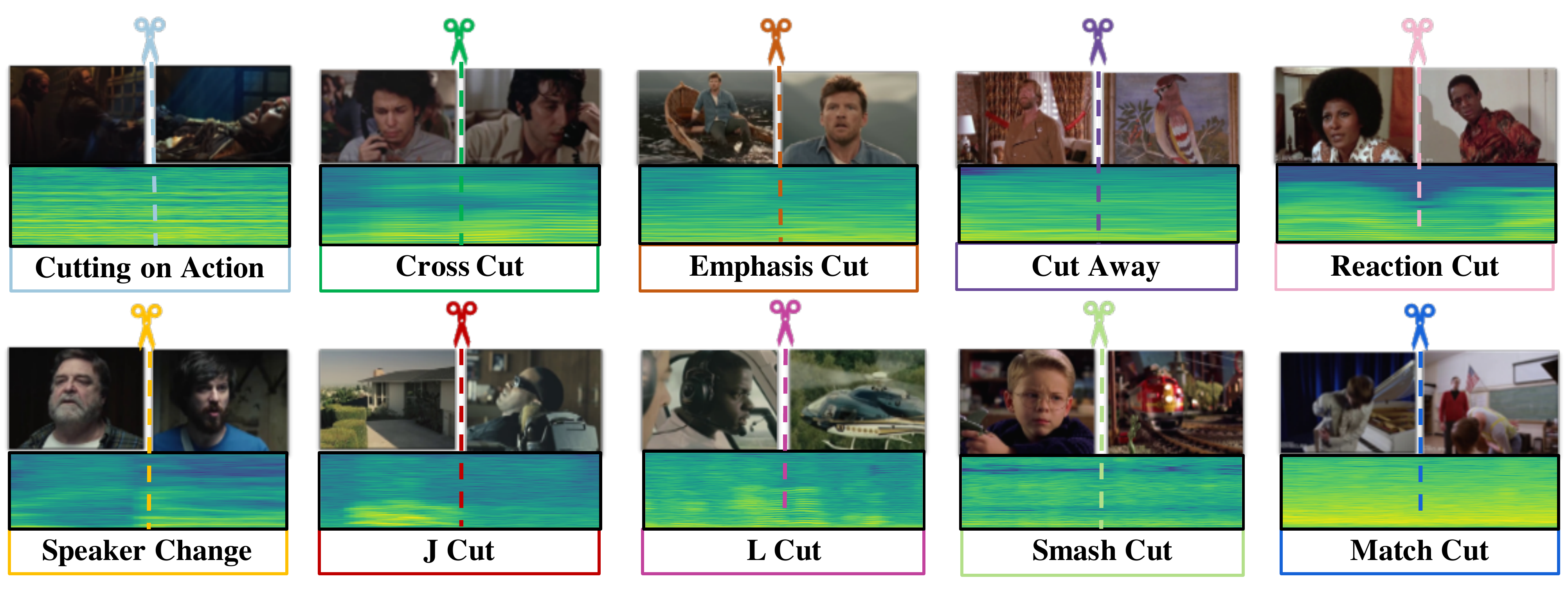}
    \end{center}
    \caption{\textbf{MovieCuts Dataset.} MovieCuts contains $173,967$ video clips labeled with 10 different Cut types. Each sample in the dataset is composed of two shots (with a cut) and their audio spectogram. Our cuts are grouped into two major categories, visual (top) and audio-visual (bottom) driven. 
    }
    \label{fig:dataset}
\end{figure*}

\noindent\textbf{Cut Type Categories}. Our goal is to find a set of cut type categories often used in movie editing. Although there exists literature in the grammar of film language \cite{filmgrammar,smith2012window} and the taxonomy of shot types \cite{wang2009taxonomy,canini2013classifying,benini2016shot}, there is no gold-standard categorization of cuts. As mentioned earlier in the related work, there exist categorization of cut types \cite{editgrammar} but it focuses on the emotional aspects of the cuts rather than the audio-visual properties of the shots composing the cut.
We gathered an initial taxonomy (17 cut types) from film-making courses (\eg \cite{transitionsref}), textbooks \cite {bordwell1993film}, and blogs \cite{transitionsblog}. We then hired ten different editors to validate the taxonomy. All the editors studied film-making, two have been nominated for the Emmys, and most have over 10 years of experience. Some of the original categories were duplicated and some of them 
were challenging to mine from movies. Our final taxonomy includes $10$ categories. Figure \ref{fig:dataset} illustrates each cut type along with their visual and audio signals:
\begin{enumerate}
    \item \textbf{Cutting on Action:} Cutting from one shot to another while the subject is still in motion.
    \item \textbf{Cross Cut:} Cutting back and forth within locations.
    \item \textbf{Emphasis Cut:} Cut from wide to close within the same shot, or the other way around. 
    \item \textbf{Cut Away:} Cutting into an insert shot of something and then back. 
    \item \textbf{Reaction Cut:} A cut to the reaction of a subject (facial expression or single word) to the comments / actions of other actors, or a cut after the reaction. 
    \item \textbf{Speaker Change:} A cut that changes the shot to the current speaker. 
    \item \textbf{J Cut:} The audio of the next shot begins before you can see it. You hear what is going on before you actually see what is going on. 
    \item \textbf{L Cut:} The audio of the current shot carries over to the next shot. 
    \item \textbf{Smash Cut:} Abrupt cut from one shot to another for aesthetic, narrative, or emotional purpose. 
    \item \textbf{Match Cut:} Cut from one shot to another by matching a concept, an action or a composition of both. \\
\end{enumerate}

\noindent\textbf{Video Collection and Processing.} We need professionally edited videos containing diverse types of cuts. Movies are a perfect source to gather and collect such data. As pointed out by Bain \etal~\cite{bain2020condensed}, there are online video channels\footnote{\href{https://www.youtube.com/c/MOVIECLIPS/videos}{MovieClips YouTube Channel is the source of scenes in MovieCuts.}} 
that distribute \textit{individual} movie scenes, thus facilitating access to movie data for research. We downloaded $9,363$ scenes, which come from $1,986$ movies. However, these movie scenes come untrimmed, and further processing is required to obtain cuts from them. We automatically detect all the cuts in the dataset with a highly accurate shot boundary detector \cite{gygli2018ridiculously} ($97.3\%$ precision and $98.5\%$ recall), which yields a total of $195,000$ candidate cuts for annotation.

\vspace{2pt}\noindent\textbf{Human Annotations and Verification}. Our goal at this stage is to collect expert annotations for $195,000$ candidate cuts. To do so, we hired Hive AI to run our annotation campaign. We choose them given their experience in labeling data for the entertainment industry. Annotators did not necessarily have a film-making background, but they had to pass a qualification test to participate in the labeling process. At least three annotators reviewed each cut/label candidate pair, and only the annotations with more than two votes were kept. The annotators also have the option to discard cuts due to: (i) errors in the shot boundary detector, and (ii) the cut not showing any of the categories in our taxonomy. We also build a handbook in partnership with professional editors to include several examples per class and guidelines on addressing edge cases. We discarded $21,033$ cuts, which left us with a total of $173,967$ cuts to form our dataset. From the $21,073$ discarded clips, we found that $12,090$ did not have enough consensus, which leads to an inter-annotator agreement of $93.8\%$. Given that inter-annotator agreement does not account for missing labels, five professional editors labelled a small subset of two thousand cuts and created a high-consensus ground truth. We found that our annotations exhibited a $90.5\%$ precision and $88.2\%$ recall when contrasted with such a gold standard.



\subsection{MovieCuts Statistics}

\begin{figure*}[t!]
    \centering
    \begin{subfigure}[t]{0.28\linewidth}
        \centering
        \includegraphics[width=0.95\linewidth]{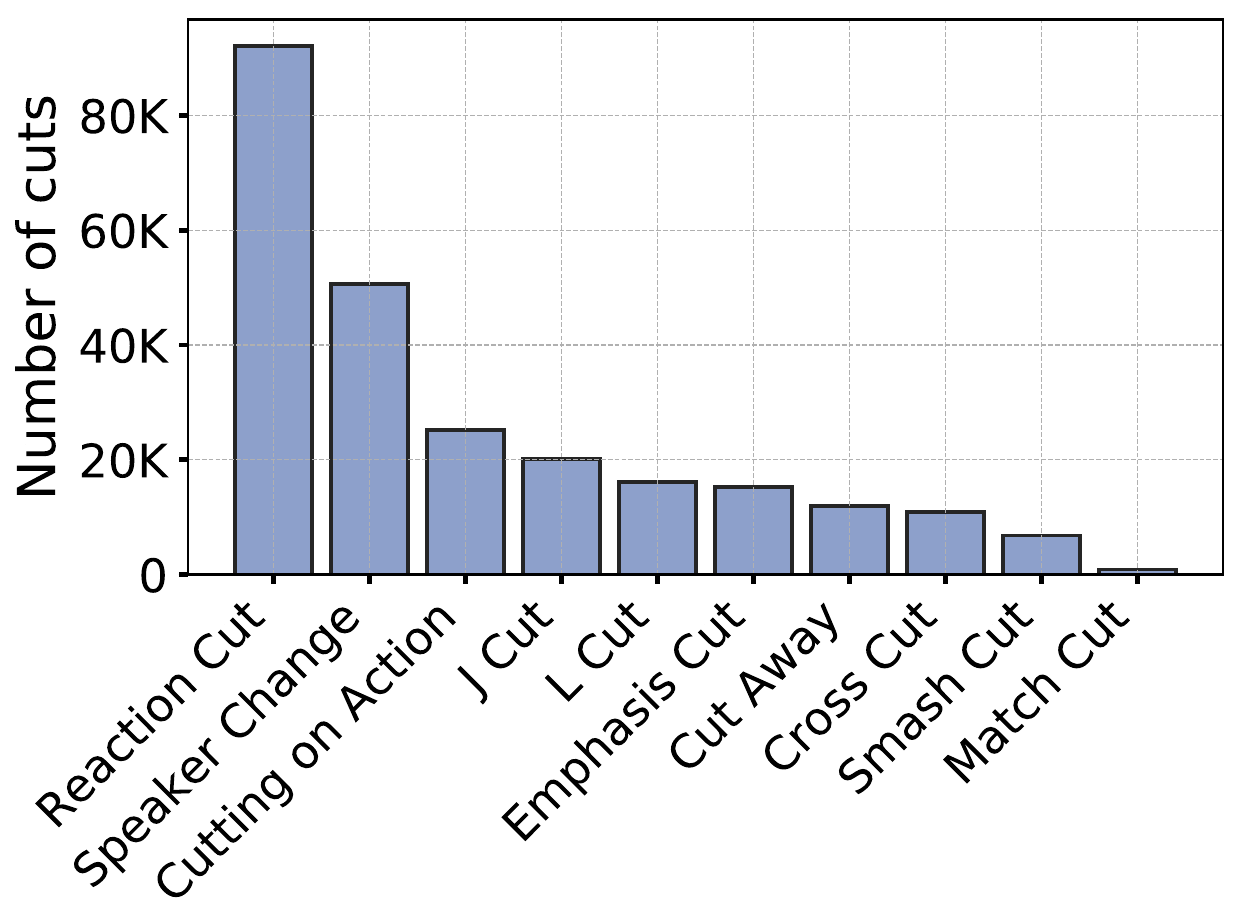}
        \caption{Cut label Distribution}
        \label{fig:dataset-statistics-label}
    \end{subfigure}%
    \hfill
    \begin{subfigure}[t]{0.38\linewidth}
        \centering
        \includegraphics[width=0.98\linewidth]{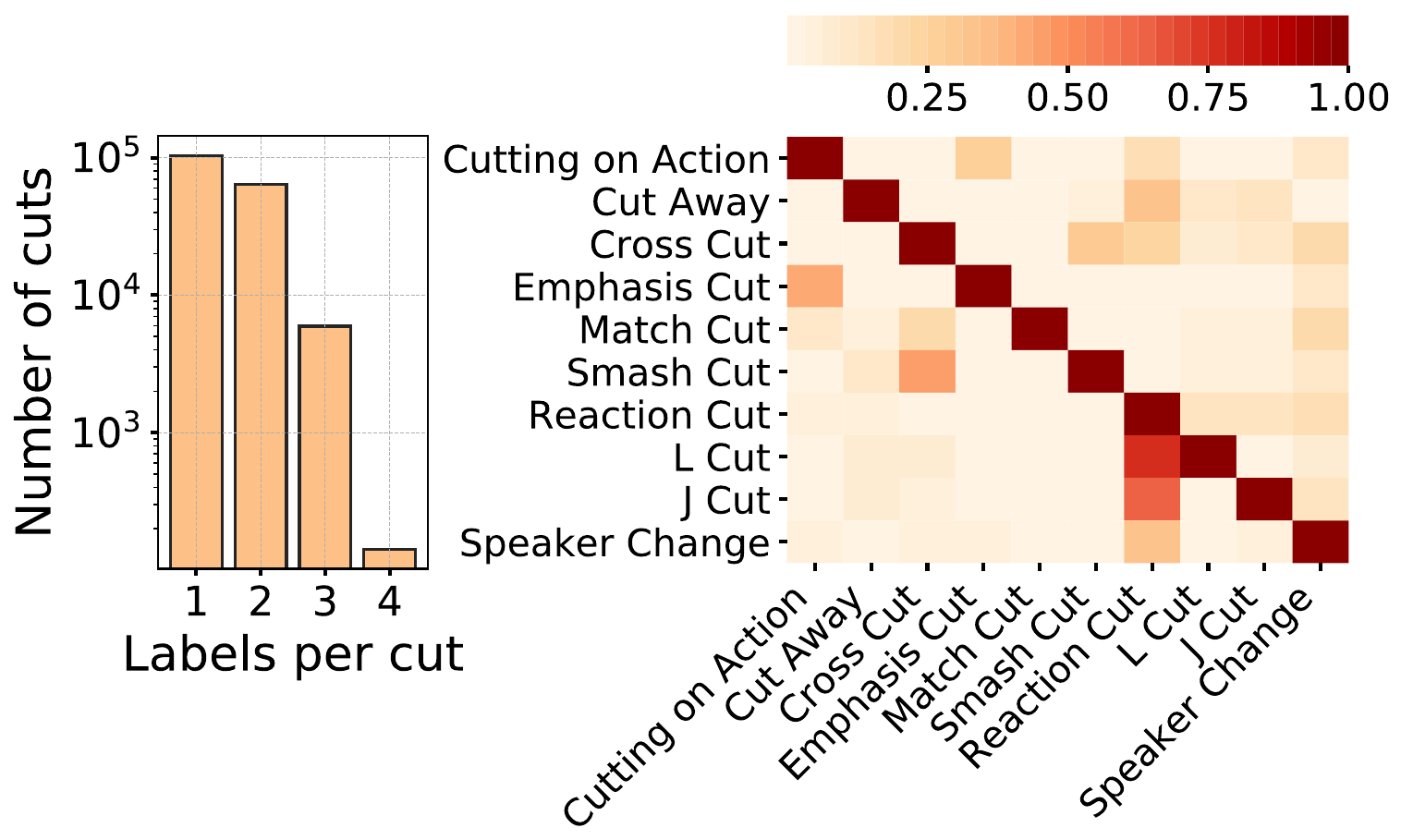}
        \caption{Multi-label distribution and co-occurrences}
        \label{fig:dataset-statistics-multilabel}
    \end{subfigure}%
    \hfill
    \begin{subfigure}[t]{0.28\linewidth}
        \centering
        \includegraphics[width=0.85\linewidth]{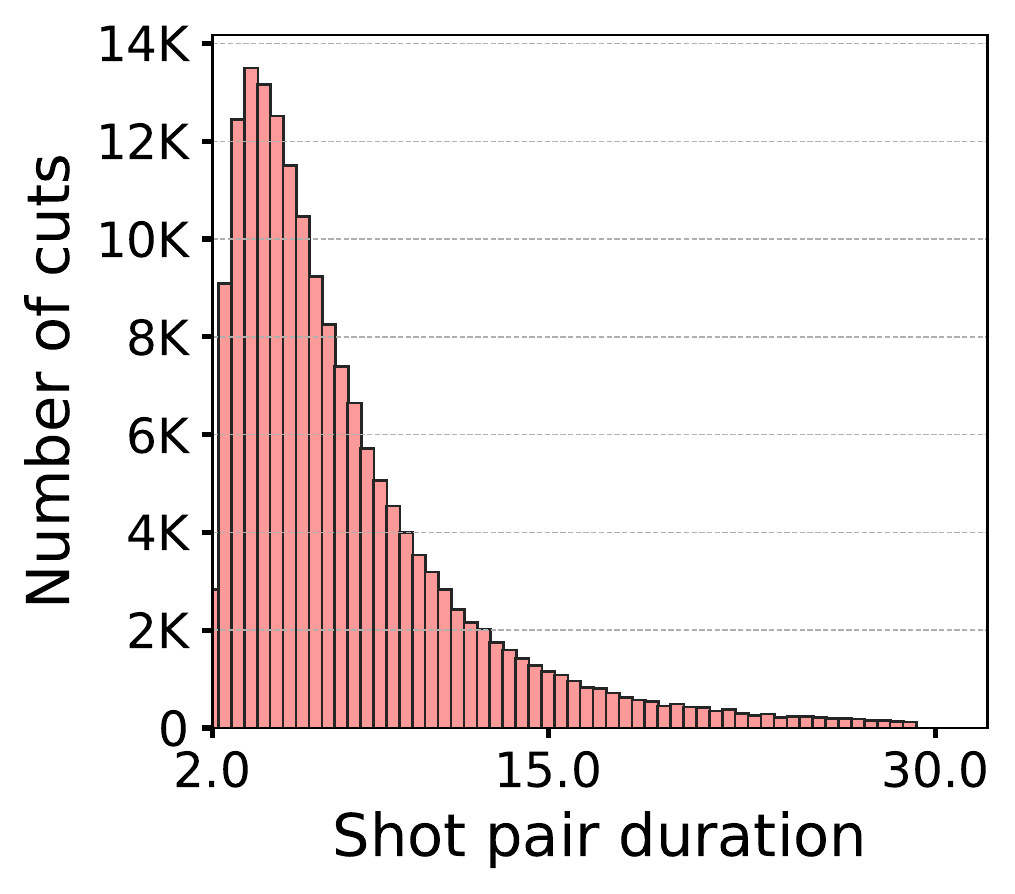}
        \caption{Duration of shot pairs}
        \label{fig:dataset-statistics-duration}
    \end{subfigure}%
    
    \begin{subfigure}[t]{0.33\linewidth}
        \centering
        \includegraphics[width=0.9\linewidth]{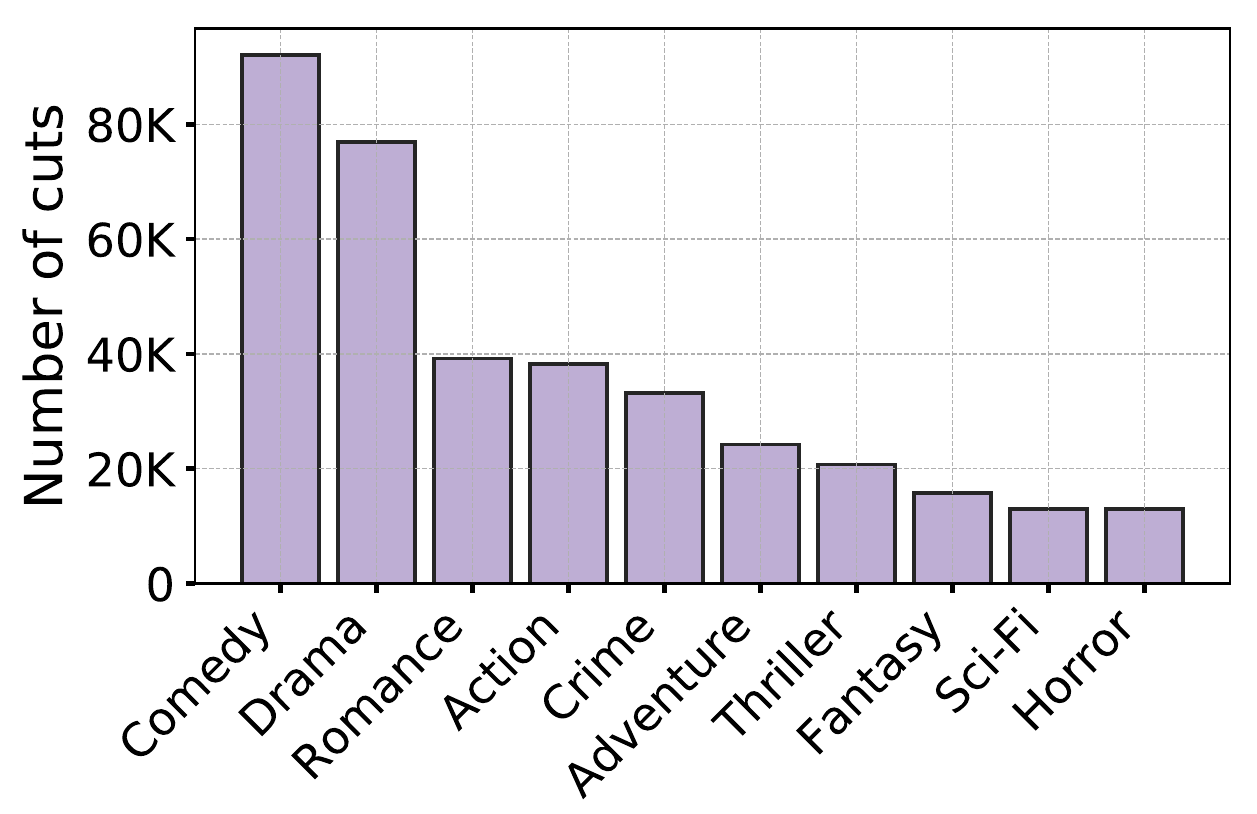}
        \caption{Cuts-per-scene by genre}
        \label{fig:dataset-statistics-genre}
    \end{subfigure}%
    \hfill
    \begin{subfigure}[t]{0.33\linewidth}
        \centering
        \includegraphics[width=0.9\linewidth]{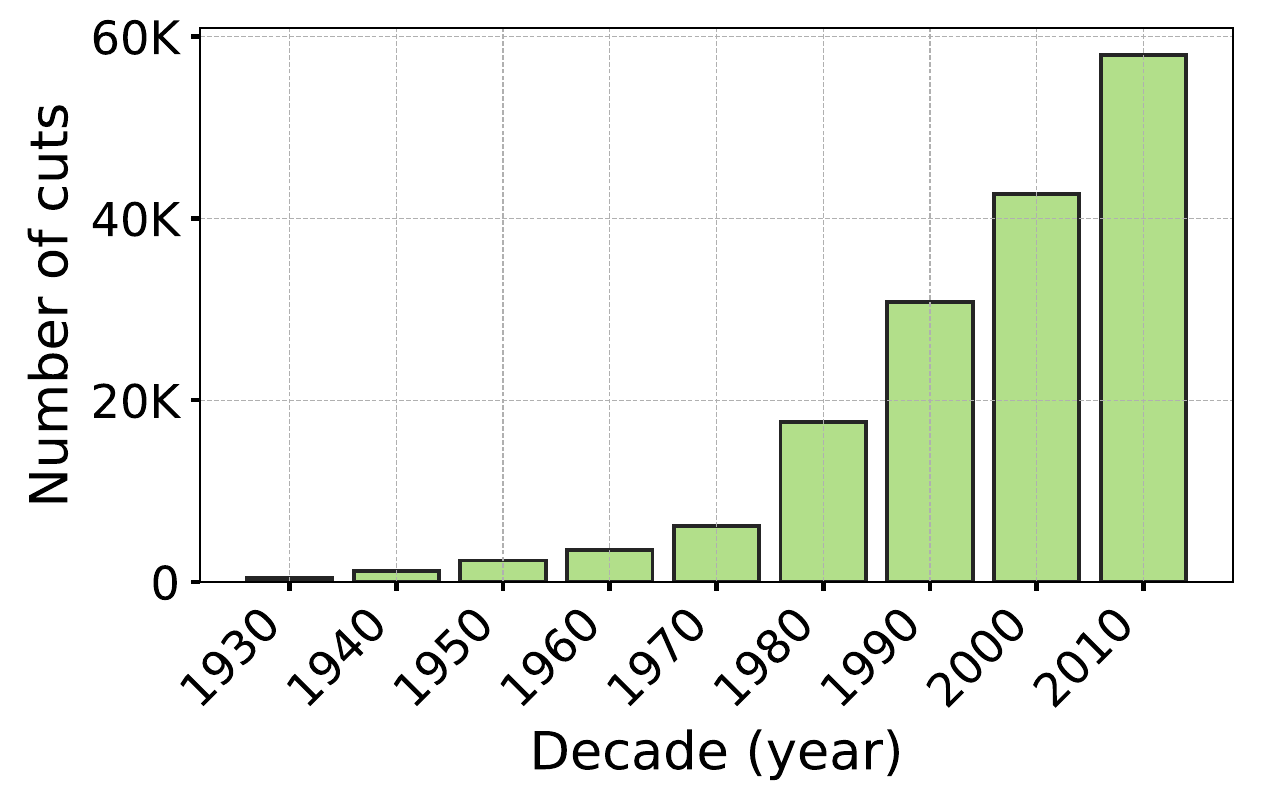}
        \caption{Cuts-per-scene by production year}
        \label{fig:dataset-statistics-year}
    \end{subfigure}%
    \hfill
    \begin{subfigure}[t]{0.3\linewidth}
        \centering
        \includegraphics[width=0.95\linewidth]{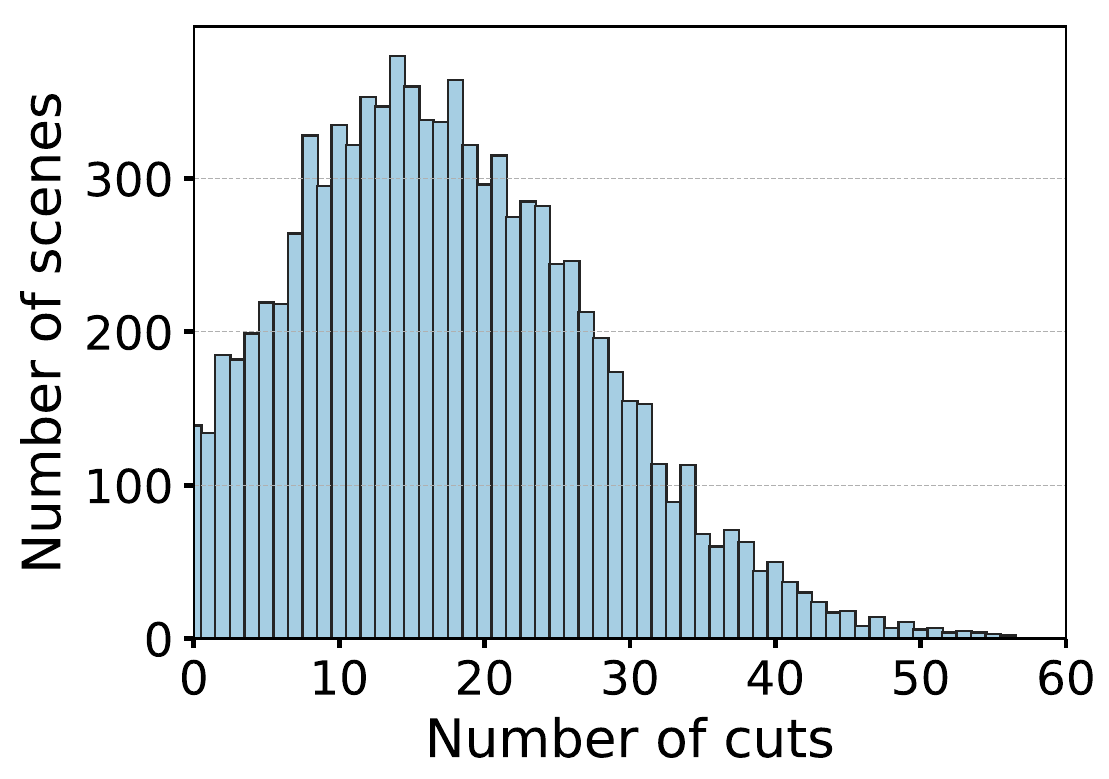}
        \caption{Cuts-per-scene distribution}
        \label{fig:dataset-statistics-scene}
    \end{subfigure}
    \caption{\textbf{MovieCuts statistics.} Figure \ref{fig:dataset-statistics-label} shows the number of instances per cut type. Labels follow a long-tail distribution. Figure \ref{fig:dataset-statistics-multilabel} indicates that a large number of instances contain more than a single cut type. Moreover, certain pairs of cut types co-occur more often. Figure \ref{fig:dataset-statistics-duration} plots the distribution of lengths (in seconds) of all the dataset instances. 
    Figure \ref{fig:dataset-statistics-genre} summarizes the production properties of the movie scenes and cuts used in our study. Figure \ref{fig:dataset-statistics-year} shows the distribution based on year of production. Finally, Figure \ref{fig:dataset-statistics-scene} shows the distribution of number of cuts per scene.}
    \label{fig:dataset-statistics}
\end{figure*}

\noindent\textbf{Cut label distribution.} Figure \ref{fig:dataset-statistics-label} shows the distribution of cut types in MovieCuts. The distribution is long-tailed, which may reflect the editors' preferences for certain types of cuts. It is not a surprise that \textit{Reaction Cut} is the most abundant label given that emotion and human reactions play a central role in storytelling. Beyond human emotion, dialogue and human actions are additional key components to advance movie storylines. We observe this in the label distribution, where \textit{Speaker Change} and \textit{Cutting on Action} are the second and third most abundant categories in the dataset. While classes such as \textit{Smash Cut} and \textit{Match Cut} emerge scarcely in the dataset, it is still important to recognize these types of cuts, which can be considered the most creative ones. We also show the distribution of cut types per movie genre in the \textit{supplementary material}.


\vspace{2pt}\noindent\textbf{Multi-label distribution and co-occurrences.} We plot in Figure \ref{fig:dataset-statistics-multilabel} the distribution of labels per cut and the co-occurrence matrix. On one hand, we observe that a significant number of cuts contain more than one label. On the other, we observe that certain pair of classes co-occur more often, \eg \textit{Reaction Cut / L Cut}. The multi-label properties of the dataset suggest that video editors compose and combine cut types quite often.  

\vspace{2pt}\noindent\textbf{Duration of shot pairs.} We study the duration of the shot pairs that surround (and form) the cuts. Figure \ref{fig:dataset-statistics-duration} shows the distribution of such shot pair duration. The most typical length is about 3.5 seconds. Moreover, we observe that the length of shot pairs ranges from 2 seconds to more than 30 seconds. In Section \ref{sec:experiments}, we study the effect of sampling different context around the cut.

\vspace{2pt}\noindent\textbf{Cut genre, year of production, and cuts per scene.} Figures \ref{fig:dataset-statistics-genre}, \ref{fig:dataset-statistics-year}, \ref{fig:dataset-statistics-scene} show statistics about the productions from where the cuts are sampled. First, we observe that the cuts are sampled across a diverse set of genres, with Comedy being the most frequent one. Second, we sourced the cuts from old and contemporary movie scenes. While many cuts come from the last decade, we also scouted cuts from movie scenes from the 1930's. Finally, we observe that the number of cuts per scene roughly follows a normal distribution with a mean of 15 cuts per scene. Interestingly, few movie scenes have a single cut, while others may contain more than 60 cuts. These statistics highlight the editing diversity in MovieCuts.

\subsection{MovieCuts Attributes}

\begin{figure*}[t!]
    \centering
    \begin{subfigure}[t]{0.37\linewidth}
        \centering
        \includegraphics[width=\linewidth]{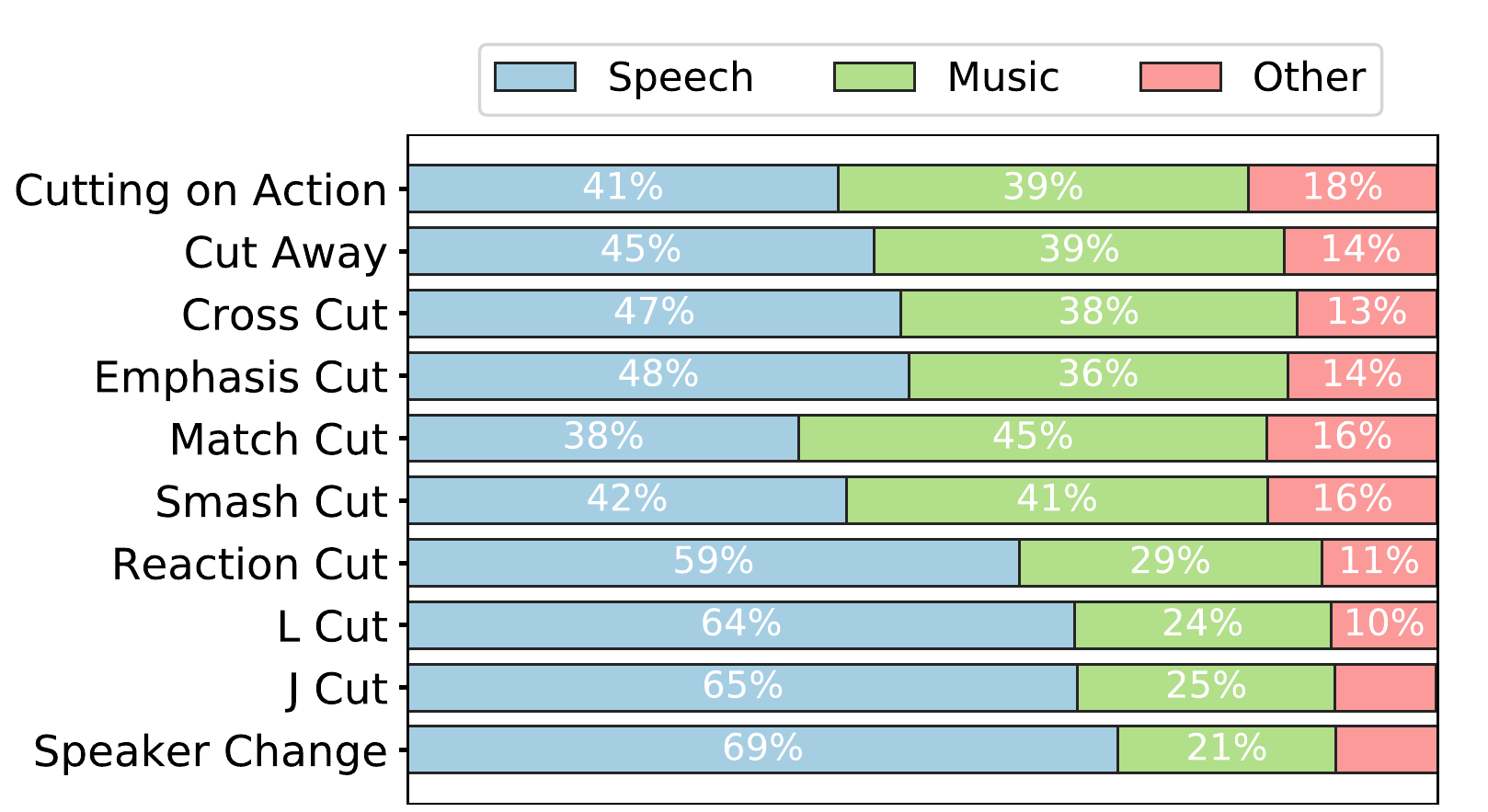}
        \caption{Sound}
        \label{fig:dataset-attributes-sound}
    \end{subfigure}%
    \begin{subfigure}[t]{0.38\linewidth}
        \centering
        \includegraphics[width=0.96\linewidth]{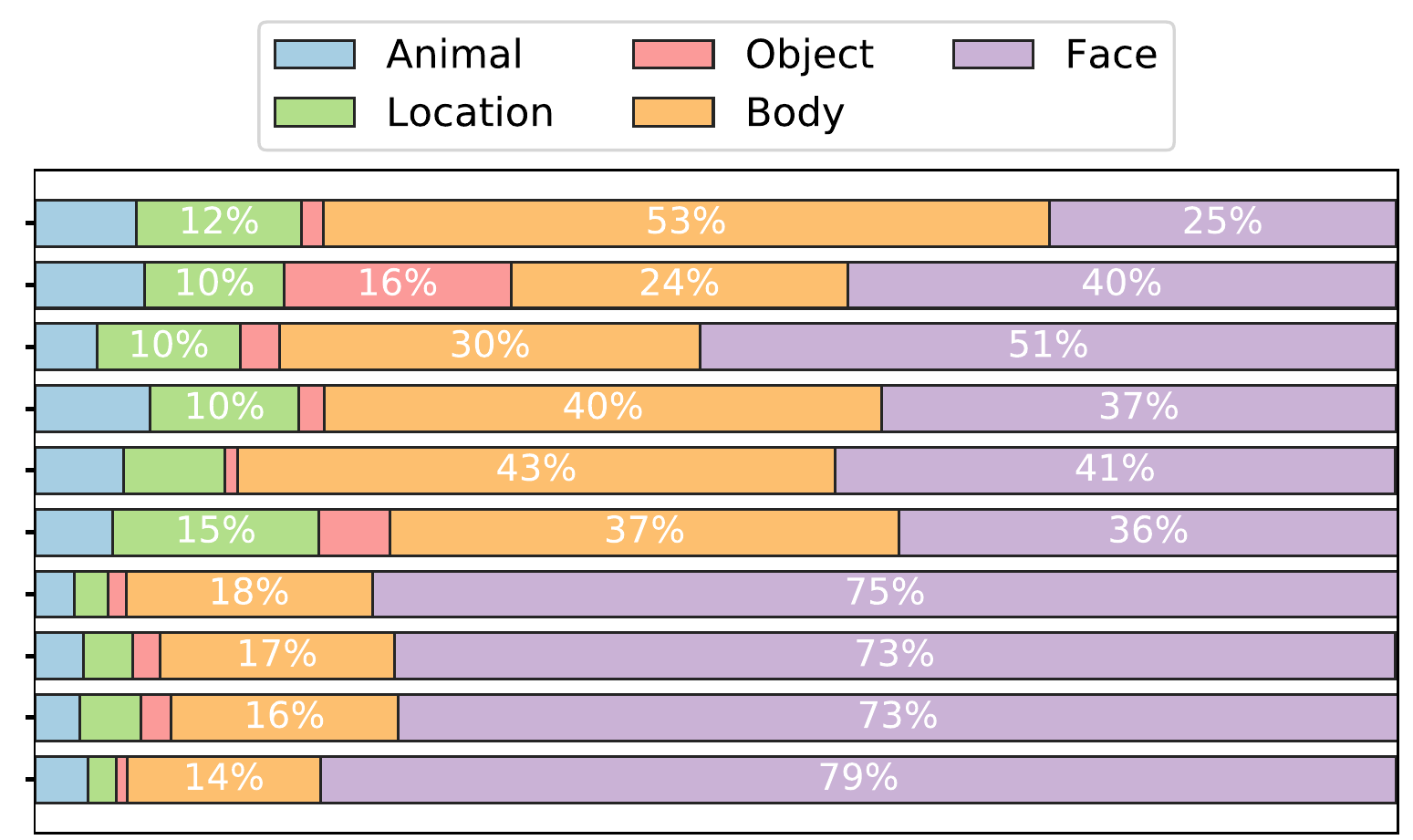}
        \caption{Subject}
        \label{fig:dataset-attributes-subject}
    \end{subfigure}%
    \begin{subfigure}[t]{0.19\linewidth}
        \centering
        \includegraphics[width=0.93\linewidth]{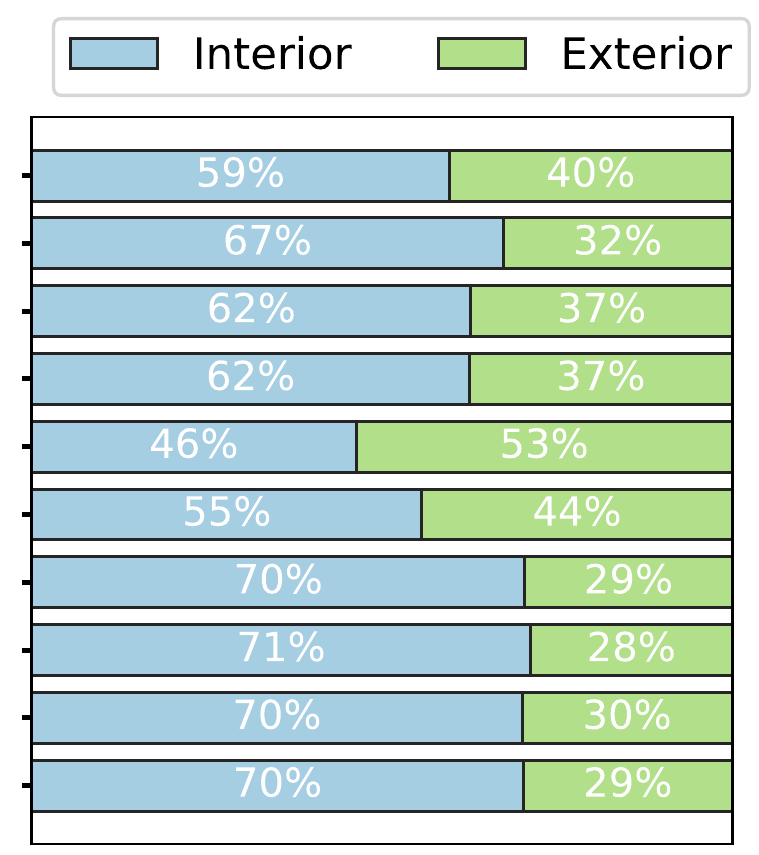}
        \caption{Location}
        \label{fig:dataset-attributes-location}
    \end{subfigure}
    \caption{\textbf{MovieCuts attributes.} MovieCuts contains diverse sounds (\ref{fig:dataset-attributes-sound}), subjects (\ref{fig:dataset-attributes-subject}), and locations (\ref{fig:dataset-attributes-location}). Some sounds co-occur more often in particular cut types. For instance, Speech is the predominant sound for dialogue related cut types such as Speaker Change or J Cut. Similarly, there exists correlation between cut types and the subjects depicted in the movie clip.}
    \label{fig:dataset-attributes}
\end{figure*}

\noindent\textbf{Sound attributes.} We leverage an off-the-shelf audio classifier \cite{Chen20} to annotate the sound events in the dataset. Figure \ref{fig:dataset-attributes-sound} summarizes the distribution of three super-groups of sound events: Speech, Music, and Other. Dialogue related cuts such as \textit{Speaker Change} and \textit{J Cut} contain a large amount of speech. Contrarily, visual-driven cuts \eg \textit{Match Cut} and \textit{Smash Cut} hold a larger number of varied sounds and background music. These attributes suggest that, while analyzing speech is crucial for recognizing cut types, it is also beneficial to model music and general sounds.

\vspace{2pt}\noindent\textbf{Subject attributes.} We build a zero-shot classifier using CLIP \cite{radford2021learning}, a neural network trained on 400M image-text pairs, to tag the subjects present in our dataset samples (\ref{fig:dataset-attributes-subject}). 

Interestingly, dialogue and emotion-driven cuts (\eg \textit{Reaction Cut}) contain many face tags, which can be interpreted as humans framed in medium-to-close-up shots. Contrarily, Body is the most common attribute in the \textit{Cutting on Action} class, which suggests editors often opt for framing humans in long shots when actions are occurring.

\vspace{2pt}\noindent\textbf{Location Attributes.} We reuse CLIP \cite{radford2021learning} to construct a zero-shot classifier of locations on our dataset. Figure \ref{fig:dataset-attributes-location} summarizes the distribution of locations (Interior/Exterior) per cut type. On one hand, we observe that most cut types contain instances shot in Interior locations 60\%-70\% of the time. On the other hand, \textit{Match Cuts} reverse this trend with the majority (53\%) of cuts shot in Exterior places. The obtained distribution suggests that stories in movies (as in real-life) develop (more often) in indoor places.

\section{Experiments}
\label{sec:experiments}

\subsection{Audio-visual Baseline} \label{section:baselines}

\begin{wrapfigure}{l}{0.55\textwidth}
    \begin{center}
        \includegraphics[width=0.9\linewidth]{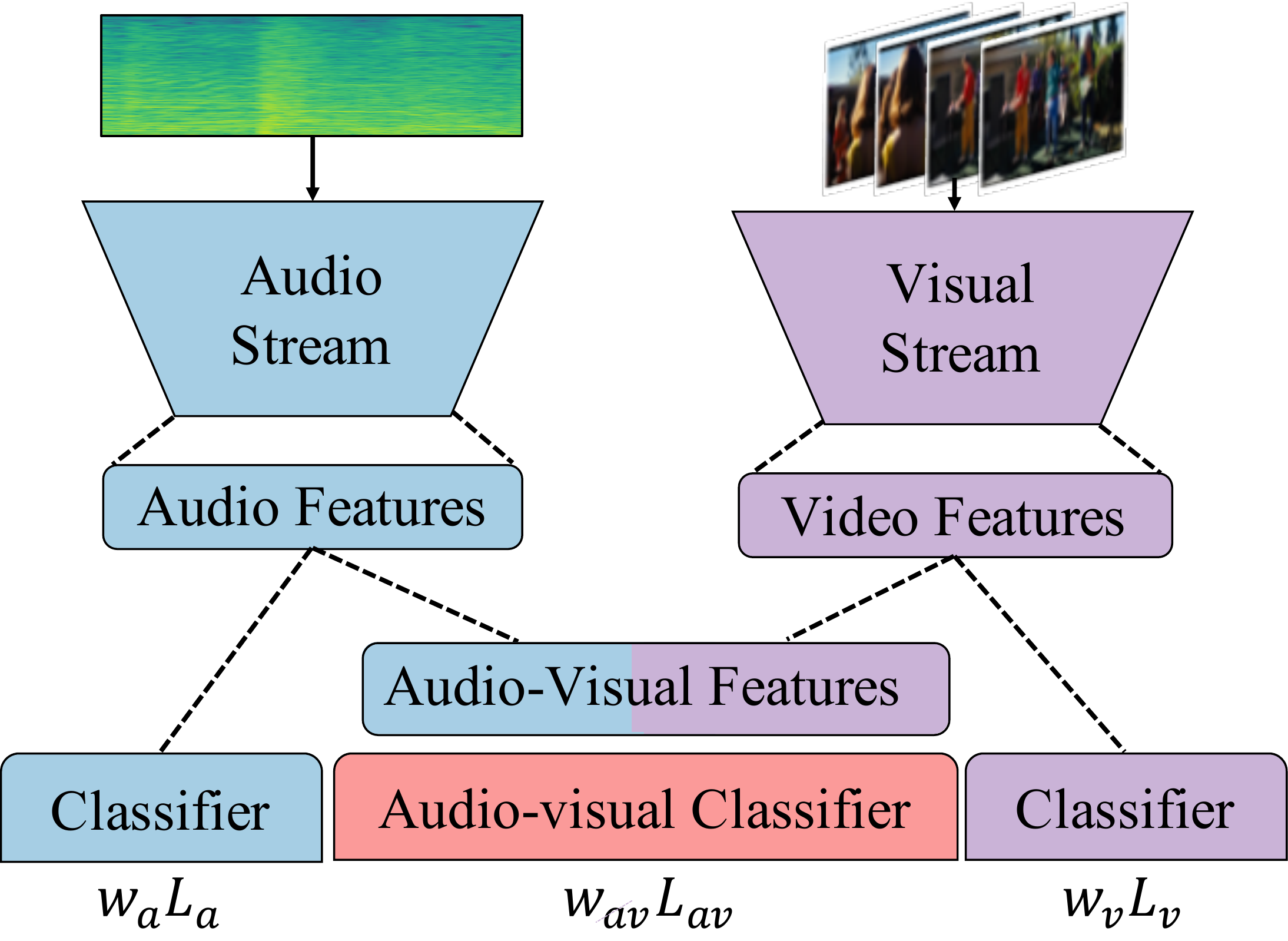}
    \end{center}
    \caption{\textbf{Audio-visual pipeline.} A late-fusion multi-modal network processes audio and visual streams. We train both networks jointly using audio loss $L_a$, visual loss $L_v$, and audio-visual loss $L_{av}$, weighted with $w_a$, $w_v$, and $w_{av}$, respectively.}
    \label{fig:pipeline}
\end{wrapfigure}

Our base architecture is shown in Figure \ref{fig:pipeline}. Similar to \cite{wang2020makes}, it takes as input the audio signal and a set of frames (clip), which are then processed by a late-fusion multi-modal CNN. 
We use a visual encoder and an audio encoder to extract audio and visual features per clip. 
Then, we form an audio-visual feature by concatenating them. 
Finally, a Multi Layer Perceptron (MLP) computes the final predictions on the audio-visual features. We optimize a binary cross-entropy (BCE) loss $\mathcal{L}$ per modality and for their combination in a one-vs-all manner to deal with the problem's multi-label nature. Our loss is summarized as:
\begin{align} \label{equation:loss}
    \text{loss} = \omega_a\mathcal{L}\left(\hat{y}_a,y\right) + \omega_v\mathcal{L}\left(\hat{y}_v,y\right) + \omega_{av}\mathcal{L}\left(\hat{y}_{av},y\right),
\end{align}

\noindent where $\omega_a$, $\omega_v$, and $\omega_{av}$ are the weights for the audio, visual, and audio-visual losses, respectively. Using this architecture, we propose several baselines: \\

\noindent\textbf{MLP classifier.}  We use the backbone as a feature extractor for each stream and train an MLP to predict on top of them and their concatenation. \\
\noindent\textbf{Encoder fine-tuning.} We train the whole backbone starting from Kinetics-400 \cite{kay2017kinetics} weights for the visual stream and from VGGSound \cite{Chen20} weights for the audio.\\ 

\noindent\textbf{Modality variants.} We train our model using the different modalities: audio only, visual only, and audio-visual. For audio-visual, we combine the losses in a naive way giving each one of them the same weight, \ie $\omega_a = \omega_v = \omega_{av}$. \\
\noindent\textbf{Modality blending.} To combine losses from multiple modalities in a more effective way, Wang~\etal~\cite{wang2020makes} proposed \textit{Gradient Blending} (GB), a strategy to compute the weight of each modality loss at training time. We use the offline GB algorithm to calculate $\omega_a$, $\omega_v$, and $\omega_{av}$. For further details, refer to Algorithms 1 and 2 of the paper~\cite{wang2020makes}.\\


\subsection{Experimental Setup}

\noindent\textbf{Dataset summary.} We divide our dataset into training, validation, and test sets by using $70\%$, $10\%$, and $20\%$ percent of the data, respectively. Thus, we use $121,423$ clips for training, $17,505$ clips for validation, and $35,039$ clips for testing. We make sure that the sets are \textit{i.i.d.}\wrt the movie genre, we show distributions per genre for each split in the supplementary material. We report all experiments on the validation set unless otherwise mentioned.  \\ 

\noindent\textbf{Metrics.} Following \cite{wu2020distribution}, we choose the mean Average Precision (mAP) across all classes and per-class AP to summarize and compare baseline performances. This metric helps to deal with MovieCuts' multi-label nature. We also report Precision-Recall curves in supplementary material. 

\noindent\textbf{Implementation details.} 
For all experiments, we use ResNet-18 \cite{he2016deep} as the backbone for both visual and audio streams. \textbf{For the audio stream}, we use a ResNet with 2D convolutions pre-trained on VGGSound \cite{Chen20}. This backbone takes as input a spectrogram of the audio signal 
and processes it as an image. To compute the spectogram we take the audios of each pair of clips, and apply consecutive Fourier Transforms with $512$ points windows with $353$ overlapping points between them. If the audio is longer than 10 seconds we trim it to 10 seconds only.
\textbf{For the visual stream}, we use a ResNet-(2+1)D \cite{tran2018closer} pre-trained on Kinetics-400 \cite{kay2017kinetics}. We sample 16 frames from a window centered around the cut as the input to the network 
Using single streams, \ie only audio or only visual, we use the features after the average pooling followed by an MLP composed of a $512 \times 128$ Fully-Connected (FC) layer followed by a ReLU and a $128 \times N$ FC-layer, where $N$ is the number of classes ($N=10$). Using two streams, we concatenate the features after the first FC-layer of the MLP to obtain an audio-visual feature per clip of size $128 \times 2 = 256$. Then, we pass it through a second FC-layer of size $256 \times N$ to compute the predictions. We train using SGD with momentum $0.9$ and weight decay of $10^{-4}$. We also use a linear warm-up for the first epoch. We train for 8 epochs with an initial learning rate of $3\times10^{-2}$, which decays by a factor of 10 after 4 epochs. We use an effective batch-size of 112 and train on one NVIDIA A100 GPU. 

\subsection{Results and Analysis}

As described in Section \ref{section:baselines}, we benchmark the MovieCuts dataset using several combinations of modalities. Results are reported in Table \ref{table:results}.

\vspace{2pt}\noindent\textbf{Linear Classifier \textit{vs.} Fine-tune}: We evaluate the performance of using frozen \textit{vs.} fine-tuned features. 
As one might expect, the fine-tuning of the backbone shows consistent improvement over all the classes regardless of the modality used. For instance, the Audio-Visual backbone performance increases from $30.82\%$ to $46.57\%$. These results validate the value of the dataset for improving the audio and visual representations encoded by the backbones.

\noindent\textbf{Modality Variants}: Consistently across training strategies,
we observe a common pattern in the results: the visual modality performs better ($43.98\%$) at the task than its audio counterpart ($27.24\%$). Nonetheless, combining both modalities still provides enhanced results for several classes and the overall mAP ($46.57\%$). 
We observe that cuts driven mainly by visual cues, such as Cutting on Action, Cut Away, and Cross Cut, do not improve their performance when audio is added. However, the rest of the classes improve when using both modalities. In particular, L Cut, J Cut, and Speaker Change improve drastically, since these types of cuts are naturally driven by audio-visual cues.

\vspace{2pt}\noindent\textbf{Gradient Blending}: The second-to-last row in Table \ref{table:results} shows the results of using the three modalities combined with the GB weights $\omega_a= 0.08$, $\omega_v = 0.57$ and $\omega_{av} = 0.35$. GB performs slightly better ($47.43\%$) than combining the losses naively ($46.57\%$), where $\omega_a=\omega_v=\omega_{av}$. 

\vspace{2pt}\noindent\textbf{Scaled Gradient Blending}: By experimenting with the Gradient Blending weights, we found that scaling them all by a constant factor can help. We empirically found that scaling the weights by a factor of 3 ($\omega_a=1.31$, $\omega_v=4.95$ and $\omega_{av}=2.74$) improves the results to $47.91\%$ mAP.

\noindent\textbf{Frame Sampling}: In addition to these experiments, we explore how to pick the frames to feed into the visual network. For all the previous experiments and as mentioned, we use \emph{Fixed Sampling} by sampling frames from a window centered around the cut. However, this is not the only strategy to sample frames. We explore two other strategies: \emph{Uniform Sampling} that takes sample frames from a uniform distribution across the two shots forming the cut, and \emph{Gaussian Sampling}, which samples the frames from a Gaussian centered around the cut. We fit both audios up to 10 seconds into the audio stream. 

\emph{Fixed Sampling} gives the best results with $47.91\%$ mAP, followed by \emph{Gaussian Sampling} with ${47.44\%}$, and \emph{Uniform Sampling} gives the lowest mAP among them with ${47.17\%}$. These results suggest that the most critical information lays around the cut. We hypothesize that the model is not good enough at handling context, architectures better at handling sequential inputs may benefit from the context of the \emph{Gaussian} or \emph{Uniform} sampling. 

\begin{table}[t!]
    \footnotesize
	\centering
    \tabcolsep=0.12cm

	\begin{tabular}{ c |c | c | c c c c c c c c c c}
		
		& Model    & \textbf{mAP} & \textbf{CA} & \textbf{CW} & \textbf{CC} & \textbf{EC}  & \textbf{MC} & \textbf{SC} & \textbf{RC} & \textbf{LC} & \textbf{JC} & \textbf{SC}  \\
		\hline
		\hline
		\multirow{3}{0.9em}{\rotatebox{90}{\textbf{Linear}}} & Audio(A)  & 23.9&	36.7&	14.8&	11.8&	14.6&	1.5&	10.7&	65.3&	15.3&	18.4&	50.0 \\
        & Visual(V)                                                      & 28.8 &	53.8&	36.3&	16.9&	19.4&	1.1&	13.3&	69.7&	12.9&	16.2&	48.0 \\
        & AV                                                & 30.8&	55.5&	32.8&	16.0&	20.3&	1.7&	13.2&	73.7&	17.4&	21.6&	56.0 \\
		\hline
        \multirow{5}{*}{\rotatebox{90}{\textbf{ Fine-tune \ }}} & Audio& 27.2&	42.6&	19.0&	14.6&	15.8&	1.5&	12.9&	69.4&	18.5&	21.3&	56.7 \\
        & Visual                                                & 44.0&	64.8&	60.8&	\underline{33.2}&	\underline{30.7}&	1.5&	21.5&	81.2&	33.7&	42.0&	70.3 \\
        & AV                                    & \underline{46.6}& \underline{65.2}&	\underline{62.5}&	31.1&	30.5&	\underline{2.0}&	\underline{22.3}&	\underline{82.9}&	\underline{43.3}&	\underline{50.0}&	\underline{75.1} \\\cline{2-13}
		& AV+GB                           & 47.4&	64.8&	62.4&	32.5&	31.6&	1.8&	23.8&	83.1&	45.6&	51.0&	77.4 \\
		& AV+SGB                 & \textbf{47.9}&	\textbf{65.6}&	\textbf{63.0}&	\textbf{34.9}&	\textbf{31.8}&	\textbf{2.3}&	\textbf{24.3}&	\textbf{83.3}&	\textbf{45.0}&	\textbf{51.6}&	\textbf{77.1} \\
		\hline
	\end{tabular}
	\caption{\textbf{Baseline comparison on MovieCuts}. We show the results of our different baselines using visual, audio, and audio-visual modalities. The last two rows use both modalities combined with Gradient Blending (GB)~\cite{wang2020makes} and Scaled Gradient Blending (SGB). All the reported results are $\%$ AP. We observe three key findings. (1) Fine-tuning on MovieCuts provides clear benefits over the linear classifier trained on frozen features. (2) Audio-visual information boosts the performance of the visual only stream. (3) Gradient Blending provides further performance gains by an optimal combination of both modalities. Showing classes: Cutting on Action (\textbf{CA}), Cut Away (\textbf{CW}), Cross Cut (\textbf{CC}), Emphasis Cut (\textbf{EC}), Match Cut (\textbf{MC}), Smash Cut (\textbf{SC}), Reaction Cut (\textbf{RC}), L Cut (\textbf{LC}), J Cut (\textbf{JC}), Speaker Chance (\textbf{SC}).
	}
	\label{table:results}
	\vspace{-15pt}
\end{table}

\begin{figure*}[ht!]
    \centering
    \begin{subfigure}[t]{0.53\linewidth}
        \centering
        \raisebox{1.4em}{\includegraphics[width=0.93\linewidth]{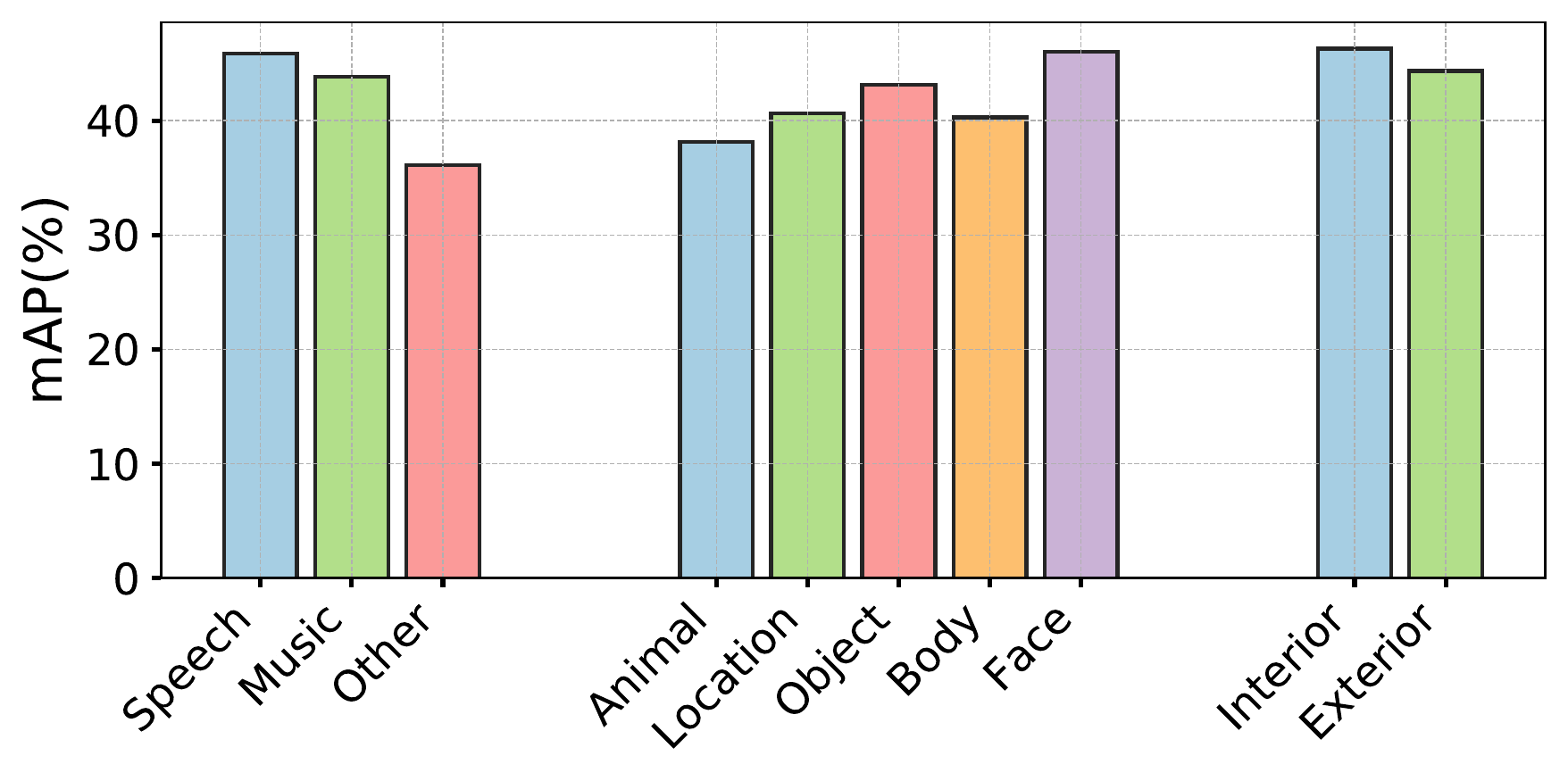}}
        \caption{Performance breakdown per attribute.}
         \label{fig:performance-breakdown-a}
    \end{subfigure}%
    \hfill
    \begin{subfigure}[t]{0.47\linewidth}
        \centering
        \includegraphics[width=\linewidth]{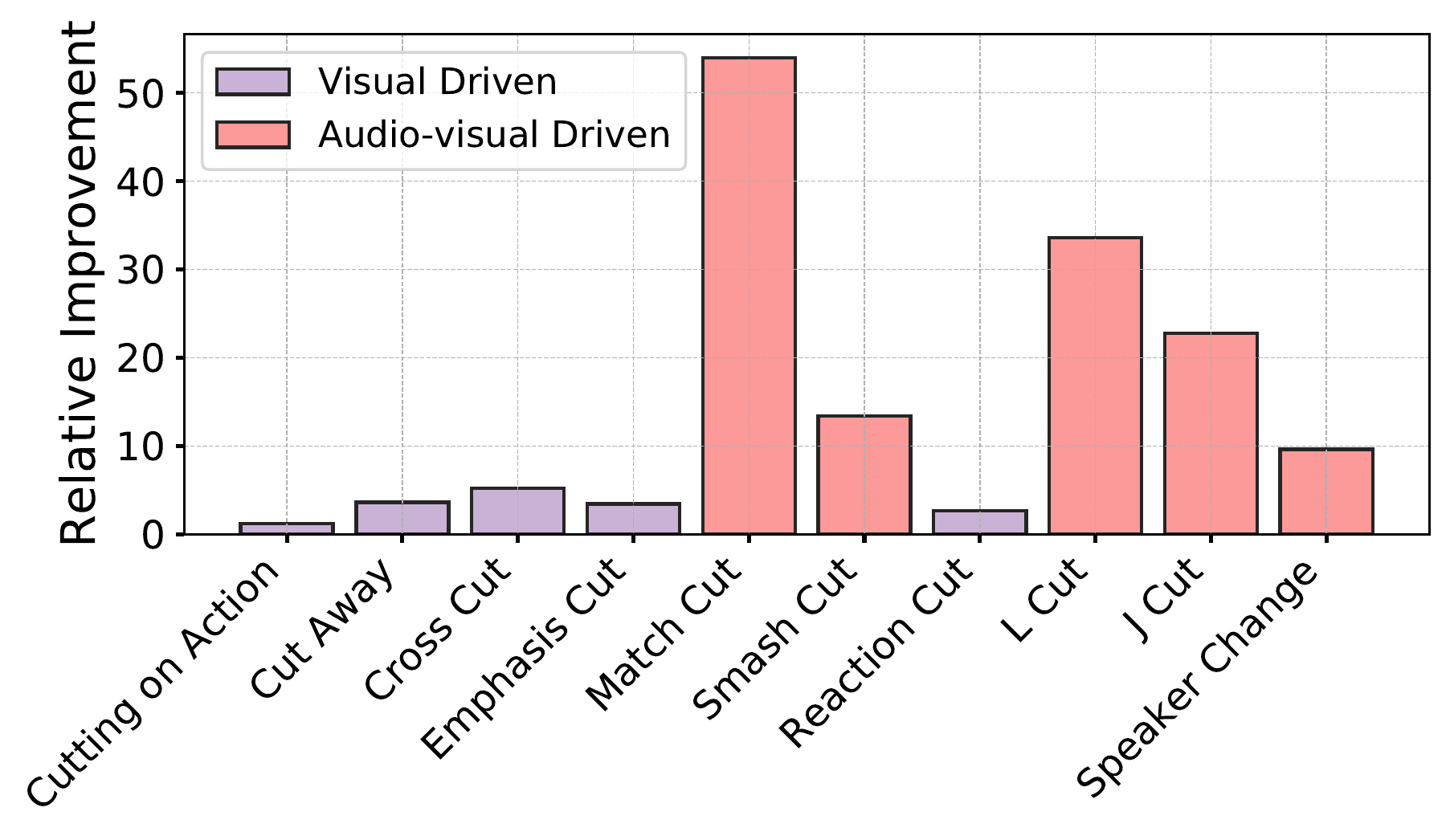}
        \caption{Audio-visual improvements per type.}
         \label{fig:performance-breakdown-b}
    \end{subfigure}%
    \caption{\textbf{Performance breakdown.} Here, we showcase a detailed performance analysis. Figure \ref{fig:performance-breakdown-a} shows the performance breakdown according to attributes of MovieCuts, such as type of sound, subjects, locations, duration per clip, and production year. Figure \ref{fig:performance-breakdown-b} shows the performance gain of the audio-visual model versus the visual-only model. For this analysis, we group the cut classes into visual-driven and audio-visual driven.} 
    \label{fig:performance-breakdown}
\end{figure*}

\noindent \textbf{Test set results}. After obtaining the best-performing model on the validation set, we evaluate this model on the test set. We obtain $47.70\%$ mAP, which is slightly lower than the results on the validation set $47.91\%$ mAP. For the full test set results, and experiments using Distribution-Balanced loss~\cite{wu2020distribution}, refer to the \textbf{supplementary material}.

\subsection{Performance Breakdown}

\noindent\textbf{Attributes and dataset characteristics.} Figure \ref{fig:performance-breakdown-a} summarizes the performance of our best audio-visual model from Table \ref{table:results} for different attributes and dataset characteristics. In most cases, the model exhibits robust performance across attributes. The largest performance gap is observed between Speech and Other sounds. We associate this result with the fact that cuts with complex audio editing, \eg those that include sound effects, often employ abstract editing such as Smash Cuts and Match Cuts, which are harder for the model to recognize. We also observe that the model is better at classifying cuts when there are faces in the scene, which aligns with the fact that it was trained on movies, which are mainly human-centered. 
These findings showcase the multi-modal nature of MovieCuts. Thus, the results can be improved by studying better audio-visual backbones.

\noindent\textbf{Audio-visual improvements per cut type.} Figure \ref{fig:performance-breakdown-b} shows the relative improvement of the audio-visual model \wrt using the visual stream only. It highlights whether the type of cut is driven by visual or audio-visual information. We observe that the audio-visual driven cuts generally benefit the most from training a joint audio-visual model. Match cuts show a relative $50\%$ improvement when adding audio. These types of cuts use audiovisual concepts to match the two shots. The second-largest gains are for cuts related to dialogue and conversations (L cut, J cut). For instance, L Cuts improve by $30\%$; this class typically involves a person on screen talking in the first shot while only their voice is heard in the second shot. By encoding audio-visual information, the model disambiguates predictions that would otherwise be confused by the visual-only model. Finally, all classes show a relative improvement \wrt the visual baseline. This suggests that the GB \cite{wang2020makes} strategy allows the model to optimize modality weights. In the worst-case scenario, GB achieves slight improvements over the visual-only baseline. In short, we empirically demonstrate the importance of modeling audio-visual information to recognize cut types.




\subsection{Machine-assisted Video Editing with MovieCuts}
\label{sec:application}

We argue that recognizing cut types can enable many applications in video editing. In this section, we leverage the knowledge of our Audio-Visual model to attempt automated video editing. Inspired by \cite{Pardo_2021_ICCV}, we use EditStock\footnote{\href{https://editstock.com/}{EditStock.com}} to gather raw footage of edited movies and perform video editing on them. Specifically, we use our model to create cuts (shot transitions) between two long sequences of shots. Further details can be found in \textbf{supplementary material}. We measure qualitatively and quantitatively our model's editing by comparing it with different automated editing methods:
(1) \textbf{Random baseline frame by frame RF:} Every frame, we perform a cut with a probability of 0.5.
(2) \textbf{Random baseline snippet by snippet RS:} Similar to how the model is trained, every 16 frames snippet we cut with a probability of 0.5. This restriction allows each shot to be on screen for at least 16 frames.
(3) \textbf{Biased Random BR:} Similar to (2), we cut every 16 snippets. However, this time we use the expected number of cuts prior. Thus, we ensure that the number of random cuts is the same as the ground truth. 
(4) \textbf{MovieCuts' AV model AV:} We use our audio-visual model's scores to score all possible cuts between the two sequences. Then, we use the top-k cuts, where k is given by the expected number of cuts. 
(5) \textbf{Human Editor GT:} From EditStock we collect the actual edited sequence edited by professional editors. We use these sequences as a reference for the quantitative study, and ground-truth for the qualitative evaluation.

In the qualitative evaluation we ask 63 humans to pick between our method and all the other methods. We observe that users picked our method (AV) over the human editor $38\%$ of the times while the BR was picked $34\%$ of the times -- RF and RS were picked only $15.7\%$ and $1.8\%$, respectively. Furthermore, for the quantitative results we use the human edit as ground-truth and measure Purity, Coverage, and F1 for each method. These metrics were implemented by  \cite{pyannote.metrics} and measure the similarity between the segmentation of two different sequences. The results are consistent with the qualitative study and show that MovieCuts' edits have an F1 of $81\%$ while BR, RS, and RF have only $77\%$, $63\%$ $17\%$, respectively. A more in-depth analysis of this study can be found in \textbf{supplementary material}. This simple experiment suggests that the MovieCuts dataset allows the model to learn about video editing by learning cut-type recognition. Thus, we argue that further improvement in Cut-type recognition tasks can translate into advances in tasks related to machine-assisted video editing.

\vspace{-10pt}
\section{Conclusion}\label{sec:conclusion}
We introduced the cut-type recognition task in movies and started research in this new area by providing a new large-scale dataset, MovieCuts accompanied with a benchmark of multiple audio-visual baselines.. We collect $173,967$ annotations from qualified human workers. We analyze the dataset diversity and uniqueness by studying its properties and audio-visual attributes. We propose audio-visual baselines by using learning approaches that address the multi-modal nature of the problem. Although we established a strong research departure point, we hope that more research pushes the envelope of cut-type recognition by leveraging MovieCuts.\\

\noindent\textbf{Acknowledgements.} This work was supported by the King Abdullah University of Science and Technology (KAUST) Office of Sponsored Research through the Visual Computing Center (VCC) funding.


\clearpage

%
%
\bibliographystyle{splncs04}
\bibliography{egbib}
\clearpage

\clearpage
\appendix
\section*{Supplementary Material}



\section{MovieCuts Attributes Details}
We leverage CLIP \cite{radford2021learning} to extract visual attributes from all instances in MovieCuts. We follow the zero-shot setup described in \cite{radford2021learning}. To do so, we create language queries using relevant classes, for each attribute type, as a set of candidate text-visual pairs and use CLIP's dual encoder to predict the most probable pair (the most probable tag). Thus, we compute an image embedding for the visual frames, and a text embedding for all candidate text queries (attribute tags) to then compute the cosine similarity between the L2-normalized embedding pairs. Instead of simply passing the tags to the language encoder, we augment the text queries using the following template: ``a photo of a \emph{subject attribute}'', and ``an \emph{location attribute} photo'' for the subject and location attributes, respectively. We retrieve tags for each of its shots by sampling a random frame before and after the shot transition from each cut.

\begin{figure}[t!]
    \begin{center}
        \includegraphics[width=0.9\linewidth]{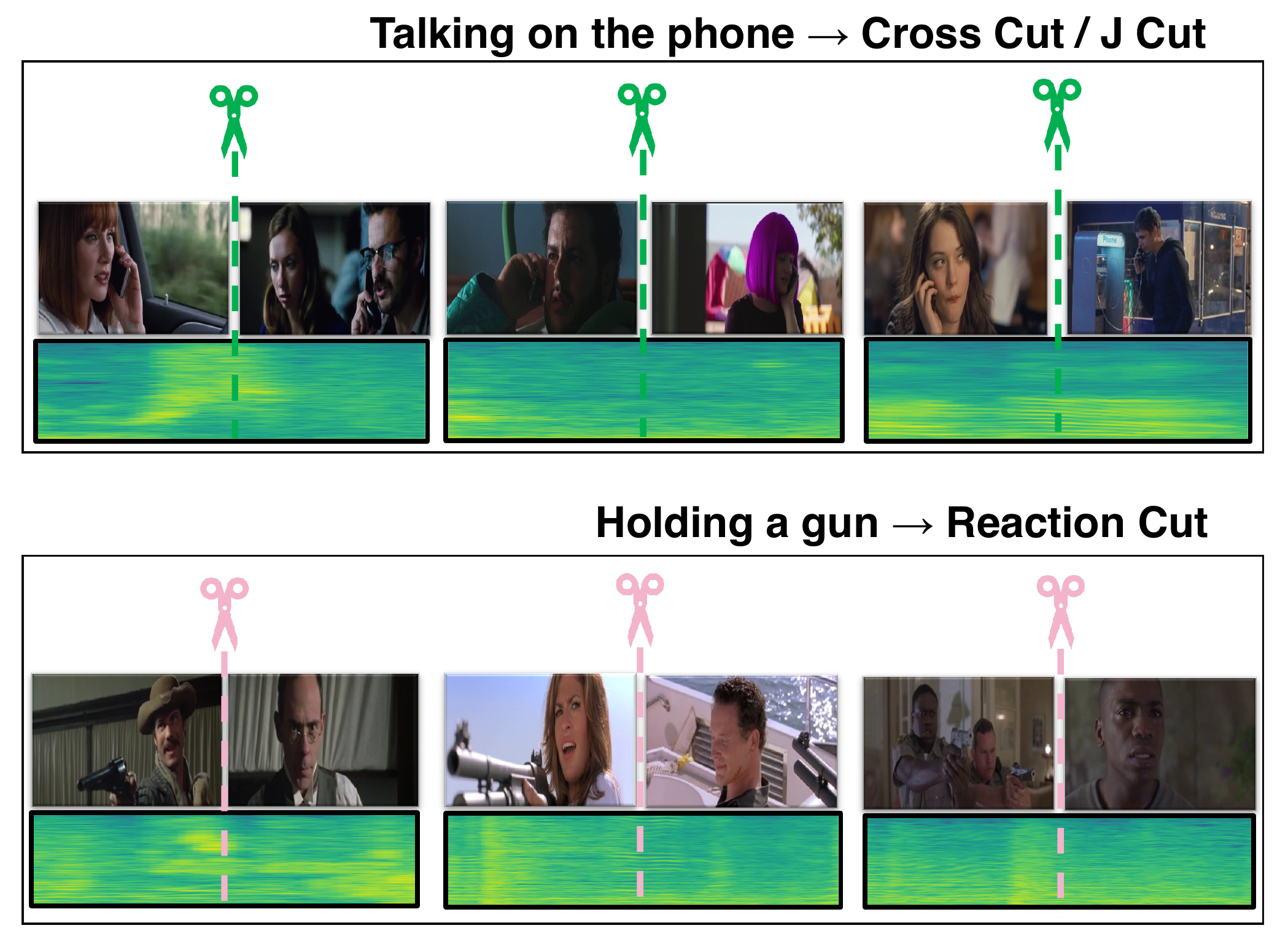}
    \end{center}
    \caption{\textbf{Actions that trigger cuts.} Some actions predominantly co-occur with a particular cut type. For instance, a Talking on the phone action is often edited via the Cross Cut and J Cuts. Another common pattern emerges when someone is Holding a gun. The predominant edit is the Reaction Cut, which first shows an actor holding the gun, and the next shot highlights a face reaction of another subject.}
    \label{fig:dataset-action-cuts}
\end{figure}

\noindent\textbf{Actions that trigger Cuts.} Our goal is to find correlations between action tags and cut types. To do so, we first build a zero-shot action classifier based on CLIP \cite{radford2021learning}. Since the zero-shot action classifier did not offer us a high accuracy, we limit our analysis with the most confident tags only. Such tags allow us to find the most common co-occurrences between actions and cut-types. Figure \ref{fig:dataset-action-cuts} showcases three common action/cut pairs. These patterns are common across different movie scenes and editors' styles. These empirical findings reaffirm the theory of the film grammar \cite{filmgrammar,smith2012attentional}, which suggests that video editing follows a set of rules more often than not. \label{section:actiontrigger}

\section{Additional Results}

\begin{table*}[ht!]
    \small
	\centering
    \renewcommand{\arraystretch}{1.1}

	\begin{tabular}{ c | c | c | c c c c c c c c c c}
		Model  & Sampling  & \textbf{mAP} & \textbf{CA} & \textbf{CW} & \textbf{CC} & \textbf{EC}  & \textbf{MC} & \textbf{SC} & \textbf{RC} & \textbf{LC} & \textbf{JC} & \textbf{SC} \\
		\hline
		\hline
		AV+Scaled GB & Uniform & 47.2&	64.3&	62.7&	32.3&	31.2&	2.6&	23.5&	83.0&	44.5&	51.6&	76.3 \\
		\hline
		AV+Scaled GB & Gaussian & 47.4&	64.0&	62.7&	34.4&	\textbf{32.0}&	2.1&	23.8&	82.7&	43.7&	\textbf{51.9}&	77.0 \\
		\hline
		AV+Scaled GB  & Fixed & \textbf{47.9}&	\textbf{65.6}&	\textbf{63.0}&	\textbf{34.9}&	31.8&	\textbf{2.3}&	\textbf{24.4}&	\textbf{83.3}&	\textbf{45.0}&	51.6&	\textbf{77.1} \\
		\hline
	\end{tabular}
	\caption{\textbf{Window Sampling Results}. Different window sampling strategies, using the Audio-Visual + Scaled Gradient Blending Model \cite{wang2020makes}. All the reported numbers are $\%$ AP. Showing classes: Cutting on Action (\textbf{CA}), Cut Away (\textbf{CW}), Cross Cut (\textbf{CC}), Emphasis Cut (\textbf{EC}), Match Cut (\textbf{MC}), Smash Cut (\textbf{SC}), Reaction Cut (\textbf{RC}), L Cut (\textbf{LC}), J Cut (\textbf{JC}), Speaker Chance (\textbf{SC}).
	}
	\label{table:window_sampling_complete}
\end{table*}
\begin{table*}[ht!]
    \small
	\centering
    \renewcommand{\arraystretch}{1.1}

	\begin{tabular}{ l | c | c c c c c c c c c c}
		Model  &  \textbf{mAP} & \textbf{CA} & \textbf{CW} & \textbf{CC} & \textbf{EC}  & \textbf{MC} & \textbf{SC} & \textbf{RC} & \textbf{LC} & \textbf{JC} & \textbf{SC} \\
		\hline
		\hline
		AV Scaled & 47.4&	65.1&	62.7&	33.1&	31.4&	1.8&	23.0&	83.0&	45.4&	50.7&	77.6 \\
		\hline
		AV Scaled+DB Loss& 47.8&	65.5&	63.0&	35.0&	31.7&	1.9&	23.4&	83.1&	45.7&	50.8&	77.7 \\
		\hline
		AV+Scaled GB+DB Loss & \textbf{47.9}&	\textbf{65.7}&	63.7&	\textbf{34.8}&	\textbf{31.5}&	1.9&	\textbf{24.0}&	\textbf{83.2}&	\textbf{45.0}&	\textbf{51.3}&	\textbf{77.4} \\
		\hline
		AV+Scaled GB & \textbf{47.9}&	\textbf{65.6}&	\textbf{63.0}&	\textbf{34.9}&	\textbf{31.8}&	\textbf{2.3}&	\textbf{24.4}&	\textbf{83.3}&	\textbf{45.0}&	\textbf{51.6}&	\textbf{77.1} \\
		\hline
	\end{tabular}
	\caption{\textbf{DB Loss Results}. We show the performance of different experiments using DB Loss on the validation set with Fixed sampling. We use Audio-Visual Model combined with Gradient Blending \cite{wang2020makes} and DB Loss \cite{wu2020distribution}. The reported number is $\%$ mAP. Showing classes: Cutting on Action (\textbf{CA}), Cut Away (\textbf{CW}), Cross Cut (\textbf{CC}), Emphasis Cut (\textbf{EC}), Match Cut (\textbf{MC}), Smash Cut (\textbf{SC}), Reaction Cut (\textbf{RC}), L Cut (\textbf{LC}), J Cut (\textbf{JC}), Speaker Chance (\textbf{SC}).
	}
	\label{table:dbloss}
\end{table*}

\noindent\textbf{Distribution-Balanced Loss Experiments.} We experiment with the Distributed-Balanced Loss (DB Loss)~\cite{wu2020distribution} introduced by Wu~\etal. The DB Loss was designed to tackle datasets with multiple labels per sample that follow a long-tail distribution. It proposes a modification to the standard binary cross-entropy loss by adding two terms to handle multiple labels and long-tail distributions.  For further details, including the loss formulation, please refer to the original publication [53]. 
For the experiments shown in Table \ref{table:dbloss} we upgrade our base BCE loss by the DB Loss. For a fair comparison, we scale the original naive combination by $3.0$ (to put the losses' magnitudes around the same scale). We observe that the DB Loss helps the base model and improves the mAP on most of the classes. However, when combined with the Scaled GB weights, there is no significance difference between using the standard BCE Loss and the DB Loss. 

\noindent\textbf{Window Sampling.} We show the results on each one of the classes for the Window Sampling study in Table \ref{table:window_sampling_complete}.

\noindent\textbf{Test Set Results.} Finally, Table \ref{table:results_test_complete} presents the results of the best performing model (AV + Scaled GB) on the test set.
\begin{table*}[ht!]
    \small
	\centering

	\begin{tabular}{ c | c | c | c c c c c c c c c c}
		Model & Sampling &  \textbf{mAP} & \textbf{CA} & \textbf{CW} & \textbf{CC} & \textbf{EC}  & \textbf{MC} & \textbf{SC} & \textbf{RC} & \textbf{LC} & \textbf{JC} & \textbf{SC} \\
		\hline
		\hline
		AV + Scaled GB & Gaussian & 47.7&	66.0&	63.0&	32.2&	32.6&	2.8&	26.5&	82.8&	43.5&	50.7&	76.7 \\
		\hline
	\end{tabular}
	\caption{\textbf{Test Set Results}. We show the performance of the fine-tuned models from table 1 of the main paper, evaluated on the test set. The reported number is $\%$ mAP. 
	}
	\label{table:results_test_complete}
\end{table*}

\begin{figure}
     \centering
     \begin{subfigure}[b]{0.48\textwidth}
         \centering
         \includegraphics[width=\textwidth] {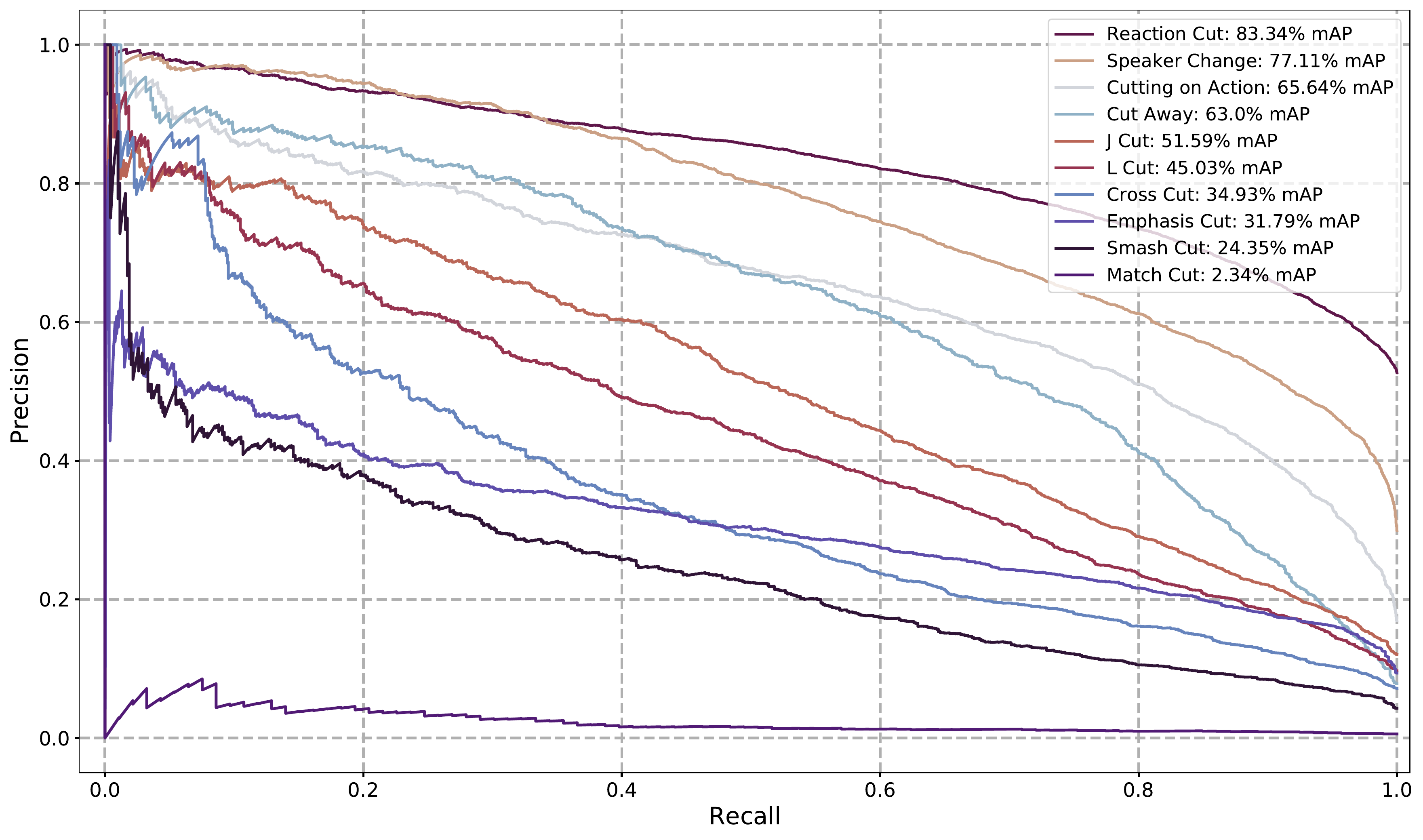}
         \caption{\textbf{Precision vs Recall.} AV + Scaled GB on validation set.}
         \label{figure:pr}
     \end{subfigure}
     \hfill
     \begin{subfigure}[b]{0.48\textwidth}
         \centering
         \includegraphics[width=\textwidth] {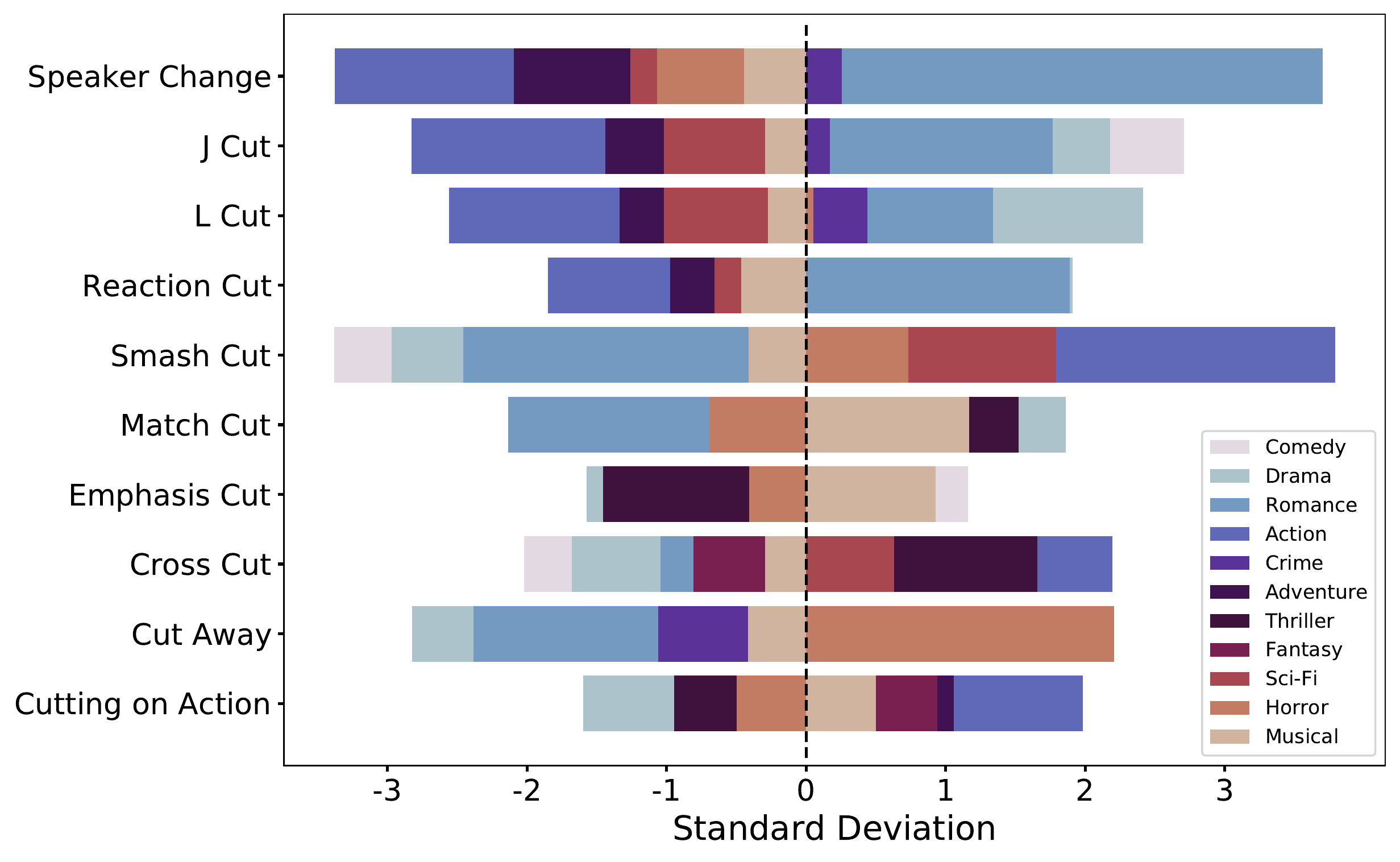}
         \caption{\textbf{Cut-types' distribution per genre.}}
         \label{figure:summary-counts}
     \end{subfigure}
     \caption{\textbf{Figure \ref{figure:pr}} shows the Precision vs Recall Curves for every class in the dataset. \textbf{Figure \ref{figure:summary-counts}} shows the summary of the cut-tupe's distribution per genre.}
\end{figure}

\noindent\textbf{Precision-Recall Curves.} Besides, in Figure \ref{figure:pr} we showcase the Precision-Recall (PR) curves for our best model on the validation set as an additional metrics to the ones shown in the main manuscript. We observe the Precision and Recall values for different confident thresholds for each one of the classes in MovieCuts.

\section{Additional Statistics}

Figure \ref{figure:summary-counts} summarizes the difference in labels' distribution across genres. To do this visualization we first calculate the average numbers of cuts for each of the classes, then, we plot the standard deviation from the classes' mean for each of the genres. Thus, we visualize how frequent or infrequent is each of the classes depending on the movie genre. For instance, we observe that for genres like Romance, and Drama the classes Speaker Change, J-cuts and L-cuts are more frequent as compared to Action, and Adventure movies. However, for Action, and Adventure, Cross-cuts and Cuts on Action are more frequent.

\begin{figure*}
\centering
\begin{subfigure}[t]{0.3\textwidth}
    \makebox[0pt][r]{\makebox[30pt]{\raisebox{50pt}{\rotatebox[origin=c]{90}{$Comedy$}}}}%
    \includegraphics[width=0.95\textwidth]
    {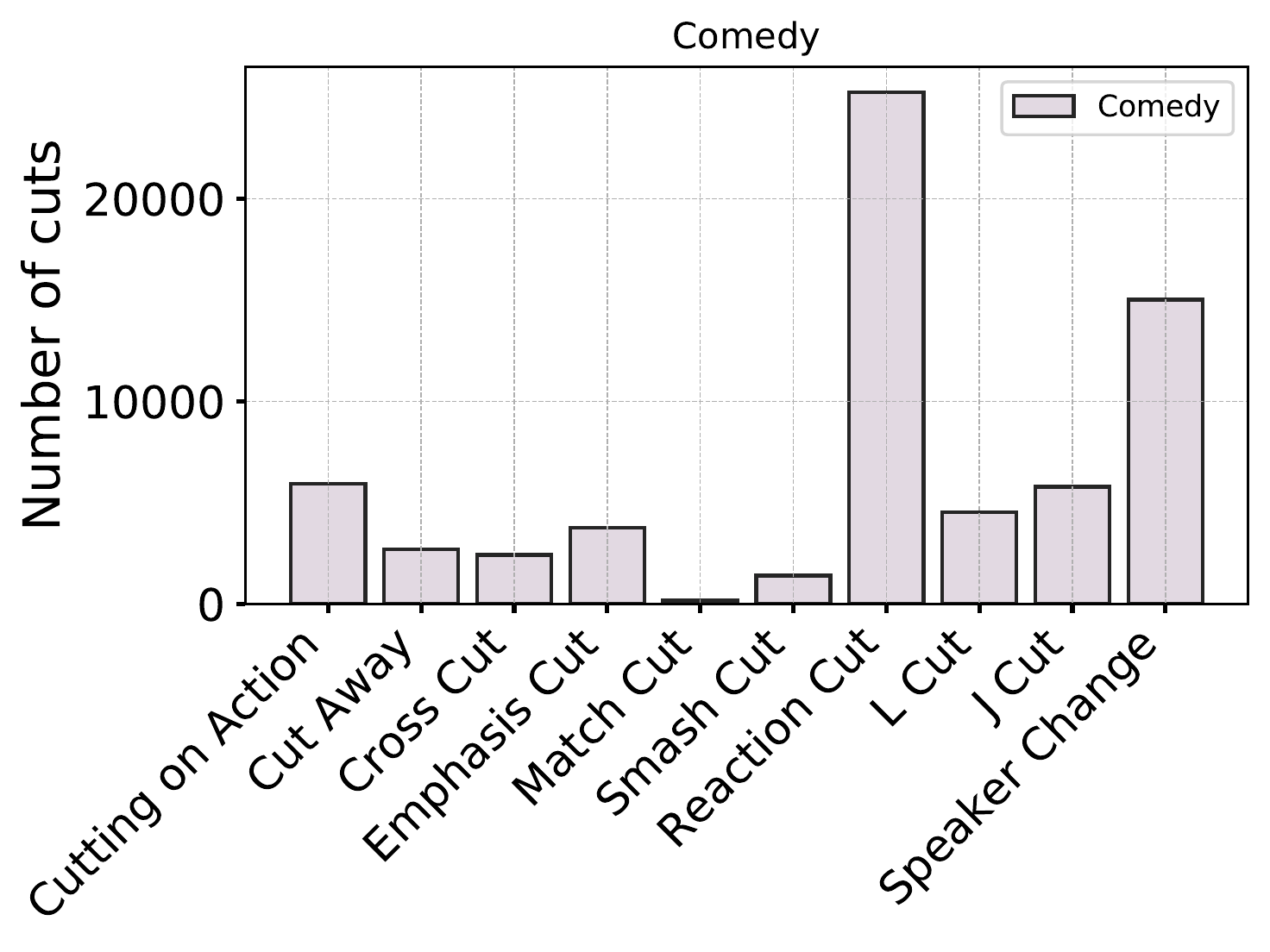}
    \makebox[0pt][r]{\makebox[30pt]{\raisebox{50pt}{\rotatebox[origin=c]{90}{$Drama$}}}}%
    \includegraphics[width=0.95\textwidth]
    {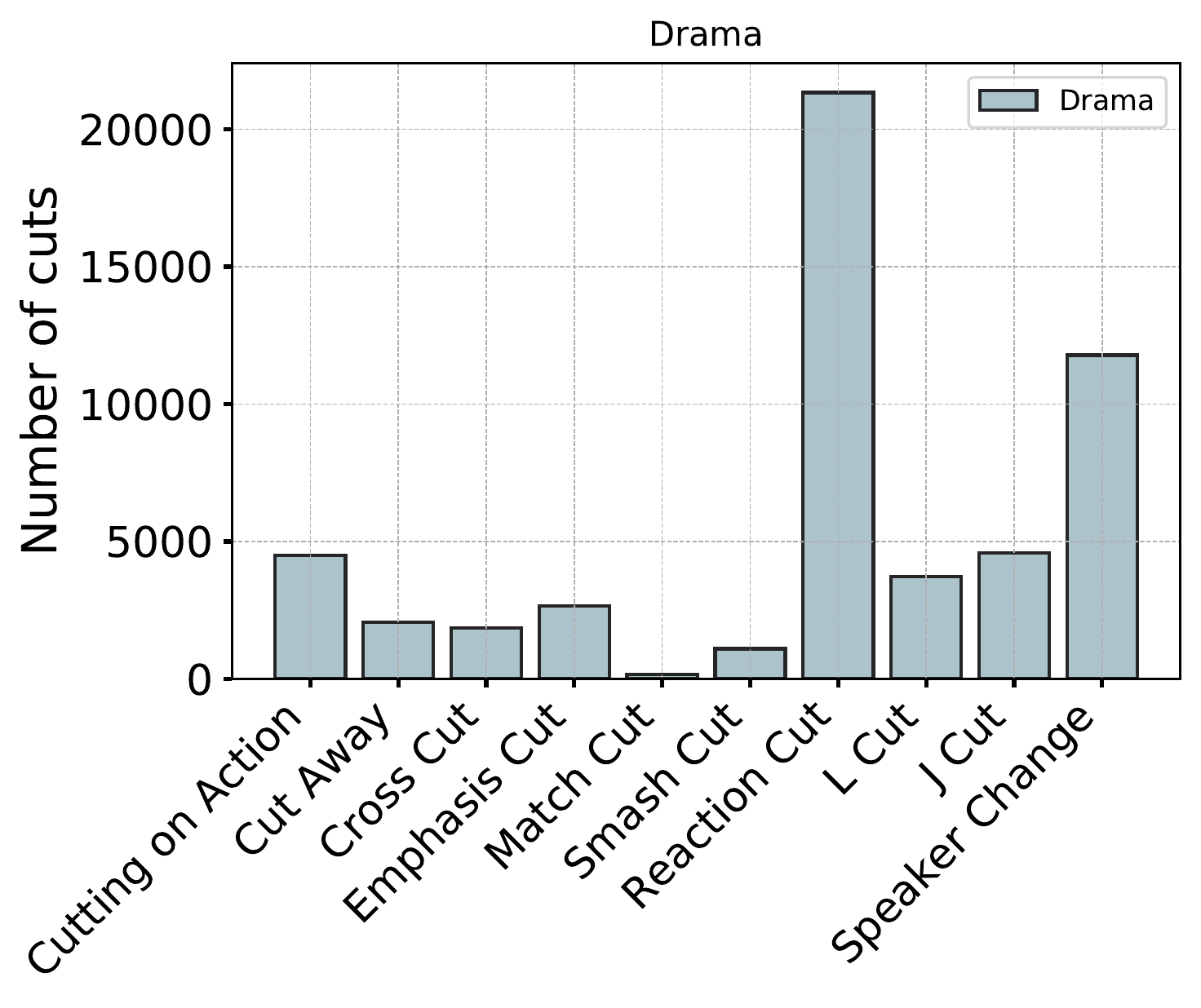}
    \makebox[0pt][r]{\makebox[30pt]{\raisebox{50pt}{\rotatebox[origin=c]{90}{$Romance$}}}}%
    \includegraphics[width=0.95\textwidth]
    {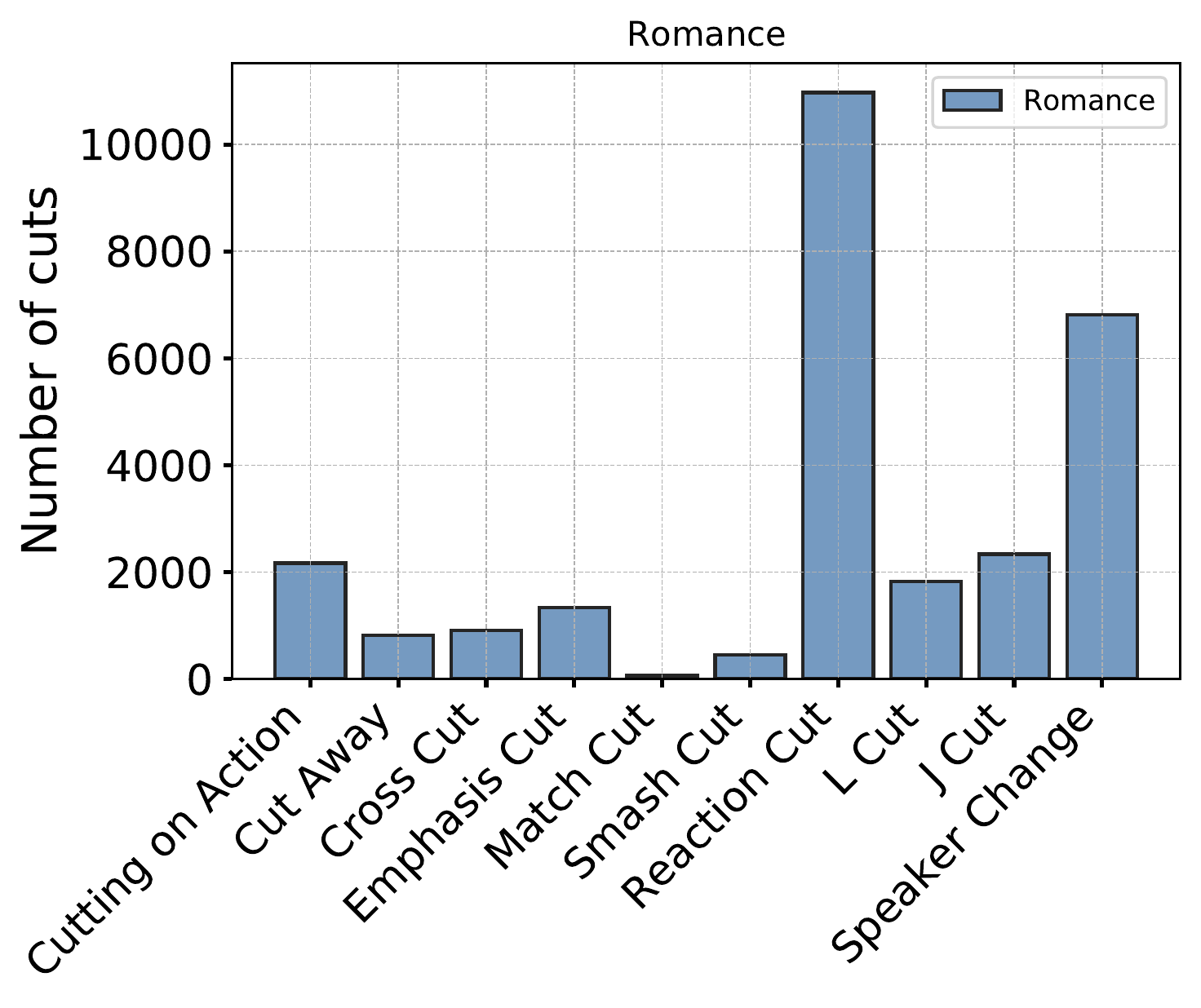}
    \makebox[0pt][r]{\makebox[30pt]{\raisebox{50pt}{\rotatebox[origin=c]{90}{$Action$}}}}%
    \includegraphics[width=0.95\textwidth]
    {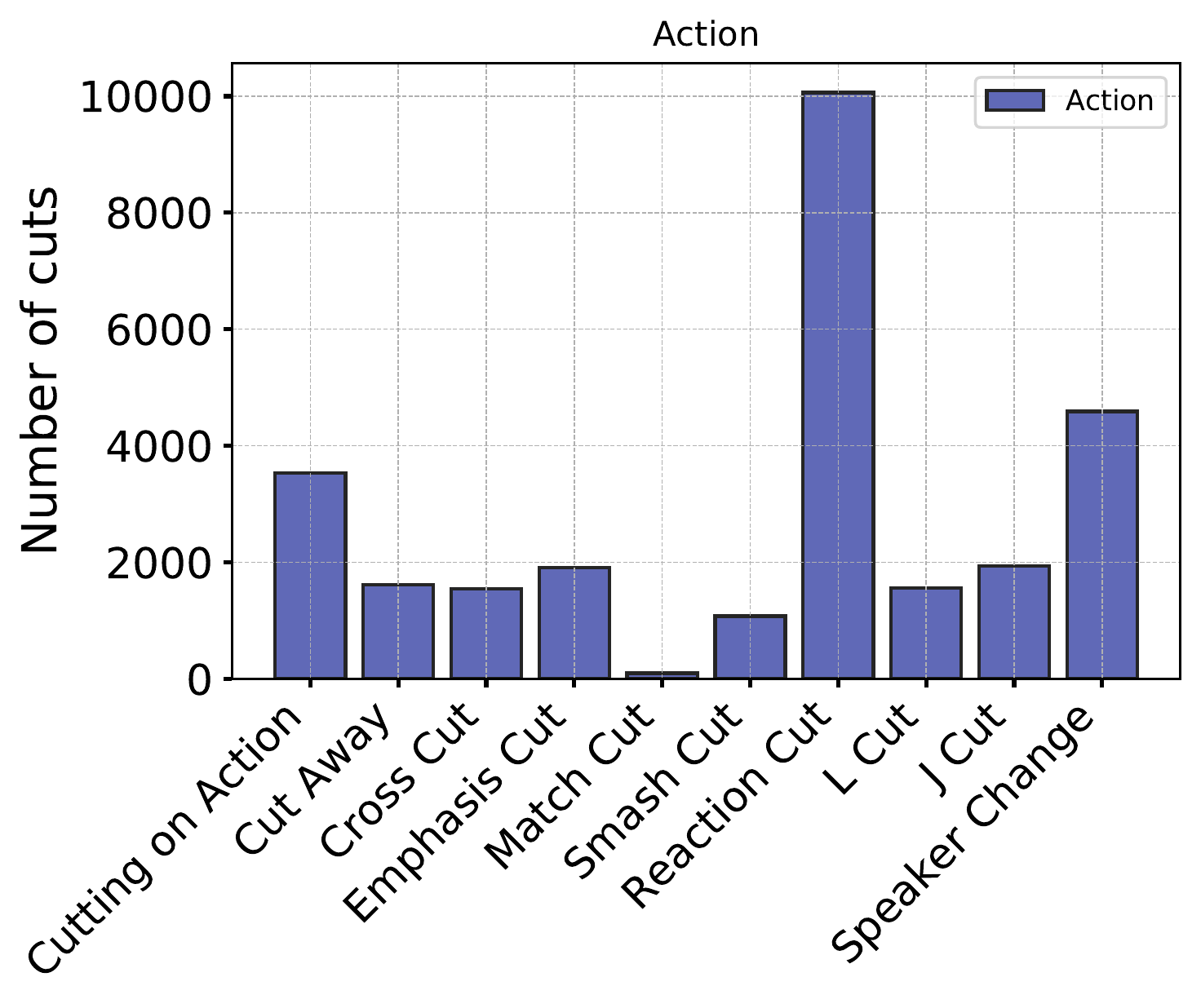}
    \makebox[0pt][r]{\makebox[30pt]{\raisebox{50pt}{\rotatebox[origin=c]{90}{$Thriller$}}}}%
    \includegraphics[width=0.95\textwidth]
    {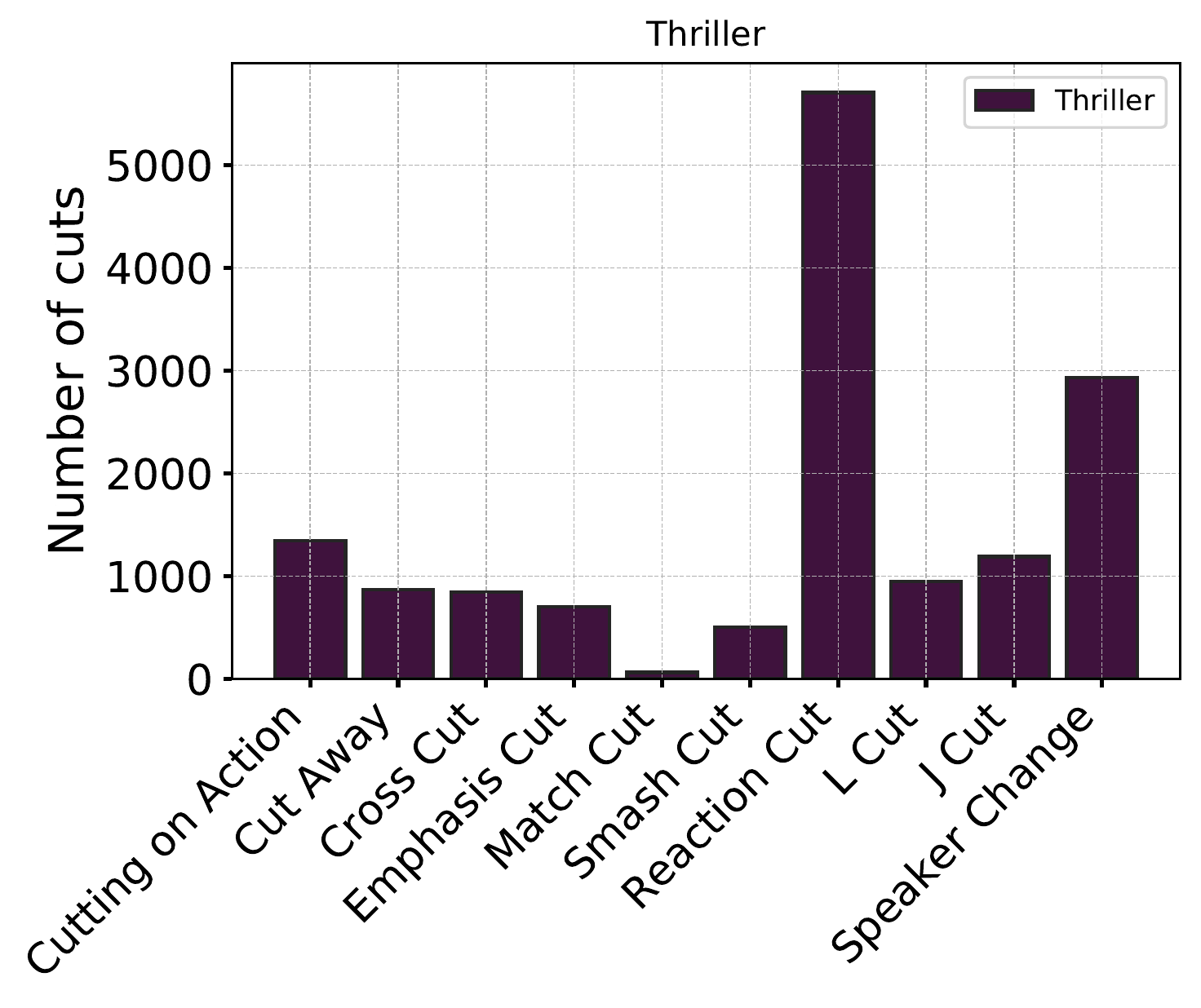}
    \caption{Train}
    \label{figure:label-distribution-train}
\end{subfigure}
\hfill
\begin{subfigure}[t]{0.3\textwidth}
    \includegraphics[width=0.95\textwidth]
    {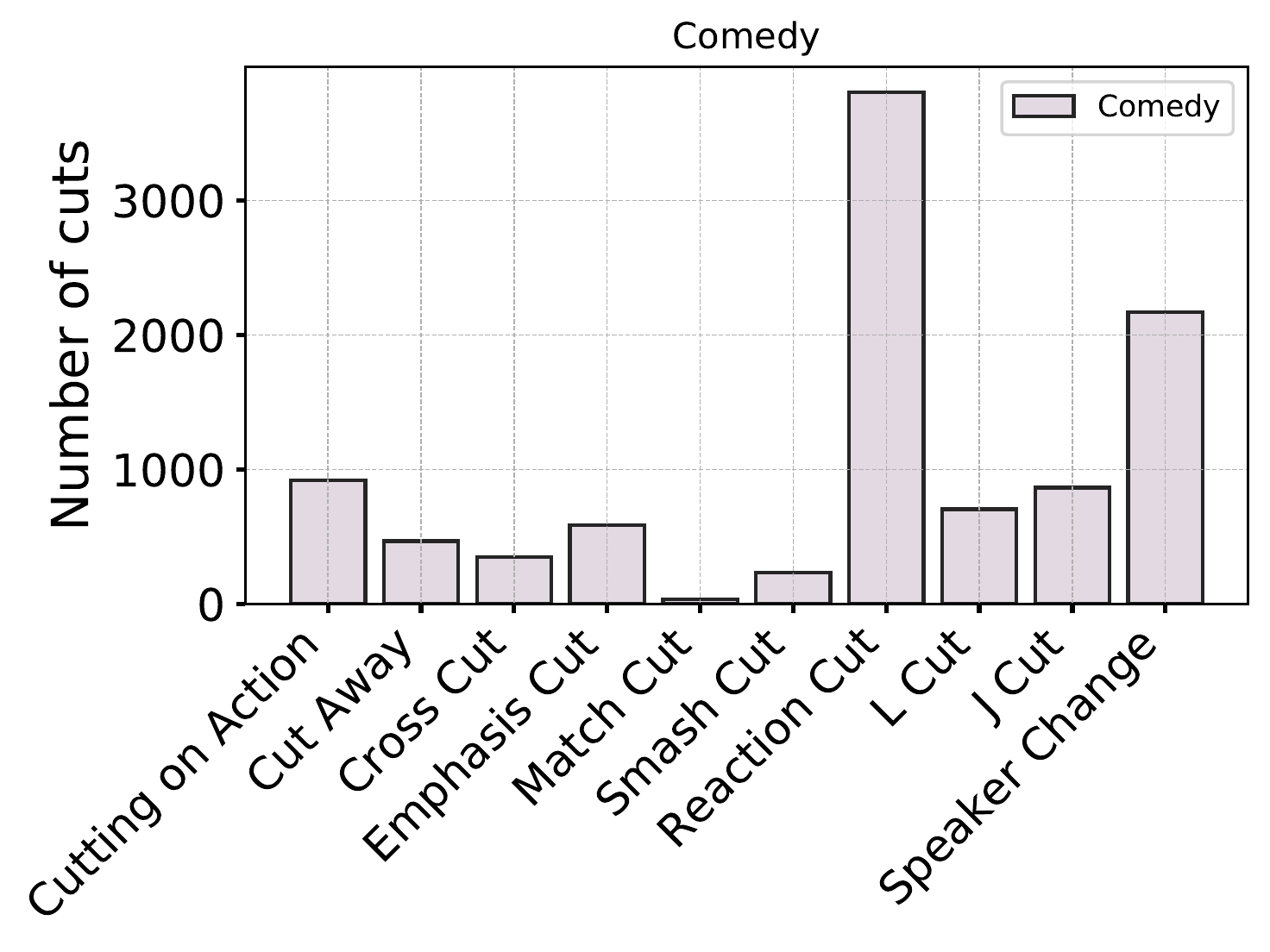}
    \includegraphics[width=0.95\textwidth]
    {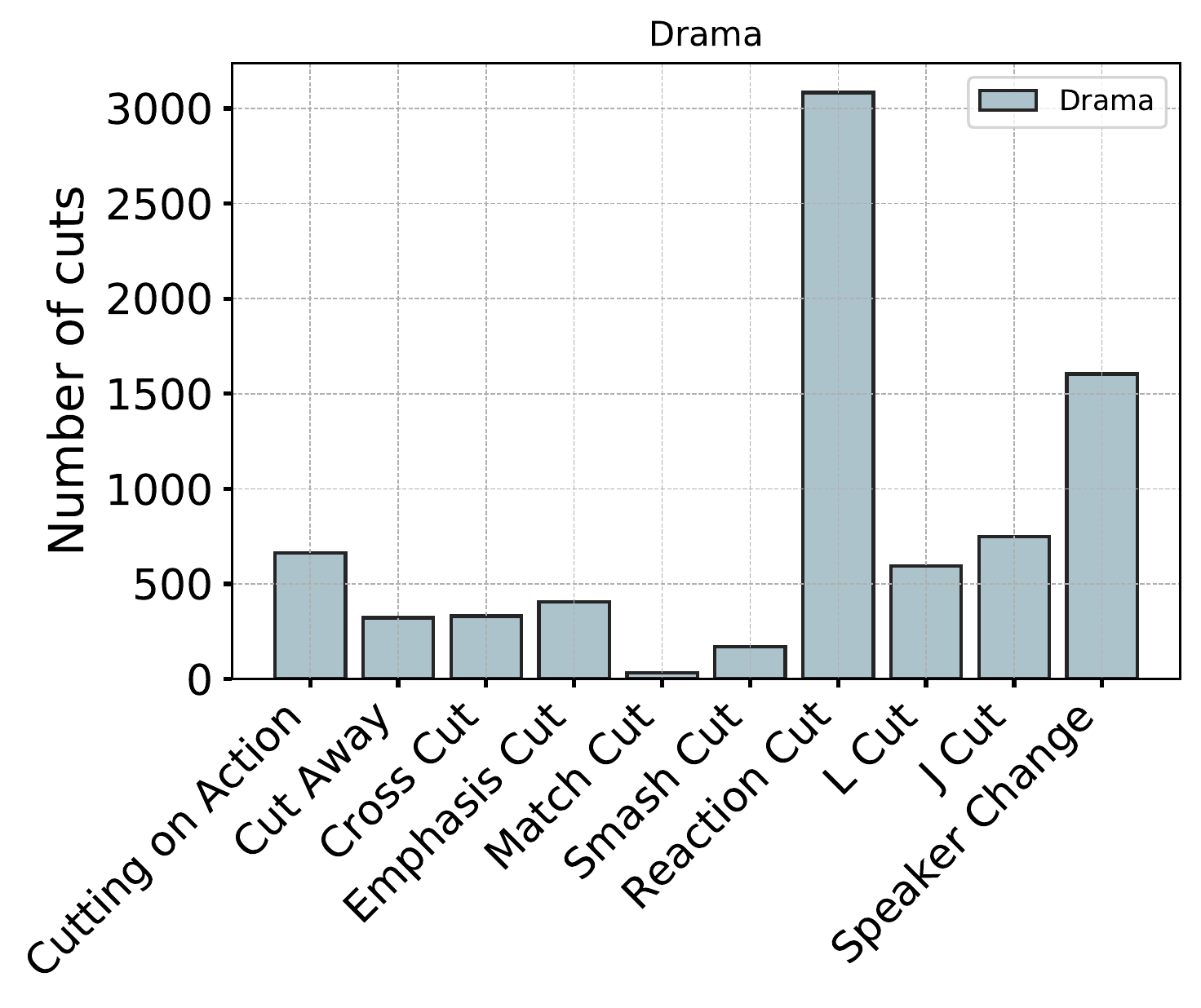}
    \includegraphics[width=0.95\textwidth]
    {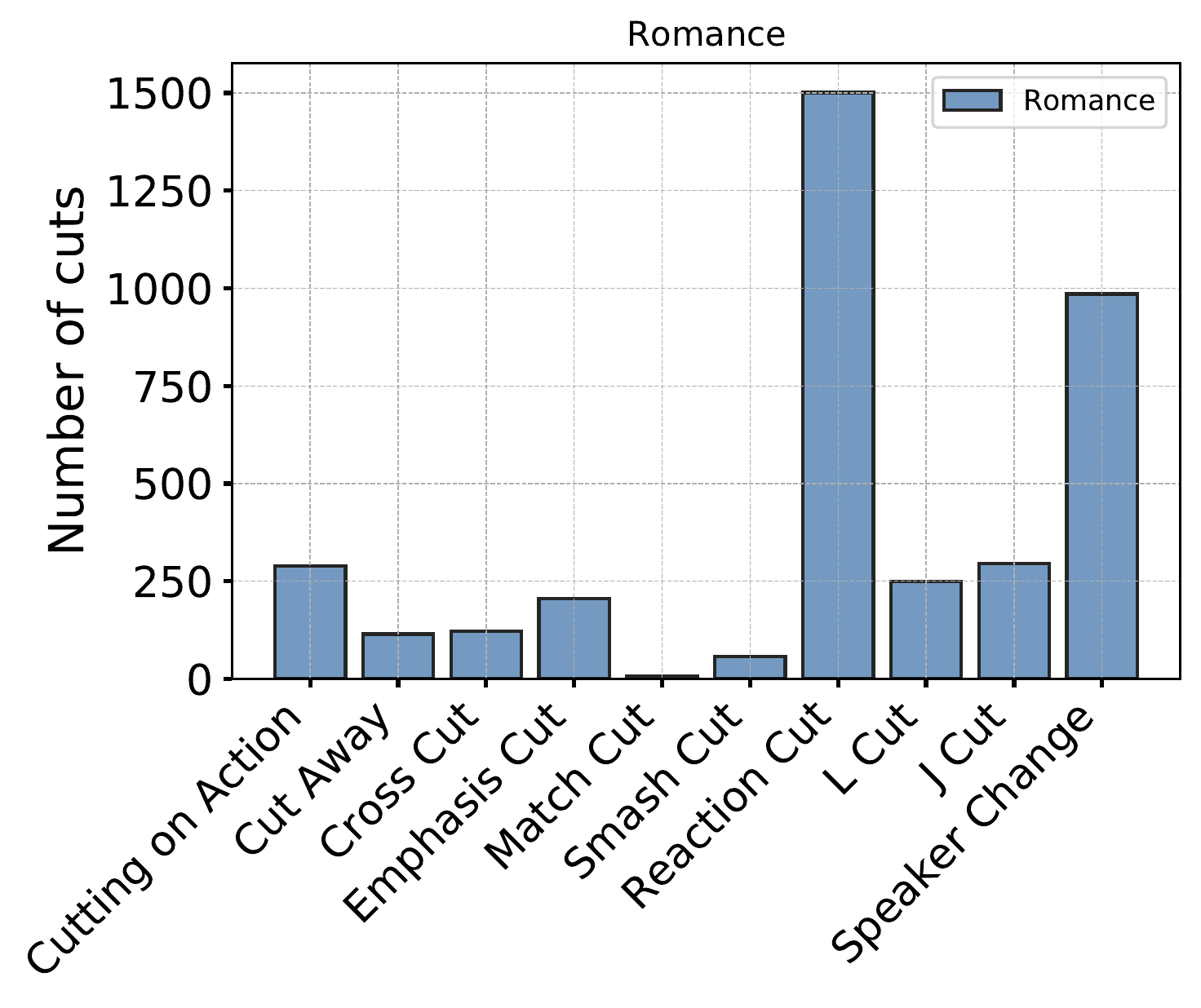}
    \includegraphics[width=0.95\textwidth]
    {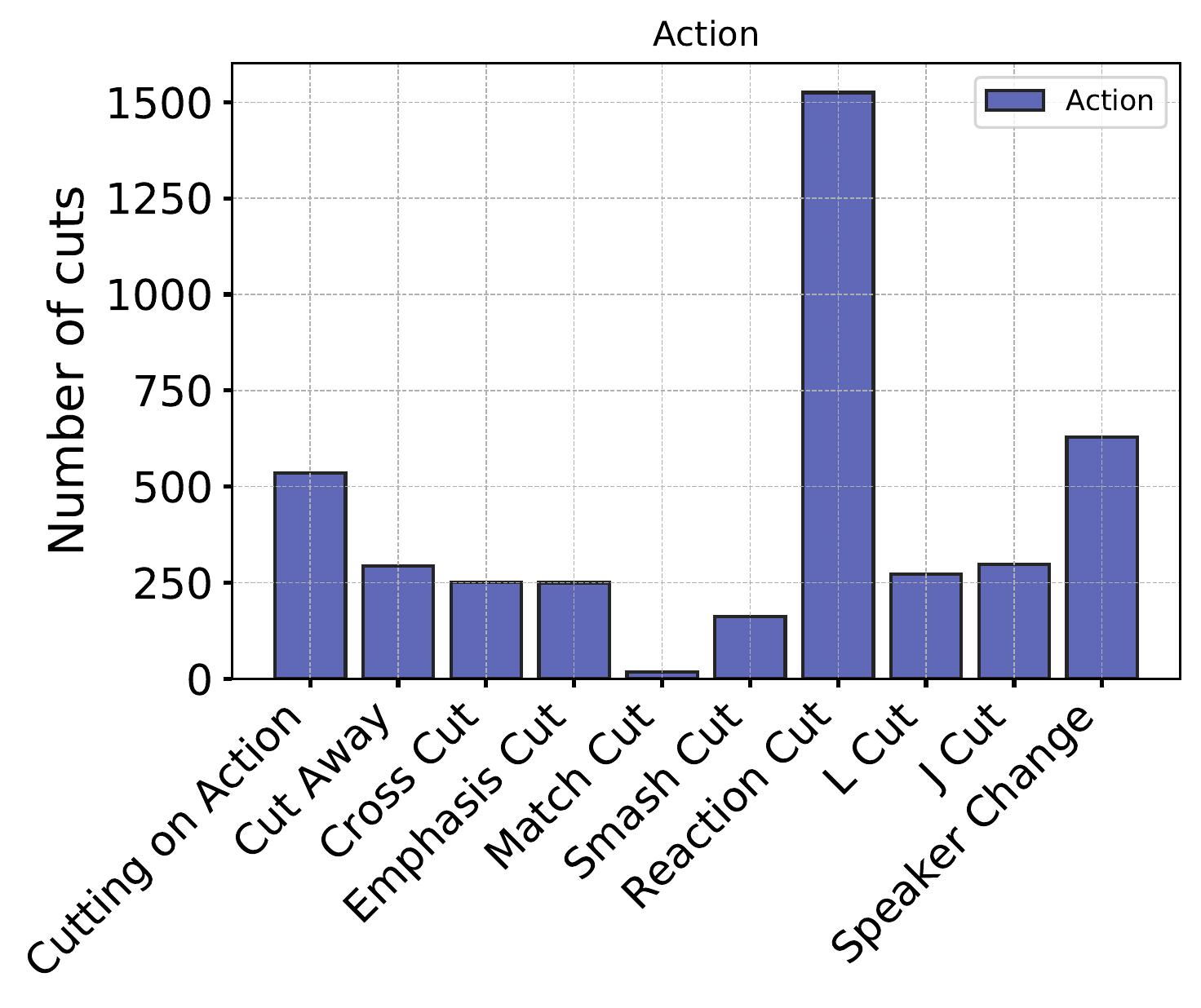}
    \includegraphics[width=0.95\textwidth]
    {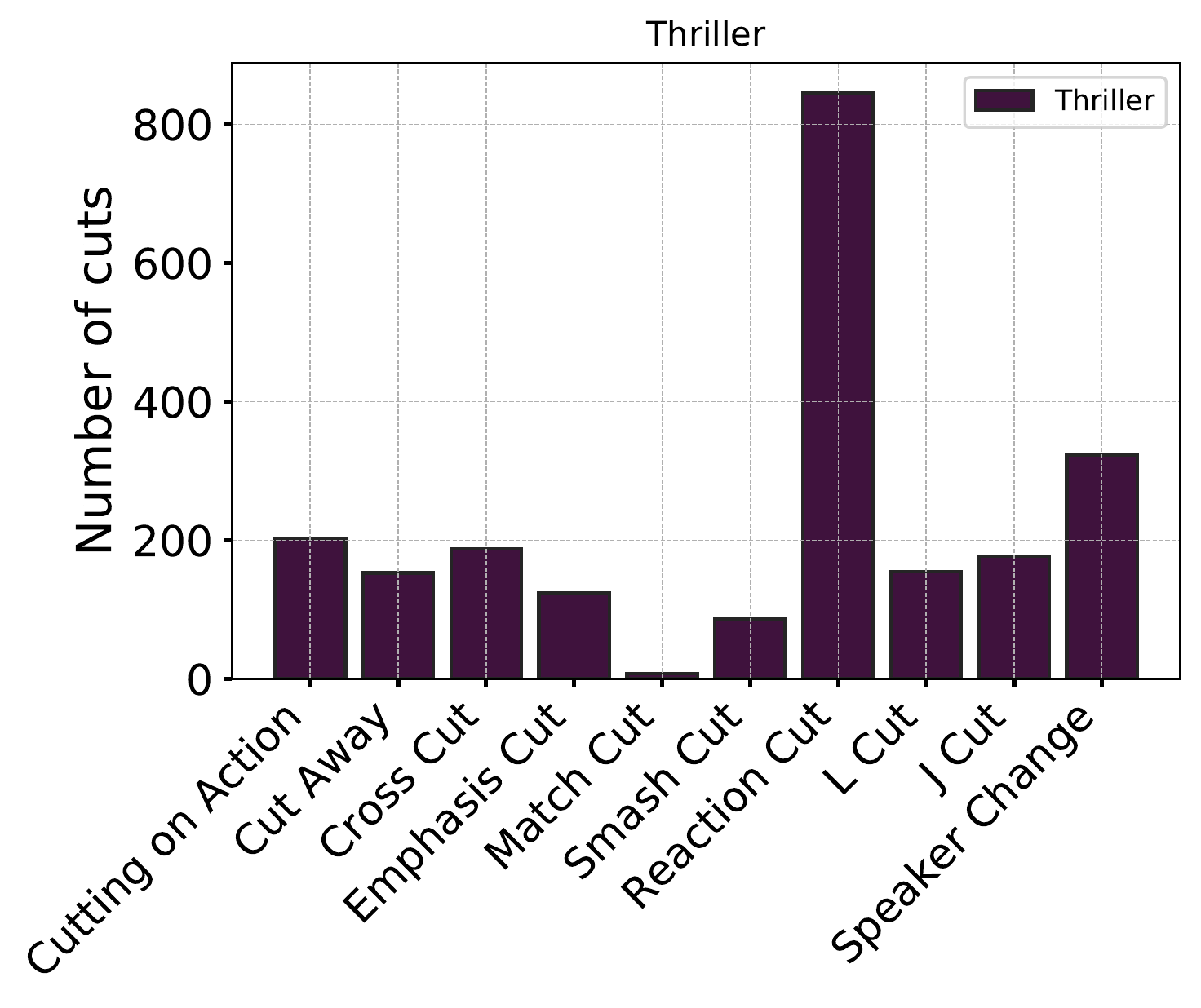}
    \caption{Validation}
    \label{figure:label-distribution-val}
\end{subfigure}
\hfill
\begin{subfigure}[t]{0.3\textwidth}
    \includegraphics[width=0.95\textwidth]
    {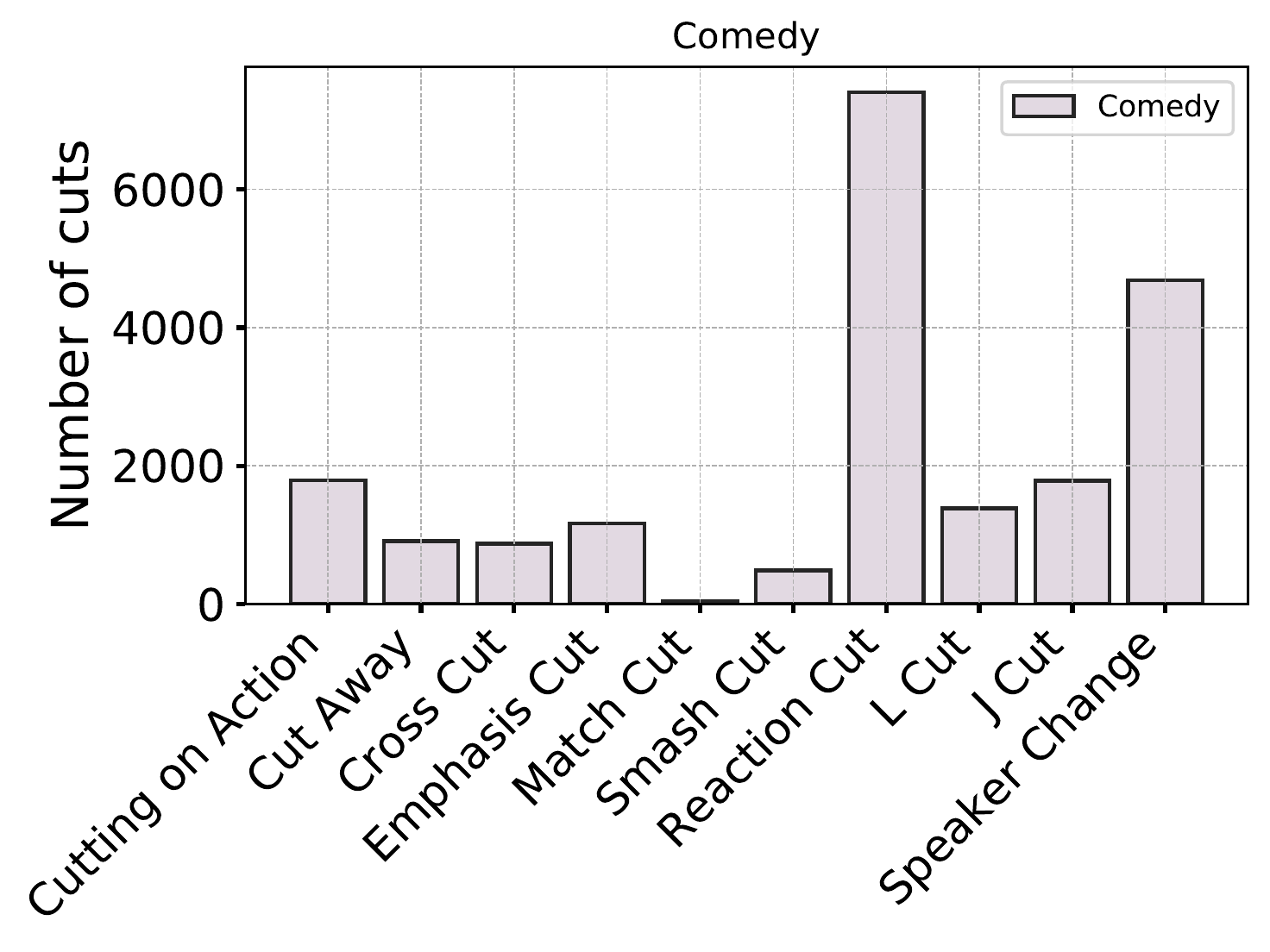}
    \includegraphics[width=0.95\textwidth]
    {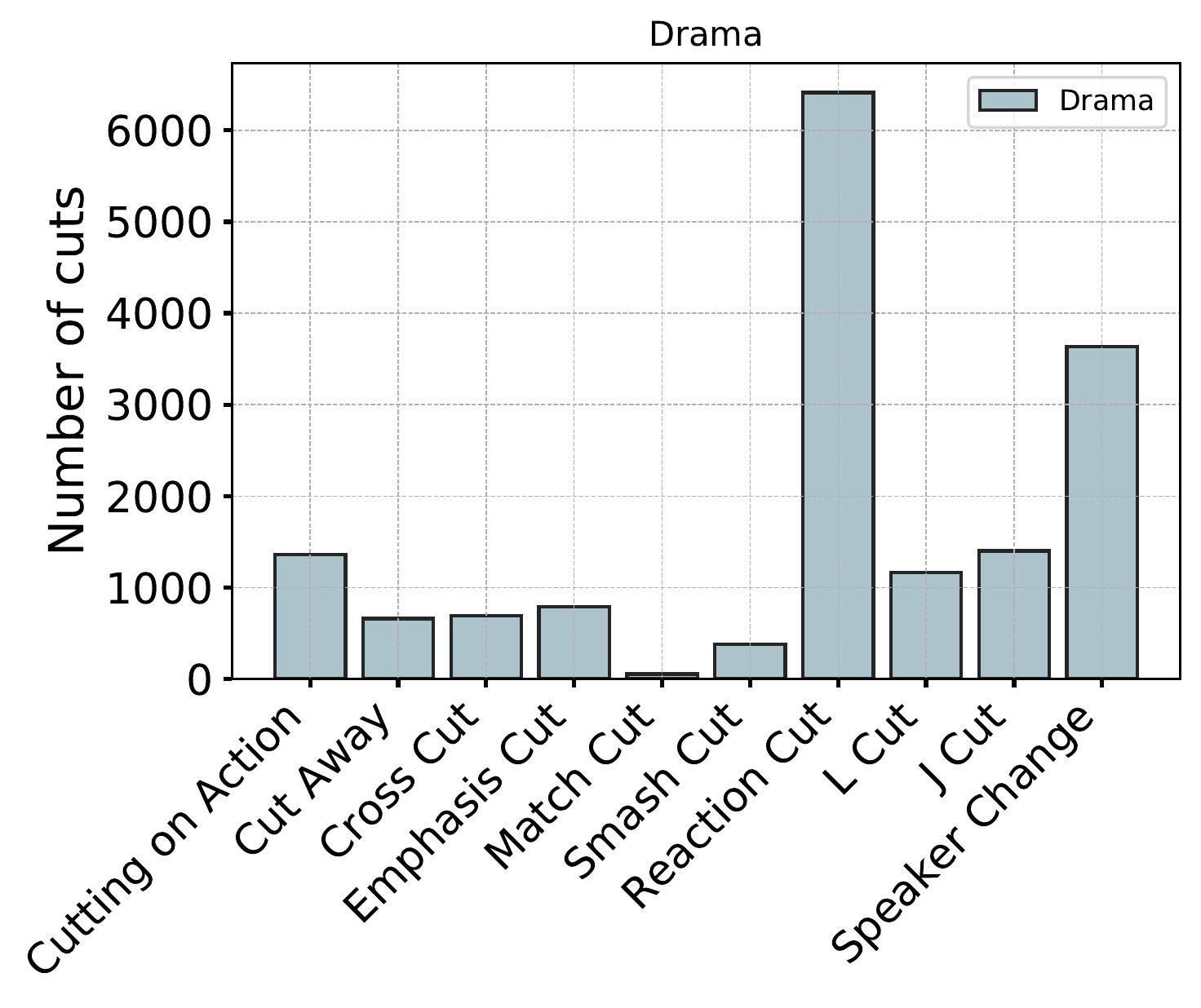}
    \includegraphics[width=0.95\textwidth]
    {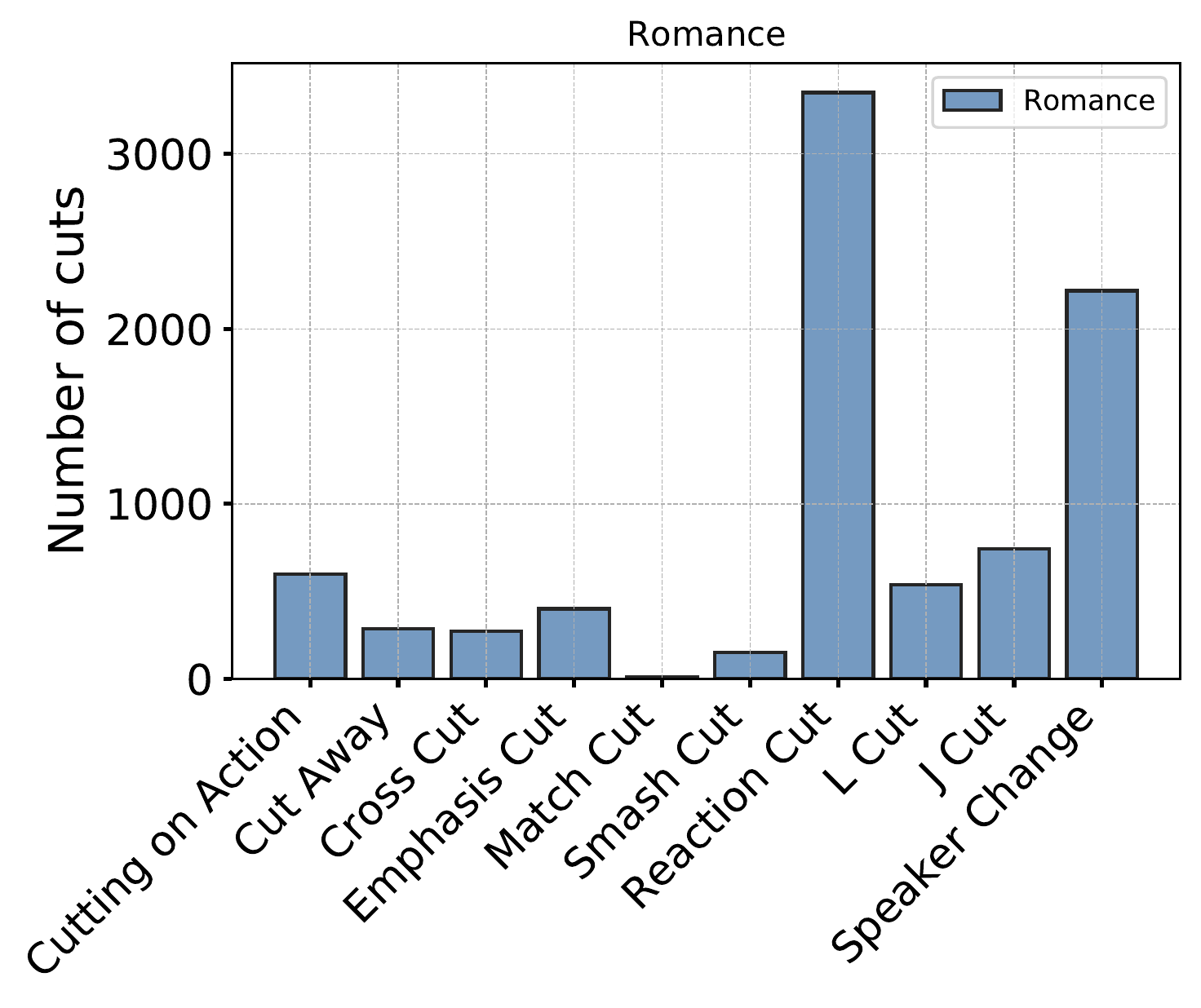}
    \includegraphics[width=0.95\textwidth]
    {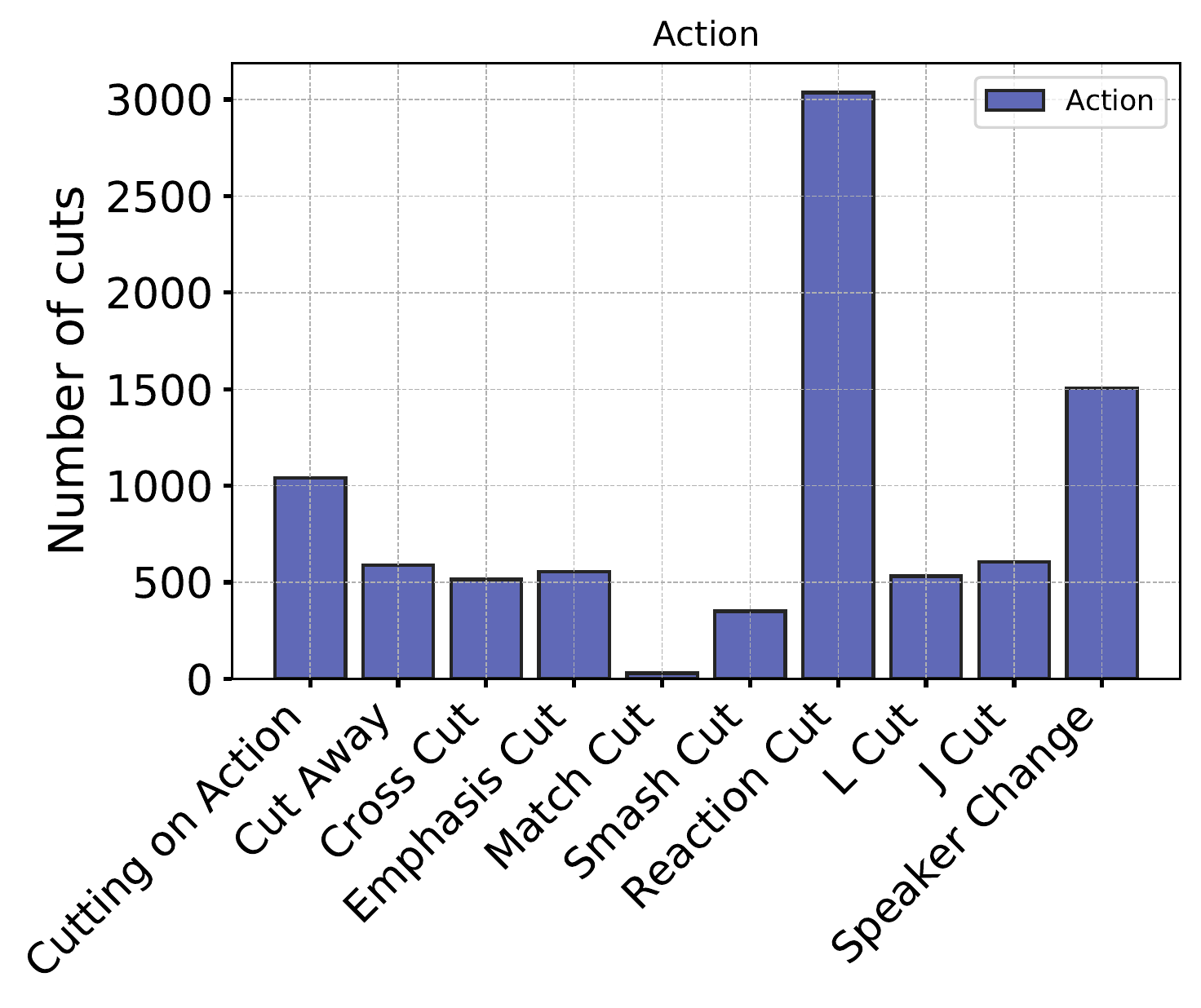}
    \includegraphics[width=0.95\textwidth]
    {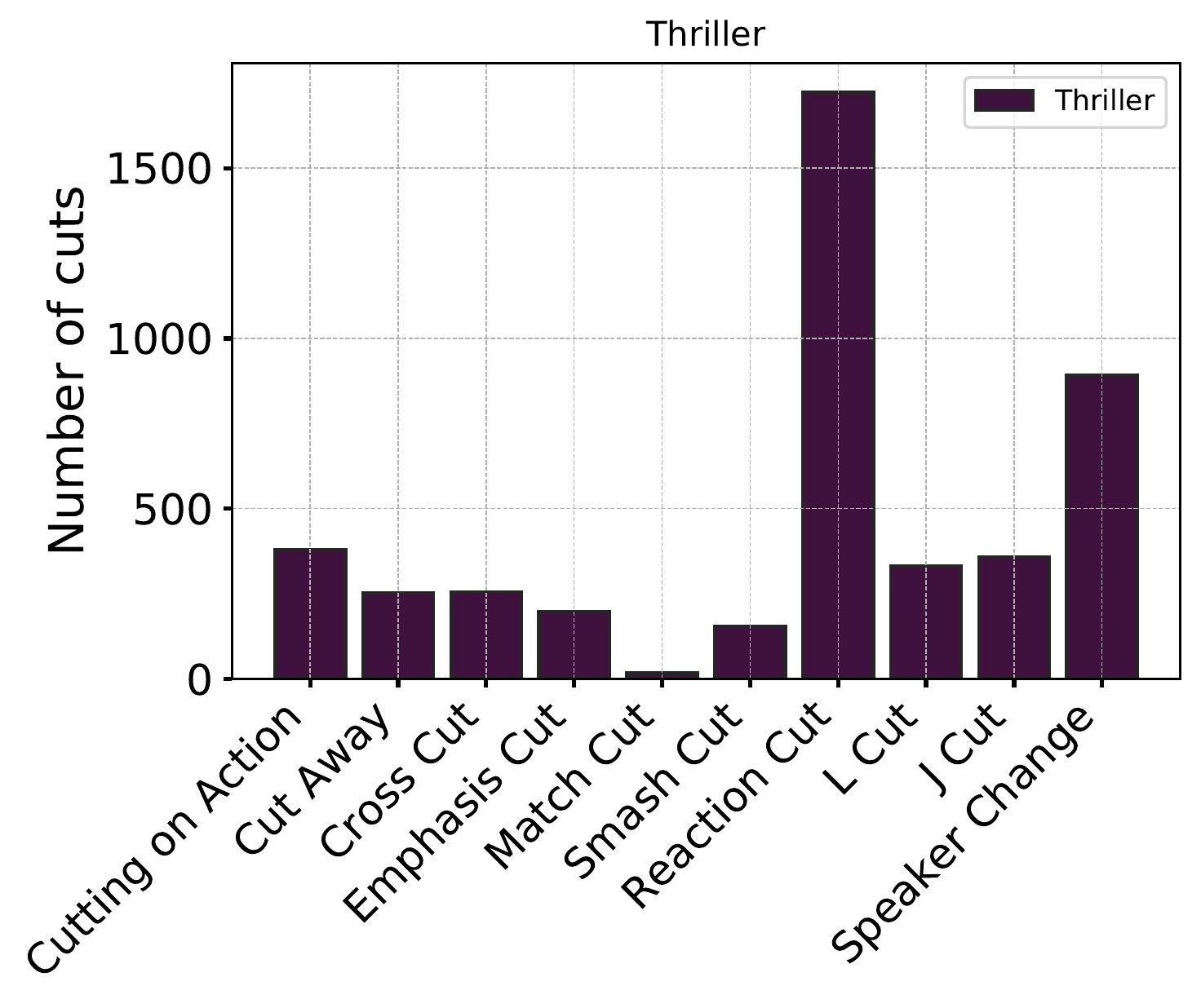}
    \caption{Test}
    \label{figure:label-distribution-test}
\end{subfigure}
\caption{\textbf{Cut-type distribution across movie genres and MovieCuts splits.}} Comedy, Drama, Romance, Action, Thriller.
\label{figure:label-distribution}
\end{figure*}

\begin{figure*}
\centering
\begin{subfigure}[t]{0.30\textwidth}
    \makebox[0pt][r]{\makebox[30pt]{\raisebox{50pt}{\rotatebox[origin=c]{90}{$Comedy$}}}}%
    \includegraphics[width=0.95\textwidth]
    {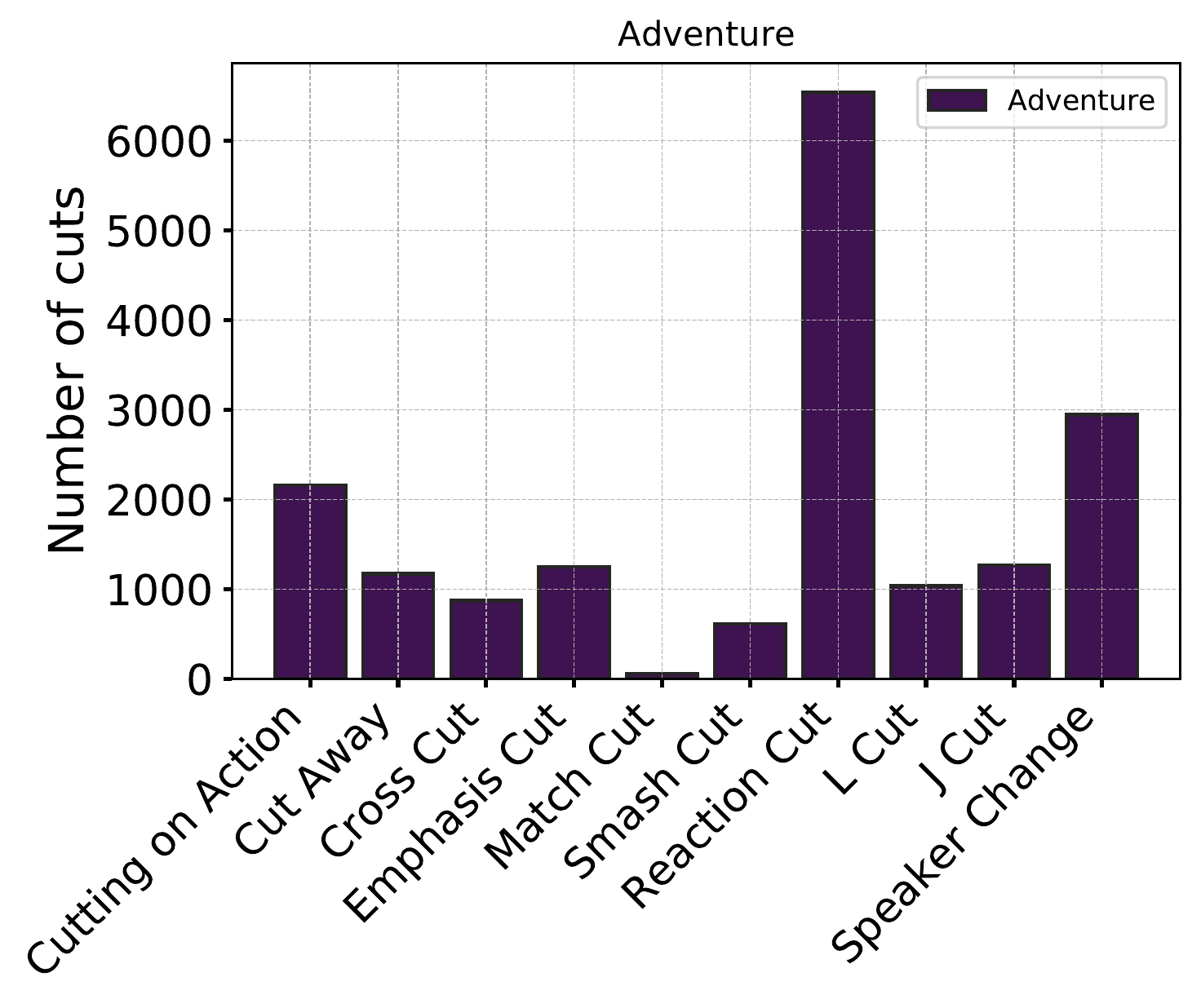}
    \makebox[0pt][r]{\makebox[30pt]{\raisebox{50pt}{\rotatebox[origin=c]{90}{$Crime$}}}}%
    \includegraphics[width=0.95\textwidth]
    {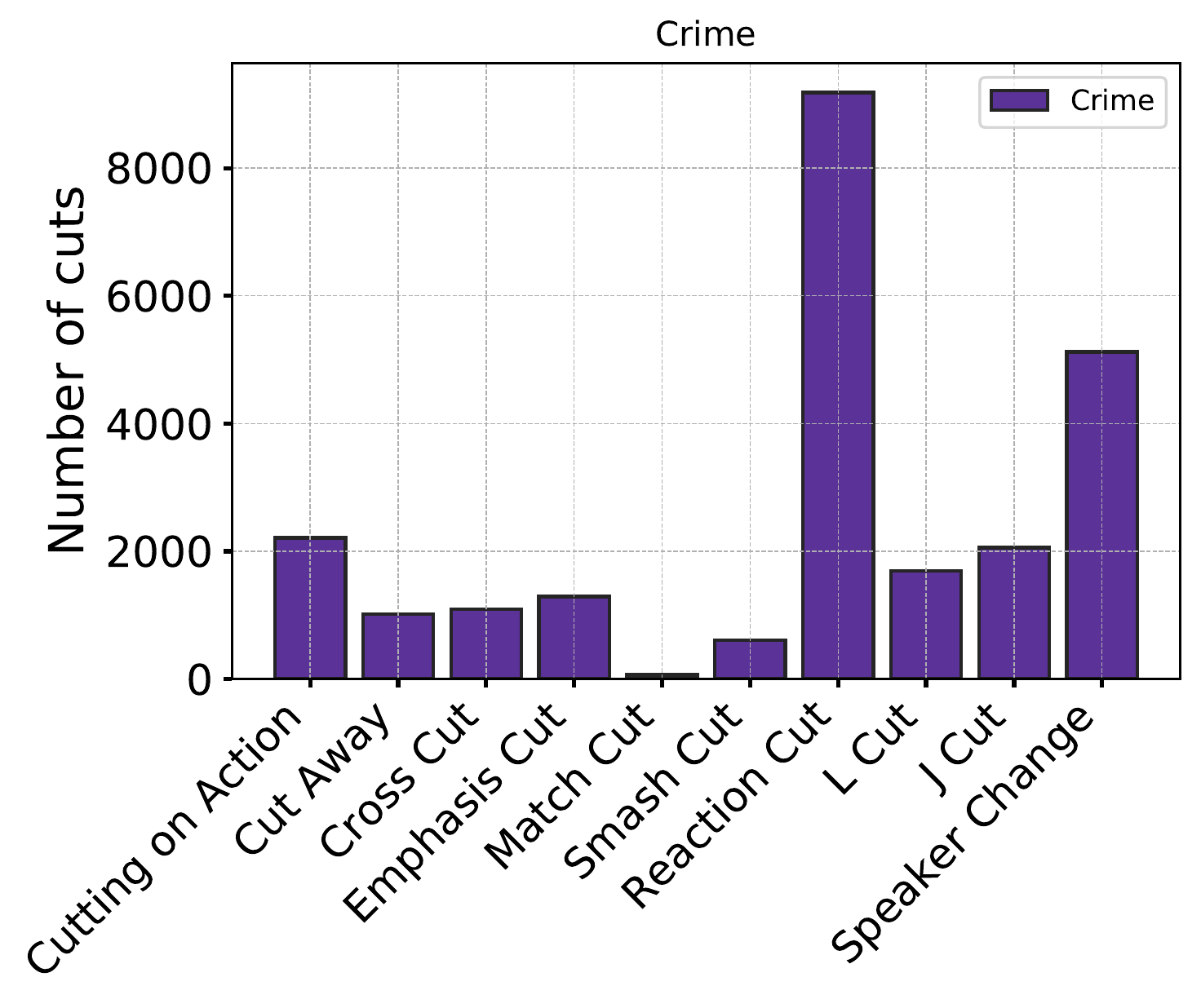}
    \makebox[0pt][r]{\makebox[30pt]{\raisebox{50pt}{\rotatebox[origin=c]{90}{$Fantasy$}}}}%
    \includegraphics[width=0.95\textwidth]
    {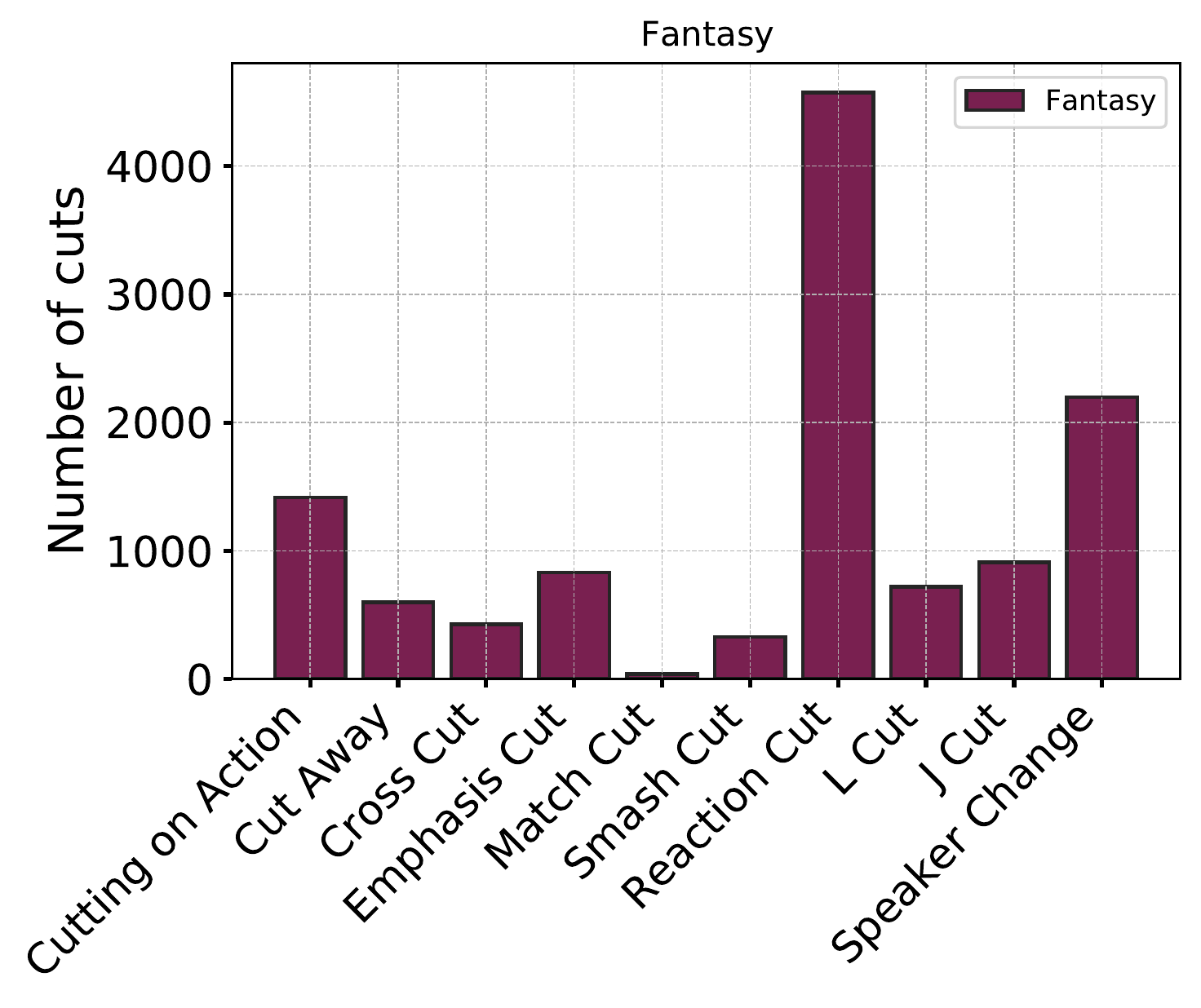}
    \makebox[0pt][r]{\makebox[30pt]{\raisebox{50pt}{\rotatebox[origin=c]{90}{$Horror$}}}}%
    \includegraphics[width=0.95\textwidth]
    {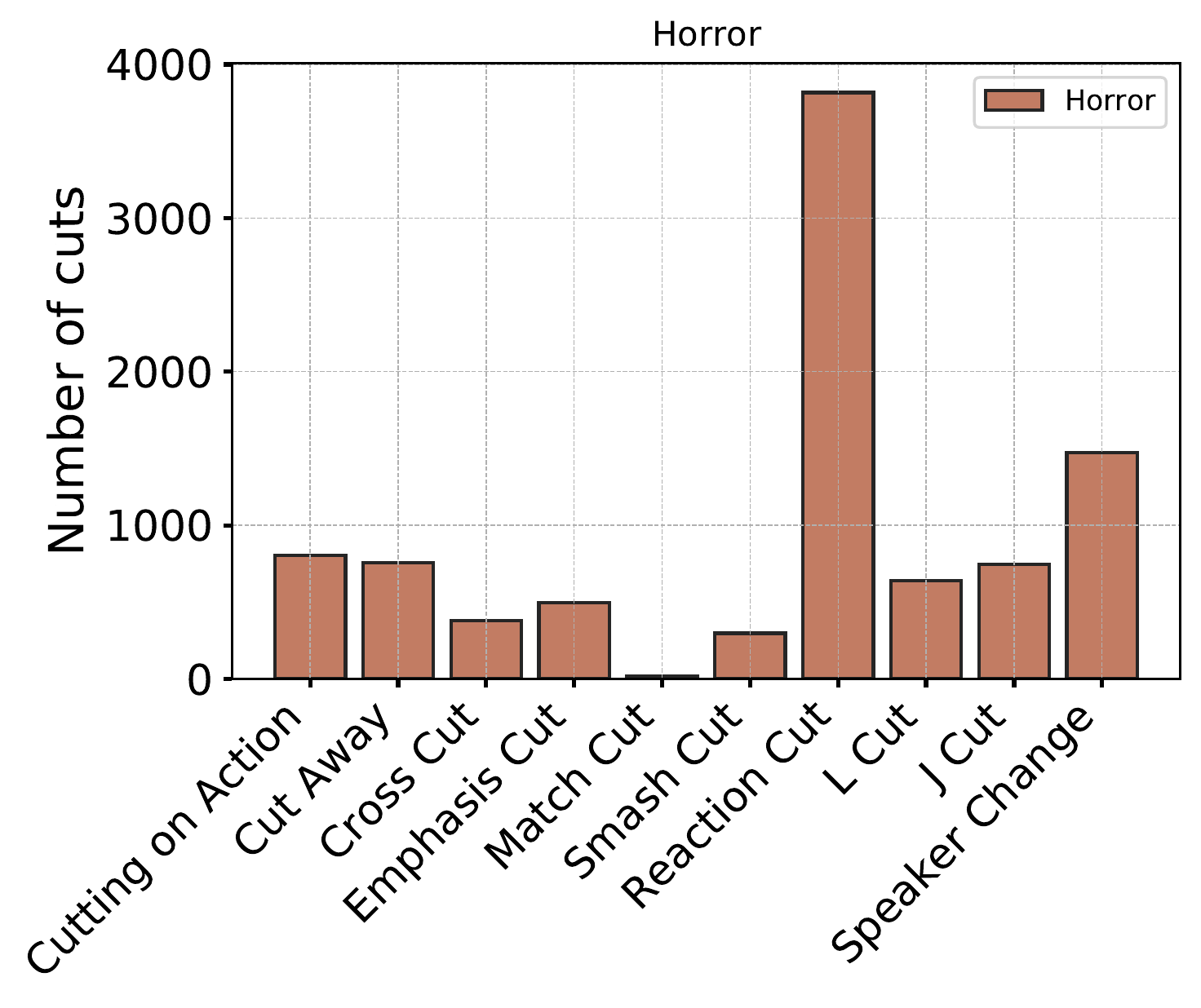}
    \makebox[0pt][r]{\makebox[30pt]{\raisebox{50pt}{\rotatebox[origin=c]{90}{$Sci-Fi$}}}}%
    \includegraphics[width=0.95\textwidth]
    {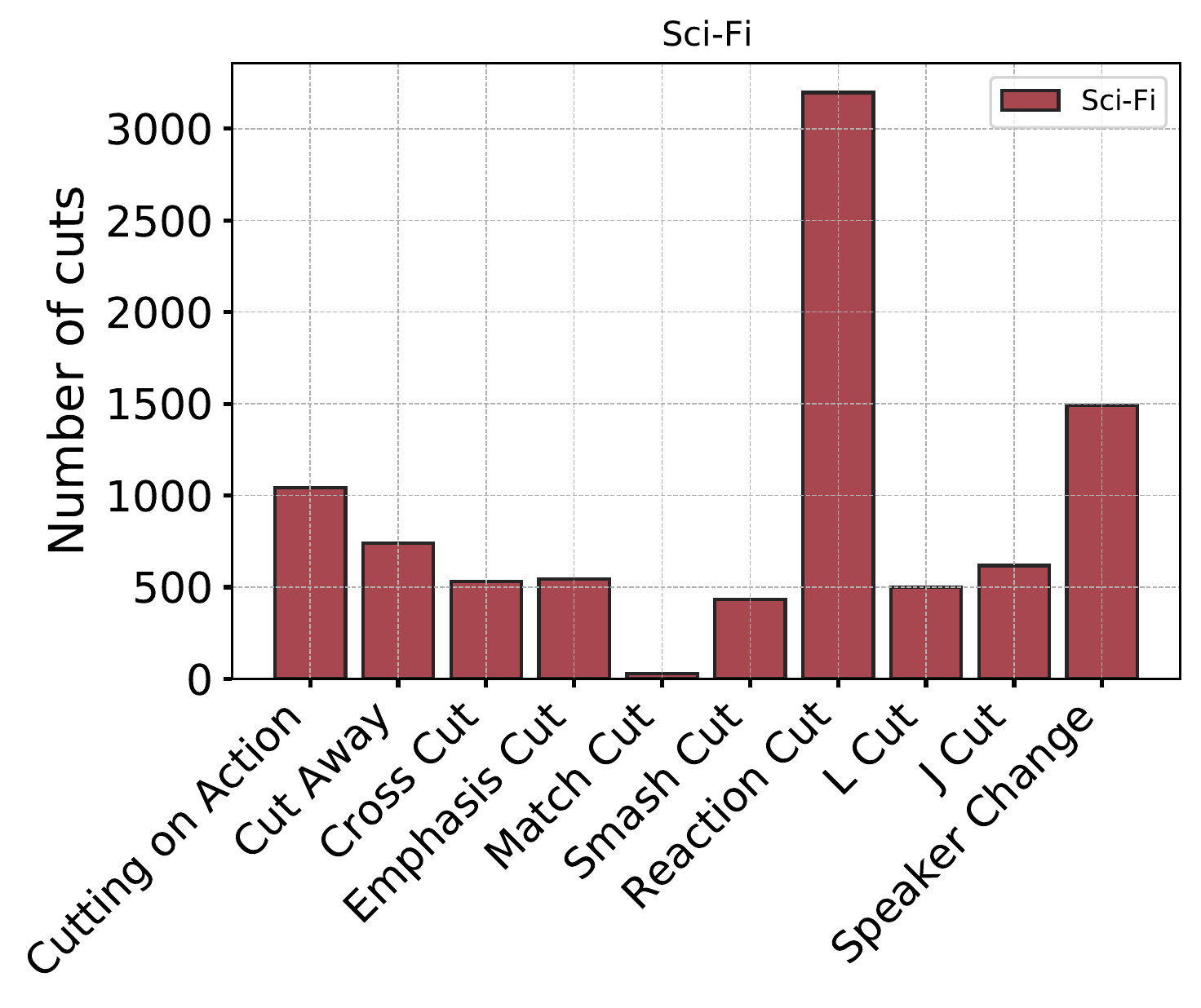}
    \caption{Train}
    \label{figure:label-distribution-train-2}
\end{subfigure}
\hfill
\begin{subfigure}[t]{0.30\textwidth}
    \includegraphics[width=0.95\textwidth]
    {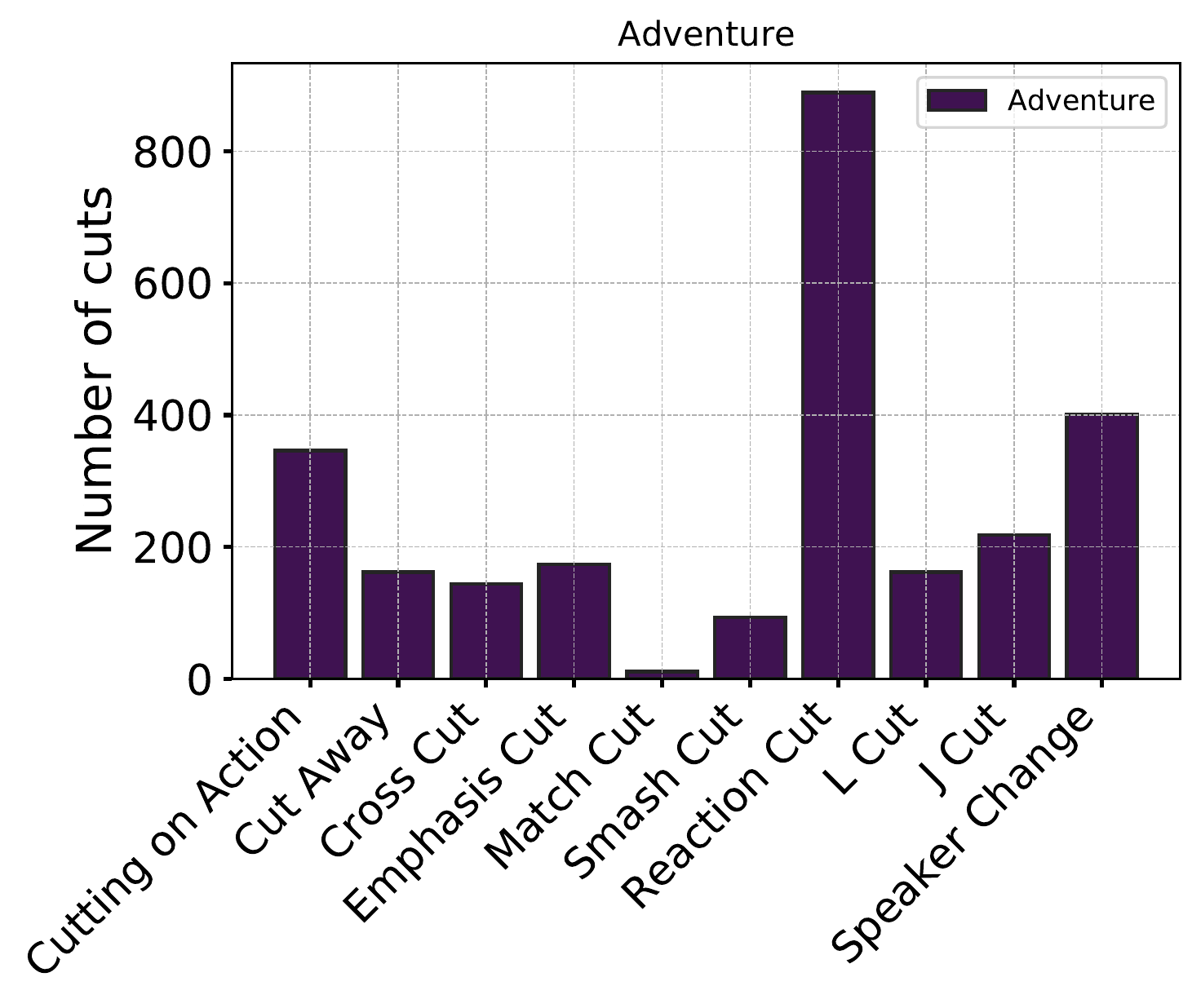}
    \includegraphics[width=0.95\textwidth]
    {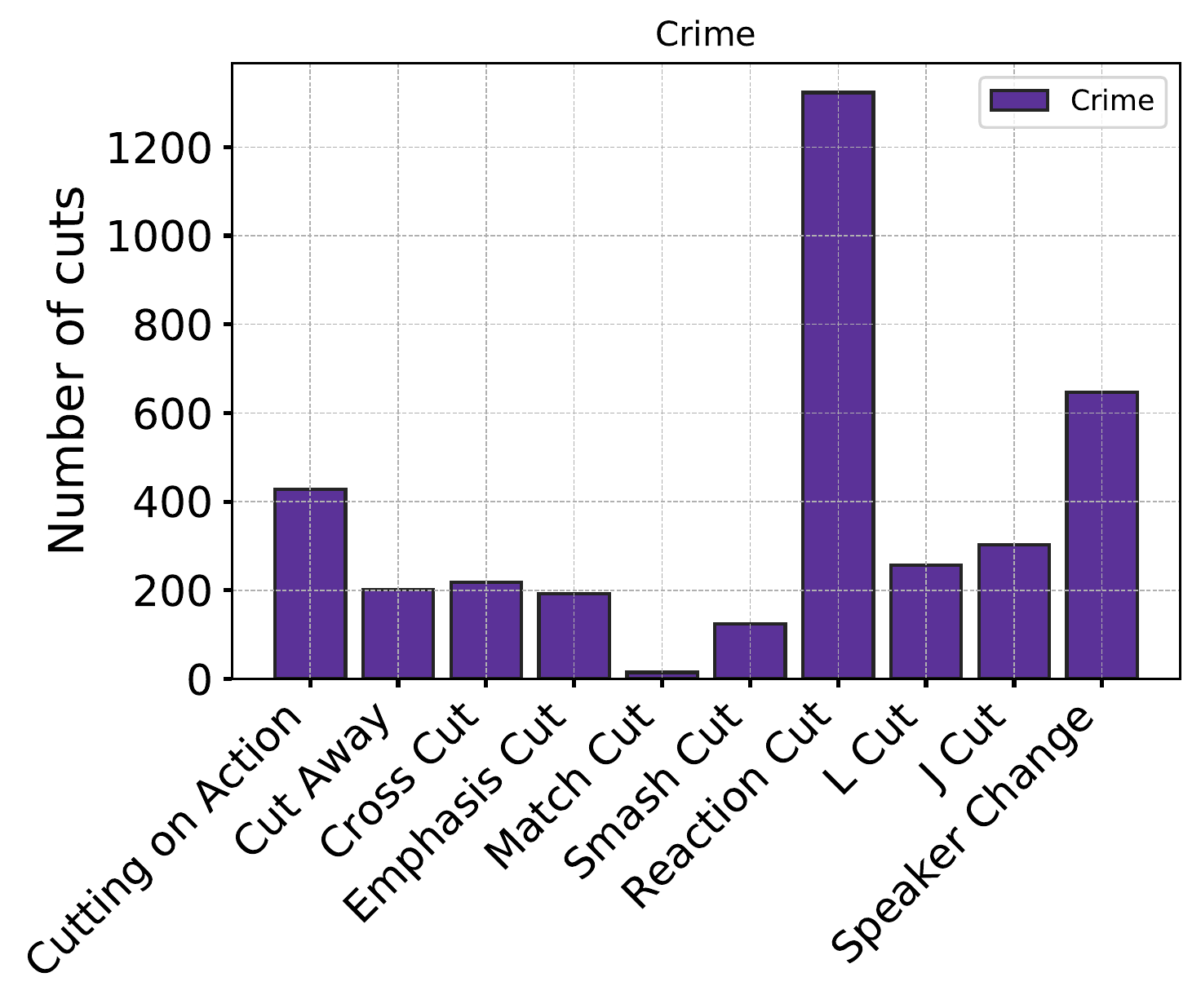}
    \includegraphics[width=0.95\textwidth]
    {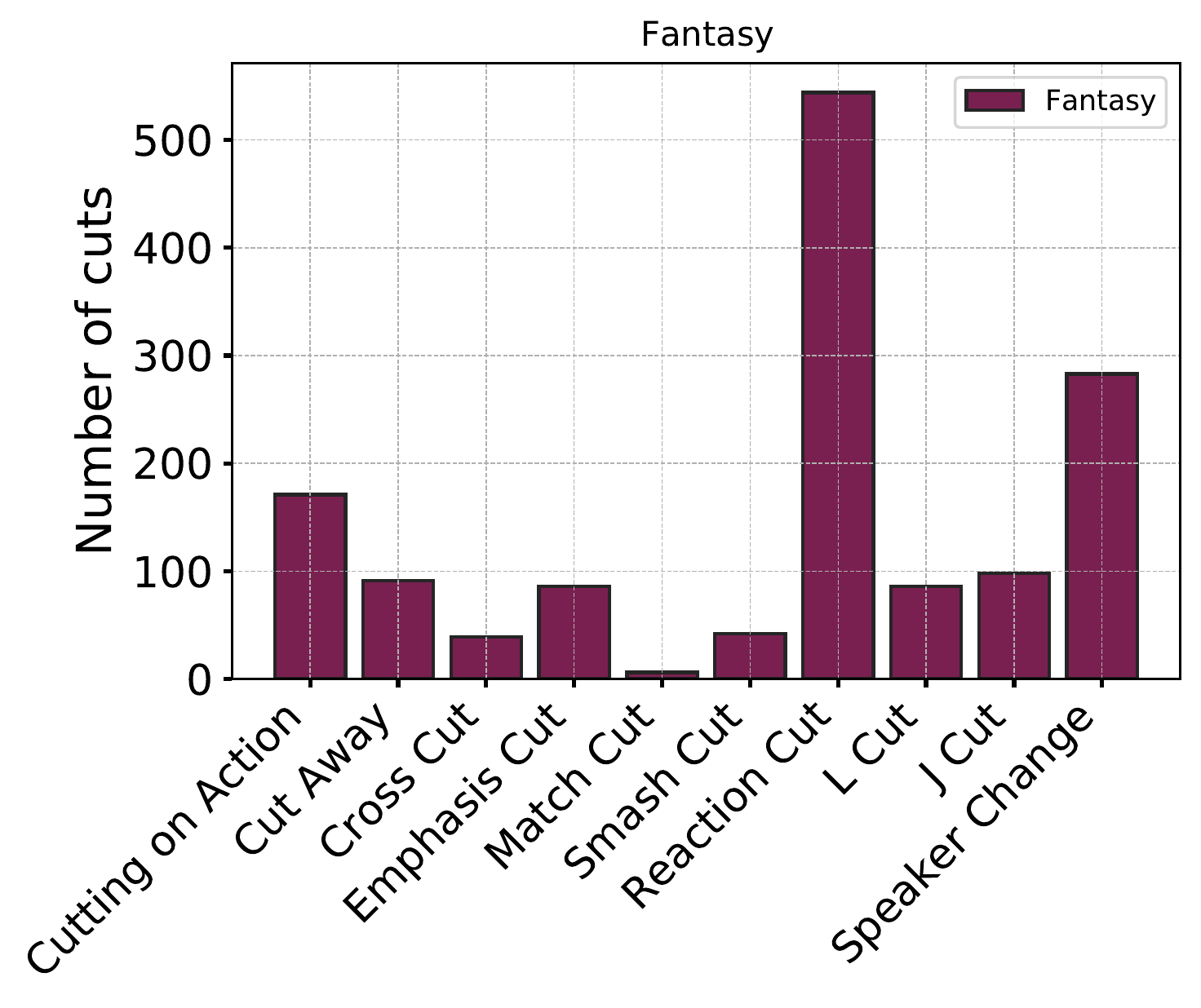}
    \includegraphics[width=0.95\textwidth]
    {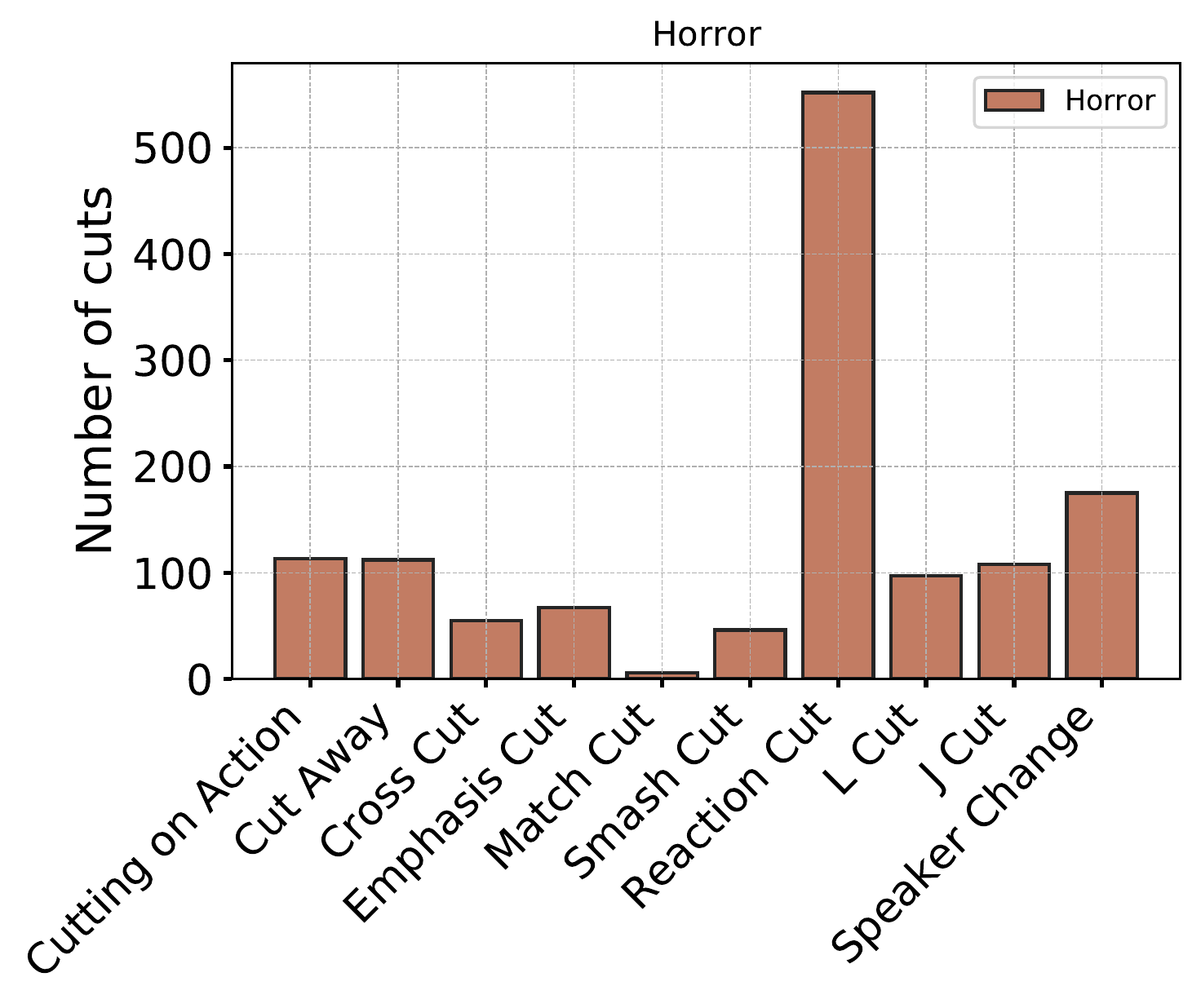}
    \includegraphics[width=0.95\textwidth]
    {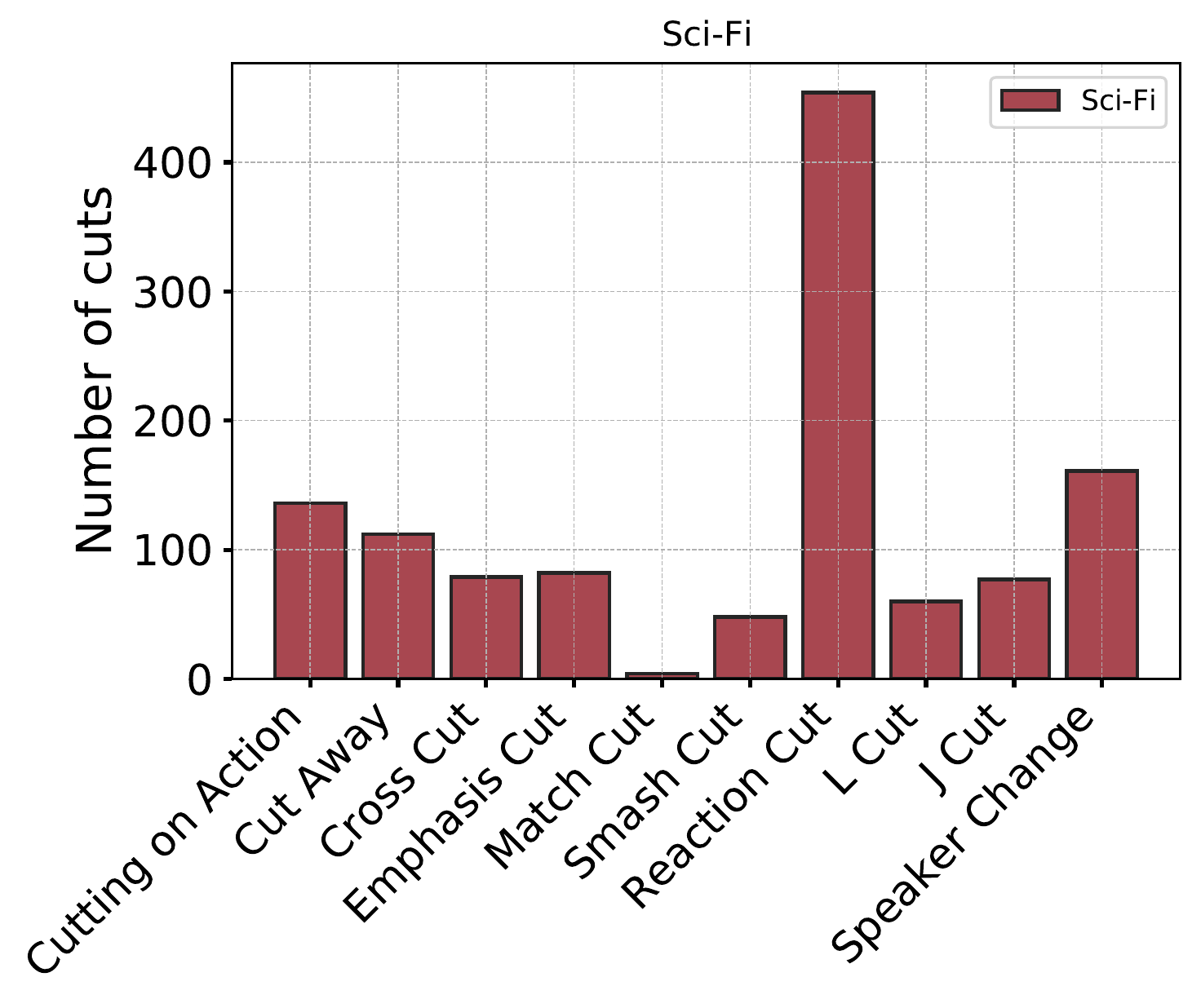}
    \caption{Validation}
    \label{figure:label-distribution-val-2}
\end{subfigure}
\hfill
\begin{subfigure}[t]{0.30\textwidth}
    \includegraphics[width=0.95\textwidth]
    {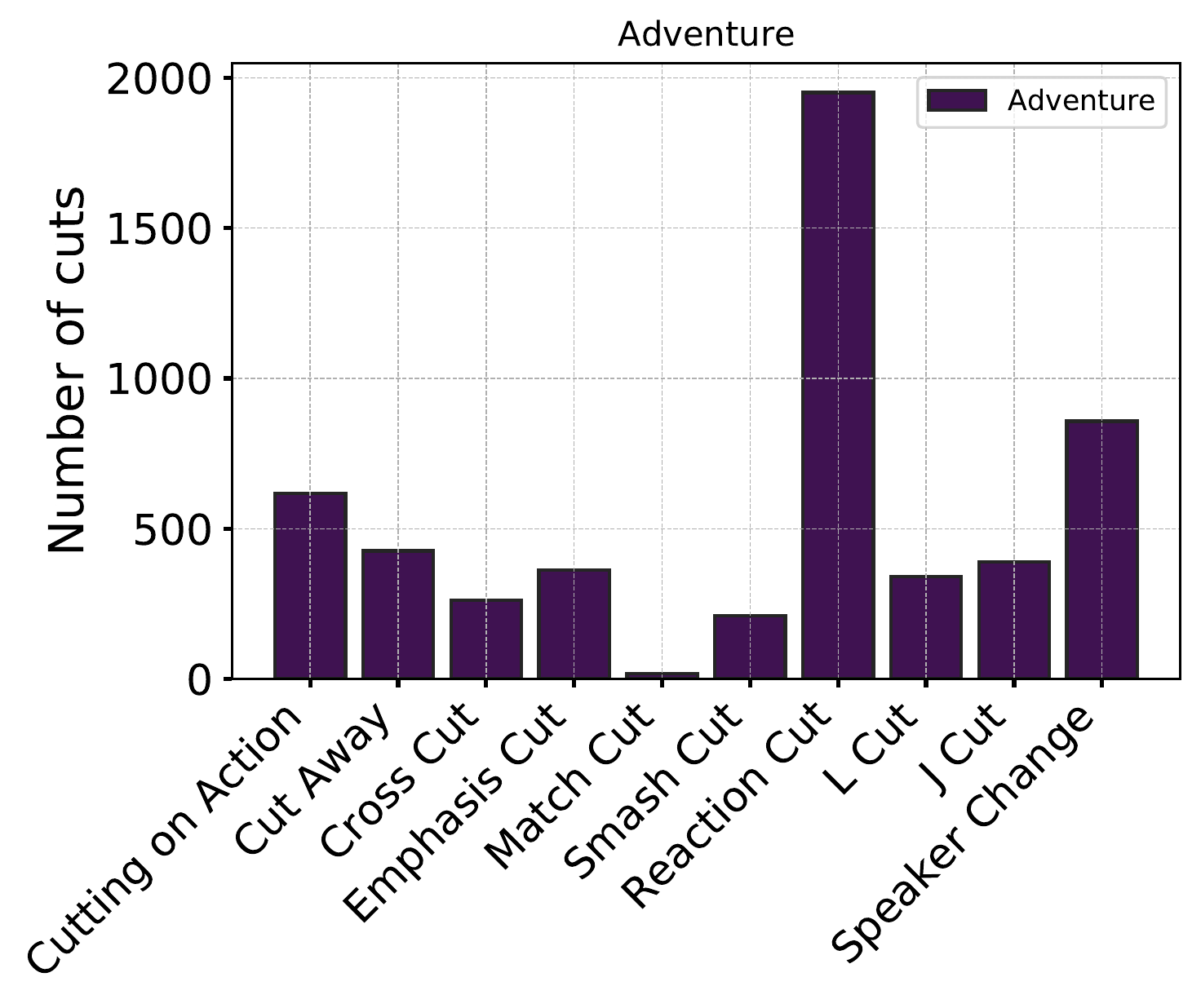}
    \includegraphics[width=0.95\textwidth]
    {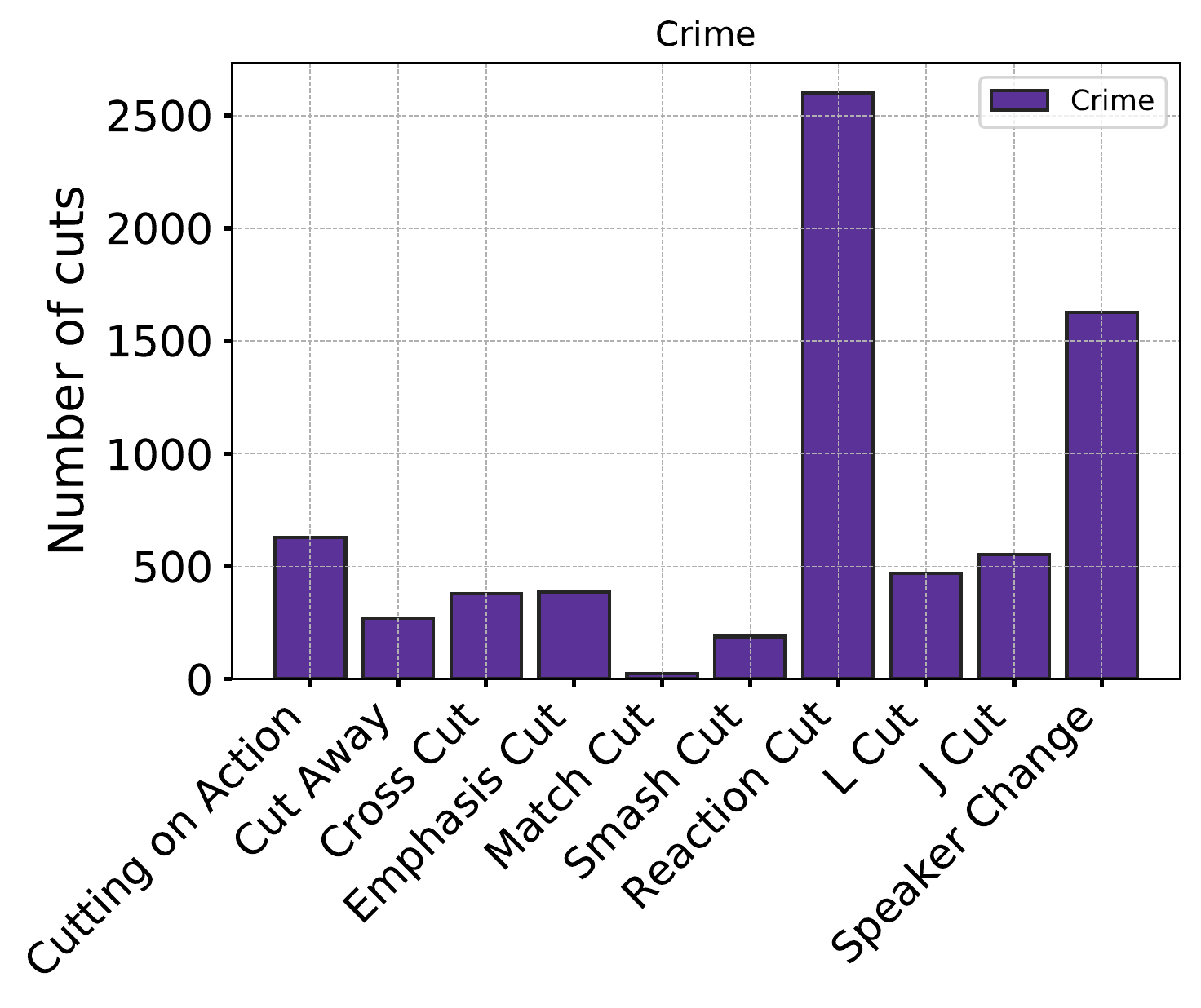}
    \includegraphics[width=0.95\textwidth]
    {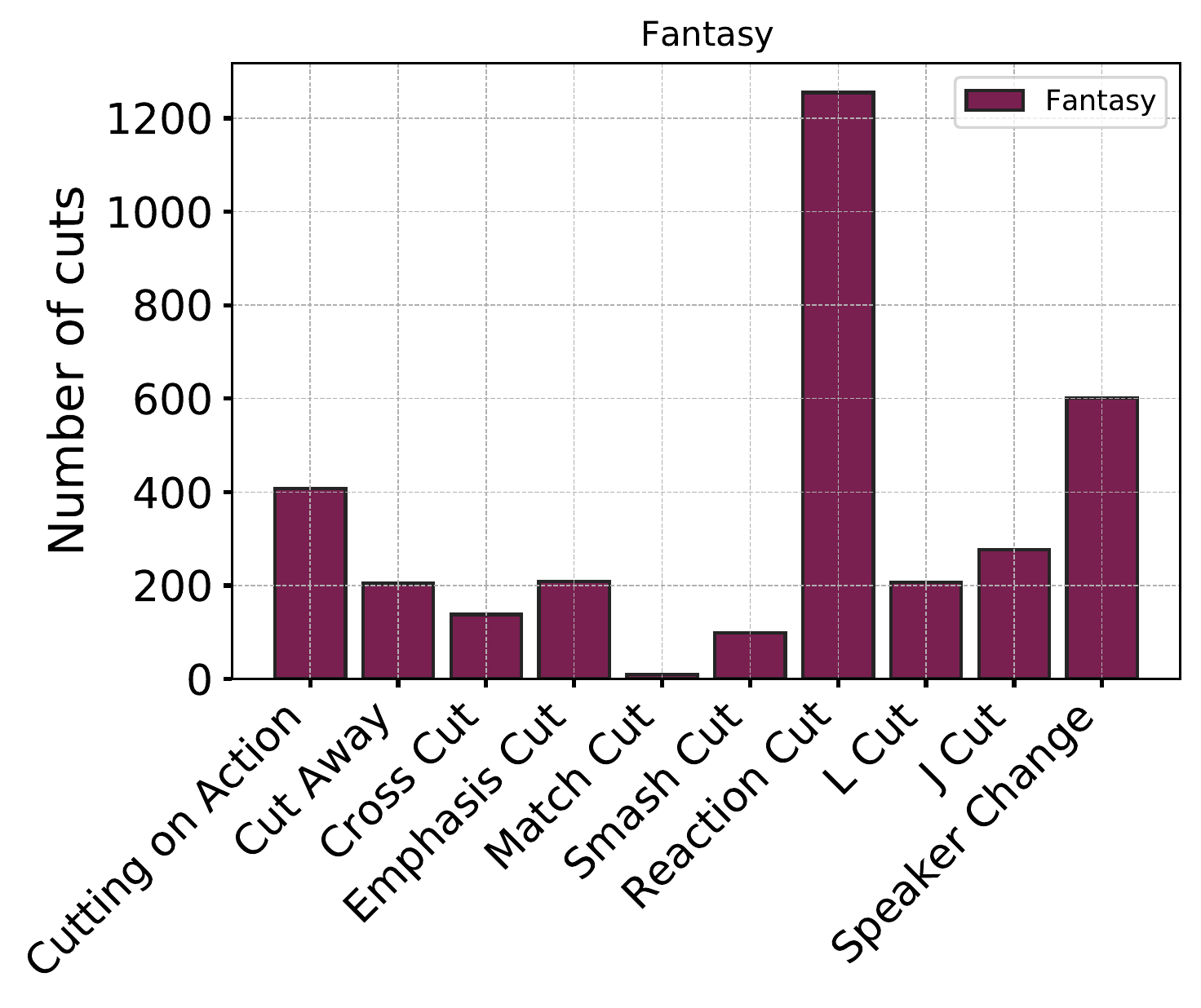}
    \includegraphics[width=0.95\textwidth]
    {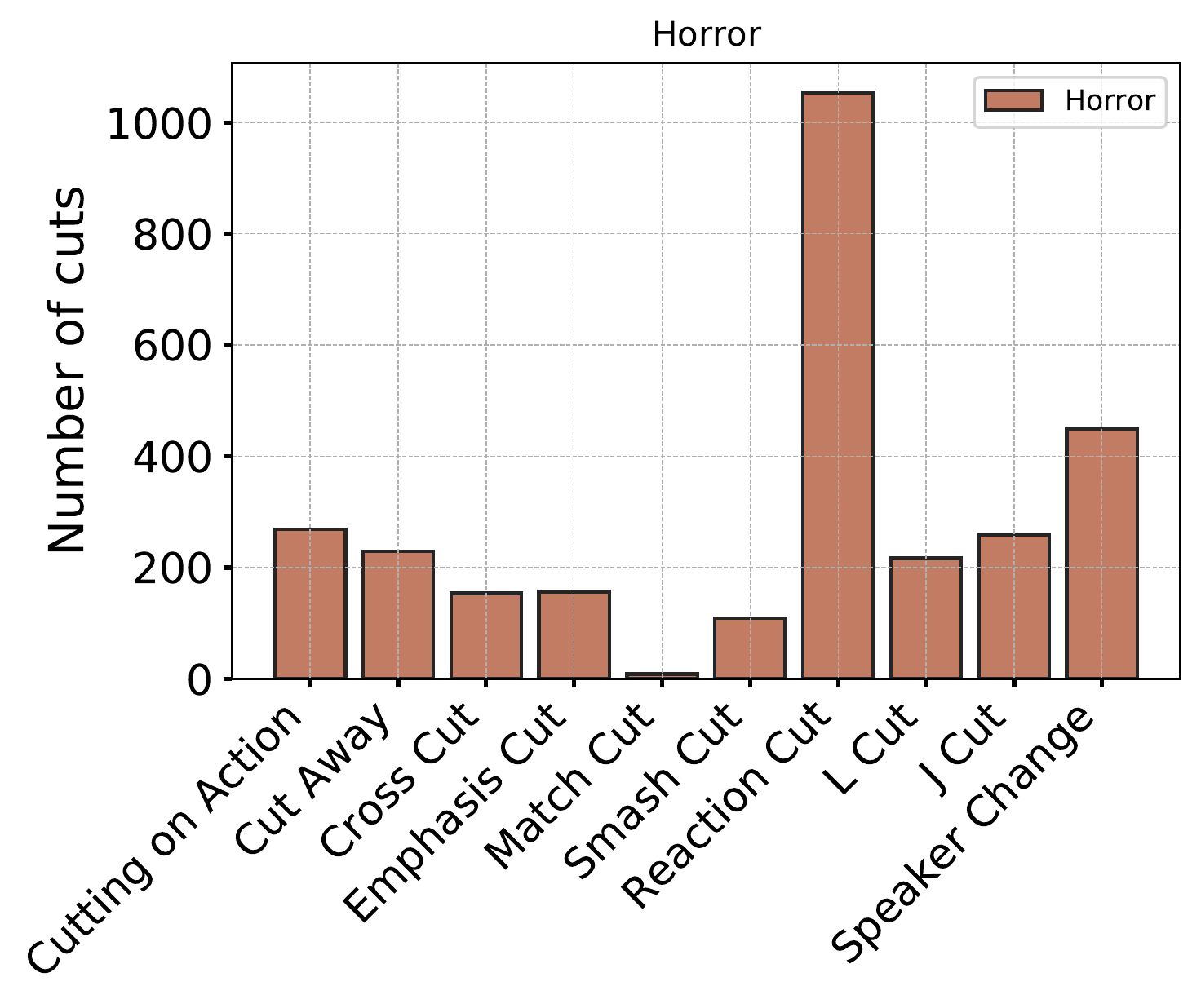}
    \includegraphics[width=0.95\textwidth]
    {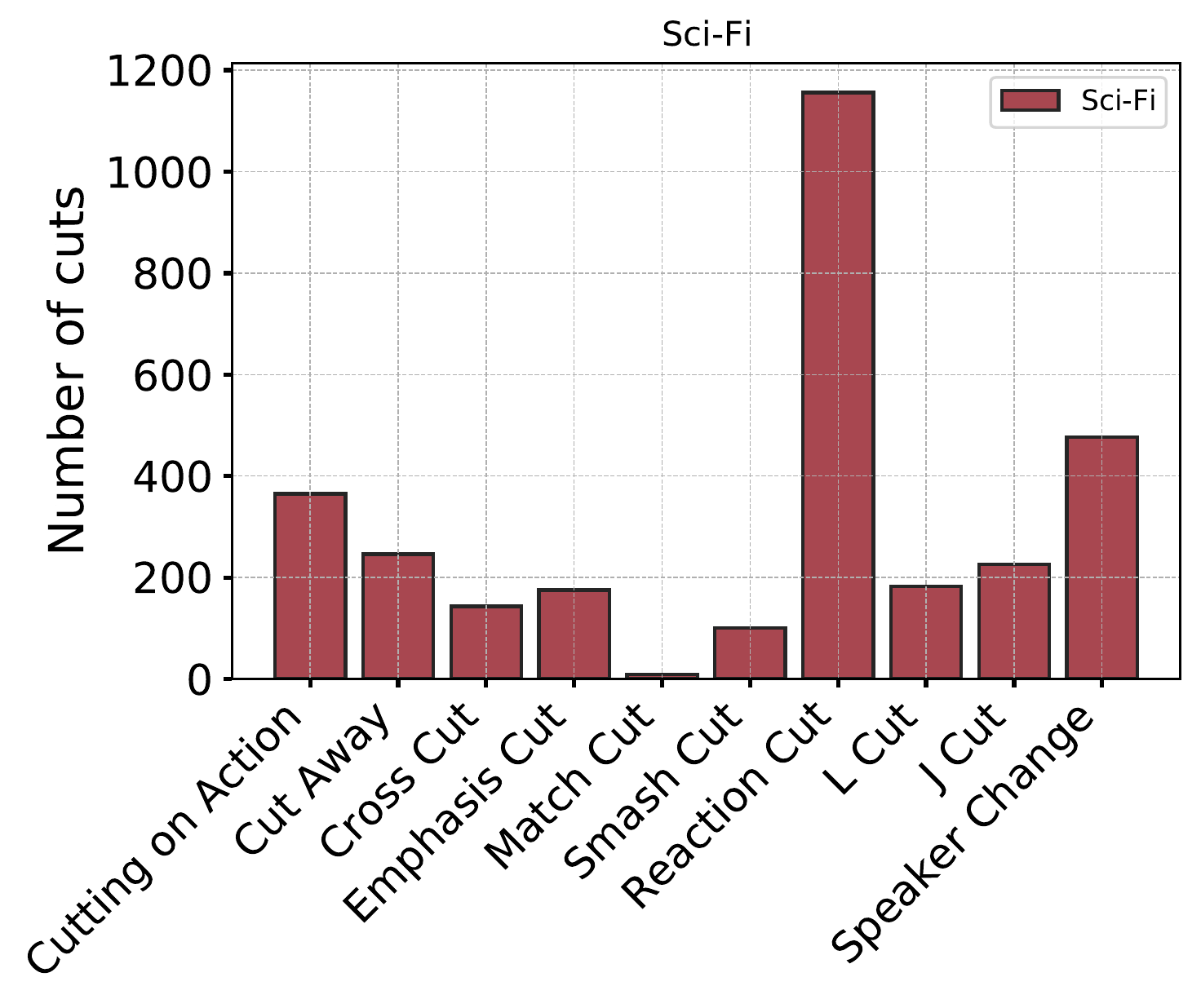}
    \caption{Test}
    \label{figure:label-distribution-test-2}
\end{subfigure}
\caption{\textbf{Cut-type distribution across movie genres and MovieCuts splits.}} Aventure, Crime, Fantasy, Horror, Sci-Fi.
\label{figure:label-distribution-2}
\end{figure*}

Additional to Figure \ref{figure:summary-counts}, in Figures \ref{figure:label-distribution} and \ref{figure:label-distribution-2} we show the distribution of classes for the most represented genres for the different splits, train \ref{figure:label-distribution-train} and \ref{figure:label-distribution-train-2}, validation \ref{figure:label-distribution-val} and \ref{figure:label-distribution-val-2}, and testing \ref{figure:label-distribution-test} and \ref{figure:label-distribution-test-2}. We see that the distributions across splits are independent and identically distributed (iid).

\subsection{Qualitative Results}

\begin{figure}[h!]
    \begin{center}
        \includegraphics[width=0.8\linewidth] {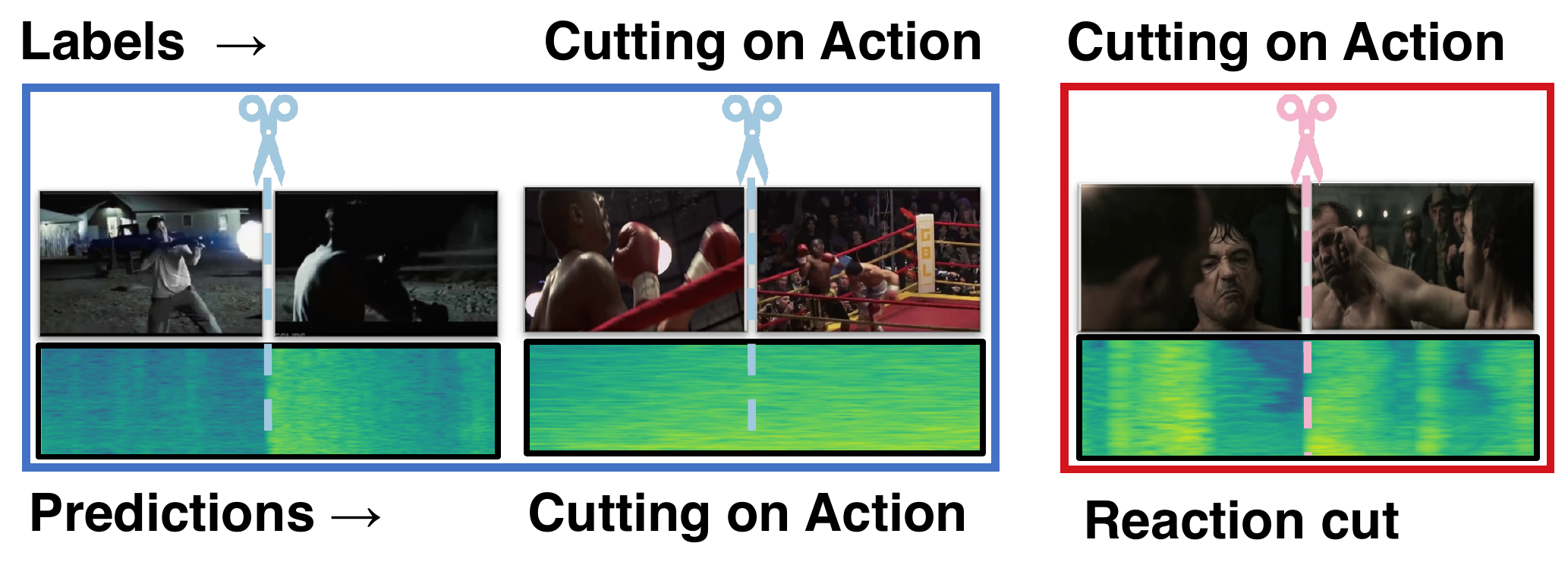}
    \end{center}
    \caption{\textbf{Qualitative results.} We showcase three examples of Cutting on Action. The blue box indicates True Positives, while the red box indicates a False Positive.}
    \label{fig:qualitative}
\end{figure}

We showcase representative qualitative results for the Cutting on Action class in Figure \ref{fig:qualitative}. We observe that the first two cuts are correctly classified as cutting on action, since the cut happens right after the action is performed (gunshot and boxing punch). The third example is a false positive. The model wrongly predicts it as a Reaction Cut. The model fails gracefully though; the shot focuses on the face of the actor right before the action, which is similar to what happens in a Reaction Cut. At the end, the actor is not reacting but is performing an action across the cut.

\section{Machine-assisted Video Editing with MovieCuts}

Section 4.5 presents the results of our model acting as an editor. We clarify the implementation details of such experiments and the user study. 

\noindent\textbf{Audio-Visual Model as a video editor.} The task we are trying to solve is the creation of an edited sequence $\mathbf{V}$ based on two raw sequences $\mathbf{A}$ and $\mathbf{B}$. The professional editor creates $\mathbf{V}$ by alternating between sequences $\mathbf{A}$ and $\mathbf{B}$ in the right places to cut (as explained in Section \ref{section:actiontrigger}). However, professional editors use different types of heuristics and rules to define these cut places. We argue that the Audio-Visual model trained on MovieCuts has some knowledge of these cut triggers and can perform cuts between the two sequences. Thus, we collected 11 edited sequences from \href{https://editstock.com/}{EditStock.com} with their corresponding unedited shots. We choose sequences that alternate between two shots. Thus, the task is to find the best places to transition between $\mathbf{A}$ and $\mathbf{B}$. After aligning the raw sequences, we create all possible cuts (transitions) from one shot to the other. Then, we score these possible cuts with our best model. We assign the maximum class score to each cut. Using these scores, we use only the top-k as good places to cut. Finally, we use these scores to perform cuts alternating between $\mathbf{A}$ and $\mathbf{B}$.
\begin{table}[b!]
    \small
	\centering
    \begin{subtable}[h]{0.48\textwidth}
    \centering
	\begin{tabular}{ l | c  c  c}
		Method & Purity & Coverage & F1 \\
		\hline
		\hline
		Random Frame & 99 & 10 & 18 \\
		\hline
		Random Snippet & 87 & 52 & 63 \\
		\hline
		Biased Random & 74 & 82 & 77 \\
		\hline
		\textbf{MovieCuts AV} & 80 & 82 & \textbf{81} \\
		\hline
	\end{tabular}
    \caption{\textbf{Quantitative Results.} Results are reported in $\%$. We use the human editor as ground truth and evaluate Purity, Coverage and F1 implemented in \cite{pyannote.metrics}.}
	\label{table:application_results_quan}
	\end{subtable}
	\hfill
	\begin{subtable}[h]{0.48\textwidth}
    \centering
	\begin{tabular}{ l | c }
		Method & vs Human Editor \\
		\hline
		\hline
		Random Frame & 1.8  \\
		\hline
		Random Snippet & 15.7  \\
		\hline
		Biased Random & 34.5 \\
		\hline
		\textbf{MovieCuts AV} & \textbf{38.1} \\
		\hline
	\end{tabular}
    \caption{\textbf{Qualitative Results.} Results are report in $\%$ of times that humans pick such method over the professional editing.}

	\label{table:application_results_qual}
	\end{subtable}
	\caption{\textbf{Video Editing Results}. Results of MovieCuts' automated video editing. Our method performs better than a set of baselines. 
	}
	\label{table:application_results}
\end{table}

We ask $63$ AMT turkers to pick among the edits done by professionals vs the automatic methods. The results are shown on table \ref{table:application_results_qual}. The AV Model trained on MovieCuts was the one picked the most over the professionals, showing that the edits made by it are preferred by users over the other methods.
Moreover, we use the edited sequences to create the ground-truth cut places and evaluate how close were the different automatic edits compared to these professionally edited sequences. In table \ref{table:application_results_quan}, we measure Purity, Coverage, and F1, implemented by \cite{pyannote.metrics}. Yet again, our method outperforms all the other automatic methods when comparing them with the professionally edited sequences. Please, be reminded that this editing process was done without training for it, but just using the model trained on MovieCuts as a scoring function for cut places. We argue that improvements in Cut-type recognition tasks can translate into advances in tasks related to machine-assisted video editing.




\end{document}